\documentclass{article}

\usepackage[a4paper, margin=1in]{geometry}

\usepackage[utf8]{inputenc} 
\usepackage[T1]{fontenc}    
\usepackage{hyperref}       
\usepackage{url}            
\usepackage{booktabs}       
\usepackage{amsfonts}       
\usepackage{amsthm}
\usepackage{nicefrac}       
\usepackage{microtype}      
\usepackage{lipsum}
\usepackage{graphicx}
\usepackage{soul}
\usepackage{multirow}
\usepackage[normalem]{ulem}
\usepackage{caption}
\usepackage{subcaption}
\usepackage{tablefootnote}
\usepackage{paralist}
\usepackage{indentfirst}
\usepackage{setspace}
\usepackage{authblk}
\usepackage{float}

\newcommand{\norm}[1]{\left\Vert#1\right\Vert}

\theoremstyle{definition}
\newtheorem{definition}{Definition}

\title{A Survey on Embedding Dynamic Graphs}

\author[1]{Claudio D. T. Barros}
\author[1]{Matheus R. F. Mendonça}
\author[2]{Alex B. Vieira}
\author[1]{Artur Ziviani}
\affil[1]{National Laboratory for Scientific Computing (LNCC),  Petrópolis, RJ, Brazil}
\affil[2]{Federal University of Juiz de Fora (UFJF),
Juiz de Fora, MG, Brazil}

\date{}

\begin{document}
\maketitle

\begin{abstract}

Embedding static graphs in low-dimensional vector spaces plays a key role in network analytics and inference, supporting applications like node classification, link prediction, and graph visualization. However, many real-world networks present dynamic behavior, including topological evolution, feature evolution, and diffusion. Therefore, several methods for embedding dynamic graphs have been proposed to learn network representations over time, facing novel challenges, such as time-domain modeling, temporal features to be captured, and the temporal granularity to be embedded. In this survey, we overview dynamic graph embedding, discussing its fundamentals and the recent advances developed so far. We introduce the formal definition of dynamic graph embedding, focusing on the problem setting and introducing a novel taxonomy for dynamic graph embedding input and output. We further explore different dynamic behaviors that may be encompassed by embeddings, classifying by topological evolution, feature evolution, and processes on networks. Afterward, we describe existing techniques and propose a taxonomy for dynamic graph embedding techniques based on algorithmic approaches, from matrix and tensor factorization to deep learning, random walks, and temporal point processes. We also elucidate main applications, including dynamic link prediction, anomaly detection, and diffusion prediction, and we further state some promising research directions in the area.
\end{abstract}


\section{Introduction}

Graph-structured networks arise naturally in several real-world complex systems, including social networks, biological networks, knowledge graphs, and finance, whose interactions between nodes allow the understanding of the structural information of these domains~\cite{Barabasi2016Network}. Therefore, graph-aware learning tasks play a key role in machine learning and network science literature, and scalable approaches to deal with these high-dimensional non-Euclidean data have been explored to address the computational challenges associated with graph data-driven analytics and inference. Embedding graphs in low-dimensional vector spaces have been applied to extract features from networks and encoding topological and semantic information, and many researchers have been using these network representation learning approaches for several applications, including node classification and clustering, link prediction, and network visualization~\cite{Hamilton2017Representation, Cai2018Comprehensive, Goyal2018Graph}.

Previous work on network representation learning has focused on static graphs, either representing existing connections at a fixed moment (i.e., a graph snapshot), or node interactions over time that are aggregated into a single static graph~\cite{Dunlavy2011Temporal}. However, several real networks display dynamic behavior, including nodes and edges being added or removed from the system~\cite{Casteigts2012Time}, labels and other properties changing over time~\cite{Li2017Attributed}, and diffusion in the network~\cite{Trivedi2018Representation}. The network temporal correlations are lost during the aggregating process and, in this sense, approaches to develop embedding methods for dynamic networks have been proposed over the past few years. These efforts have improved tasks such as link prediction~\cite{Zhu2016Scalable} and node classification~\cite{Goyal2018Dyngem} over time, while enabling novel applications, including event prediction~\cite{Trivedi2018Representation}, anomaly detection~\cite{Goyal2018Dyngem} and diffusion prediction~\cite{Li2017DeepCas}.

Several challenges arise when developing an approach to embed dynamic graphs, as (i) how to model the time domain, i.e. discrete-time or continuous-time, (ii) which dynamic behaviors are desired to be captured, and (iii) which temporal granularity will be represented in the vector space, i.e. the same granularity as the dataset, or a coarser granularity summarizing dynamics in a finer timescale. Considering the increasing number of studies proposing dynamic graph embedding techniques, these discussions are becoming more important to advance the comprehension of dynamic network representation learning. Therefore, in this survey, we overview the problem of embedding dynamic graphs, discussing its fundamental aspects and the recent advances that have been made so far. We introduce the formal definition of dynamic graph embedding, discussing different dynamic network models whose representations may be extracted, and introducing a novel taxonomy for the problem settings, i.e. embeddings input and output. Moreover, we explore and classify different dynamic behaviors that may be captured by embeddings, describe existing methods, discuss their similarities and differences, and propose a detailed taxonomy based on algorithmic approaches.

To the best of our knowledge, a few attempts have been made so far to survey dynamic graph embeddings. Kazemi et al.~\cite{Kazemi2020Representation} focus on recent representation learning techniques for dynamic graphs by using an encoder-decoder framework. Xie et al.~\cite{Xie2020Survey} propose a taxonomy based on algorithmic approaches to encode graphs. Additionally, Skarding et al.~\cite{Skarding2021Foundations} survey how the dynamic network topology can be modeled using dynamic graph neural networks. Our work is different from the aforementioned surveys since we discuss different dynamic network models that have been or may be used for embedding, in addition to detecting temporal behaviors in networks that can be captured. Moreover, we extend dynamic graph embedding techniques taxonomy, encompassing methods based on graph kernels, temporal point processes, and agnostic methods. In this sense, this survey has the following contributions:

\begin{compactitem}
    \item A taxonomy of dynamic graph embedding based on problem settings, extending graph embedding input and output to handle temporal heterogeneity (i.e. timestamps with labels, classifying network behavior over time) and temporal embeddings (i.e. different temporal granularities to represent in the low-dimensional vector space);
    \item A discussion about different dynamic behaviors in networks that embedding models may capture, including topological evolution (concerning both node and edge addition or removal), feature evolution (regarding changes of nodes/edges features or labels over time) and processes on networks (diffusion and global role of nodes and its evolution). Furthermore, we also bring some perspectives about temporal point processes on networks.
    \item A detailed analysis of embedding techniques for dynamic graphs, focused on a classification concerning their algorithmic approaches, comparing different methods proposed in the literature and discussing their similarities, differences and other particularities;
    \item The categorization, according to the topological structure, of several dynamic graph embedding applications, focused on: node related tasks, edge related tasks, node, and edge related tasks, and graph-related tasks;
    \item A discussion of future research directions in the area in terms of problem settings, solution techniques, and modeling, in addition to applications and representation learning on generalized graphs~(i.e. hypergraphs and higher-order graphs).
\end{compactitem}

The remainder of this survey is organized as follows. In Section~\ref{sec:fundamentals}, we introduce the fundamentals behind the embedding of dynamic graphs, reviewing static graph embedding, defining the problem of dynamic graph embedding, presenting different dynamic graph models explored in the embedding scenario, along with other problem settings, including the dynamic graph embedding input and output as well as the dynamic behaviors that may be captured. In Section~\ref{sec:tech}, we categorize the literature based on the embedding techniques, unraveling insights behind each paradigm and we provide a detailed comparison of different techniques. After that, we present in Section~\ref{sec:apps} some concrete examples of applications enabled by dynamic graph embedding methods discussed in Section \ref{sec:tech}, allowing the reader to better grasp the practical utility of these methods. Finally, our conclusions are presented in Section~\ref{sec:conc}, alongside some discussions on potential future research directions in the field of dynamic graph embedding.

\section{Fundamentals Behind the Embedding of Dynamic Graphs}
\label{sec:fundamentals}

In this section, we first review basic graph concepts and static graph embedding, introducing the definition of dynamic graphs. Then, we formally define the dynamic graph embedding problem. Thereafter, we describe possible problem settings, starting with dynamic graph embedding input and detailing different outputs. We expand the initial graph embedding concepts considering time-varying graph models, time granularity, and temporal aggregation, and discuss dynamic behaviors that may be captured by embeddings.

\subsection{Graphs and Static Graph Embedding}
\label{subsection:staticgraphembedding}

A \textbf{graph} $G = (V,E)$ is a mathematical structure, where $V = \{v_{1}, ..., v_{N}\}$ is a finite set of $N$ nodes (vertices), and $E \subseteq \{(v_{i},v_{j})\, \vert \,(v_{i},v_{j}) \in V\times V\}$ is a finite set of unordered pairs of vertices, whose elements $e_{ij} = (v_{i},v_{j})$ are called edges (links). The \textbf{adjacency matrix} $A$ of a graph is an $N \times N$ matrix whose element $A_{ij} = 1$ if edge $e_{ij} \in E$, or $A_{ij} = 0$ otherwise. A \textbf{directed graph} is a graph in which an edge $e_{ij} \in E$ is an ordered pair, i.e. the edge $e_{ij}$ is oriented. Otherwise, the graph is \textbf{undirected}. A \textbf{weighted graph} is a graph in which a weight function $w: E \rightarrow \mathbb{R}$ is assigned to it. Each edge has a weight associated with it, and it is possible to define a \textbf{weight matrix} $W$ such that $W_{ij} = w(e_{ij})$. Otherwise, the graph is \textbf{unweighted}. A \textbf{homogeneous graph} is a graph in which the number of node types $\mathcal{L}^{n}$ and the number of edge types $\mathcal{L}^{e}$ is 1, i.e. $\vert\mathcal{L}^{v}\vert = \vert\mathcal{L}^{e}\vert = 1$, and every node in $G$ belongs to a single node category and every edge belongs to a single edge category. A \textbf{heterogeneous graph} is a graph in which $\vert\mathcal{L}^{v}\vert > 1$ or $\vert\mathcal{L}^{e}\vert > 1$.

Different ways to define \textbf{proximity} or \textbf{similarity} between nodes in a graph may be conceived~\cite{Cai2018Comprehensive}. The \textbf{first-order proximity} $S_{ij}^{(1)}$ between nodes $v_{i}$ and $v_{j}$ is the weight of the edge $e_{ij}$, i.e., $W_{ij}$. The \textbf{second-order proximity} $S_{ij}^{(2)}$ between nodes $v_{i}$ and $v_{j}$ is a similarity between $v_{i}$'s neighbourhood $S_{i}^{(1)}$ and $v_{j}$'s neighbourhood $S_{j}^{(1)}$ given by some defined metric, where $S_{i}^{(1)} = [S_{i1}^{(1)}, ..., S_{iN}^{(1)}]$ and $S_{j}^{(1)} = [S_{j1}^{(1)}, ..., S_{jN}^{(1)}]$. \textbf{Higher-order proximities} can be defined as well, including the Katz centrality~\cite{Katz1953New}, which is a weighted summation over the paths between two vertices in the graph whose weight is an exponential decay function of its length, and the Adamic/Adar Index~\cite{Adamic2003Friends}, which counts the number of vertices connecting two nodes taking into account a weight depending on the reciprocal of the neighbor's degree.

\begin{figure}[tp]
    \centering
    \begin{minipage}[t]{0.3\textwidth}
    	\begin{subfigure}[b]{\textwidth}
         \centering
         \includegraphics[width=0.7\linewidth]{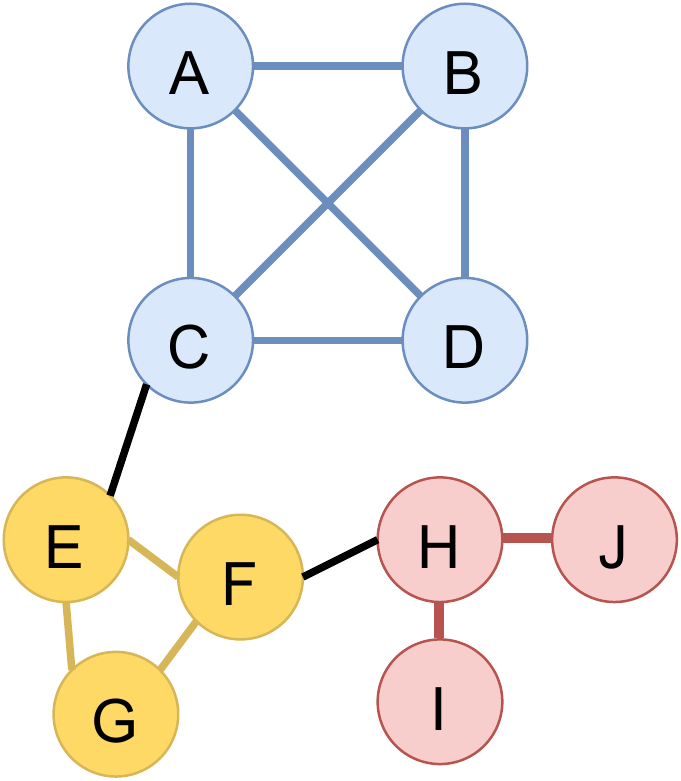}
         \caption{Graph $G$}
         \vspace*{-3cm}
         \label{fig:graph}
     \end{subfigure}
     \hfill
    \end{minipage}
    \begin{minipage}[t]{0.65\textwidth}
    	\begin{subfigure}[b]{0.35\textwidth}
        \centering
         \includegraphics[width=\linewidth]{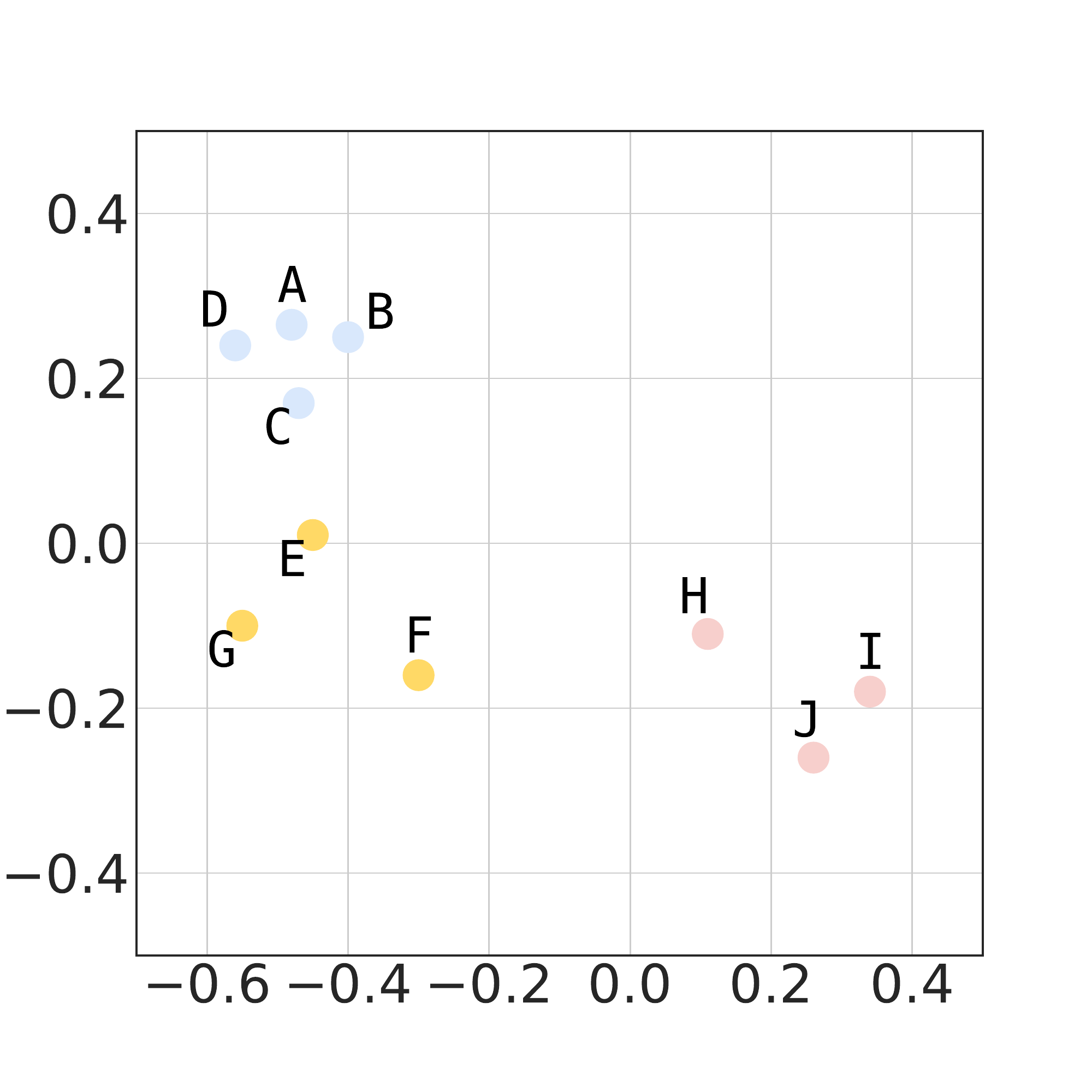}
         \caption{Node Embedding}
         \label{fig:nodeembedding}
     \end{subfigure}
     \begin{subfigure}[b]{0.35\textwidth}
         \centering
         \includegraphics[width=\linewidth]{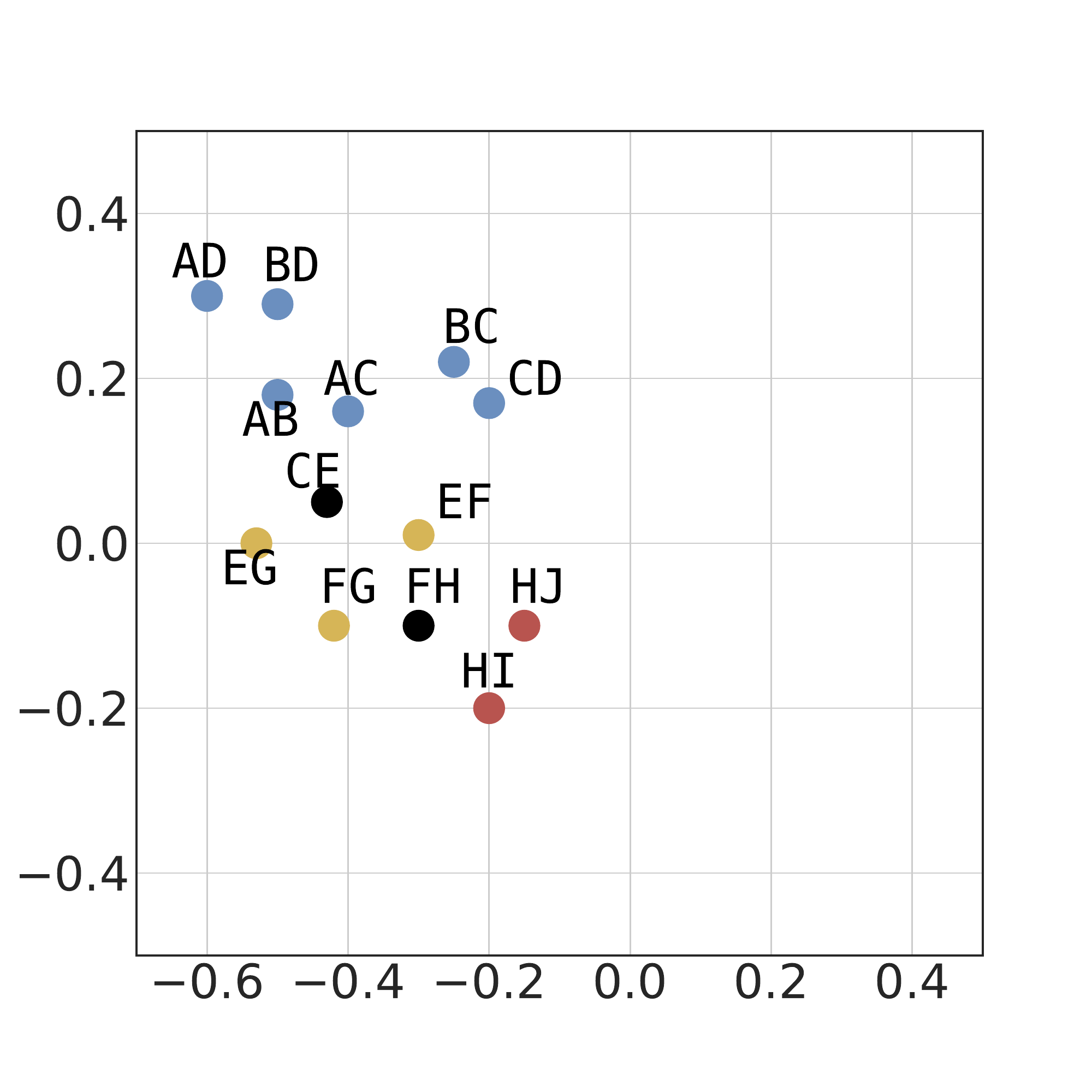}
         \caption{Edge Embedding}
         \label{fig:edgeembedding}
     \end{subfigure}
     \begin{subfigure}[b]{0.35\textwidth}
         \centering
         \includegraphics[width=\linewidth]{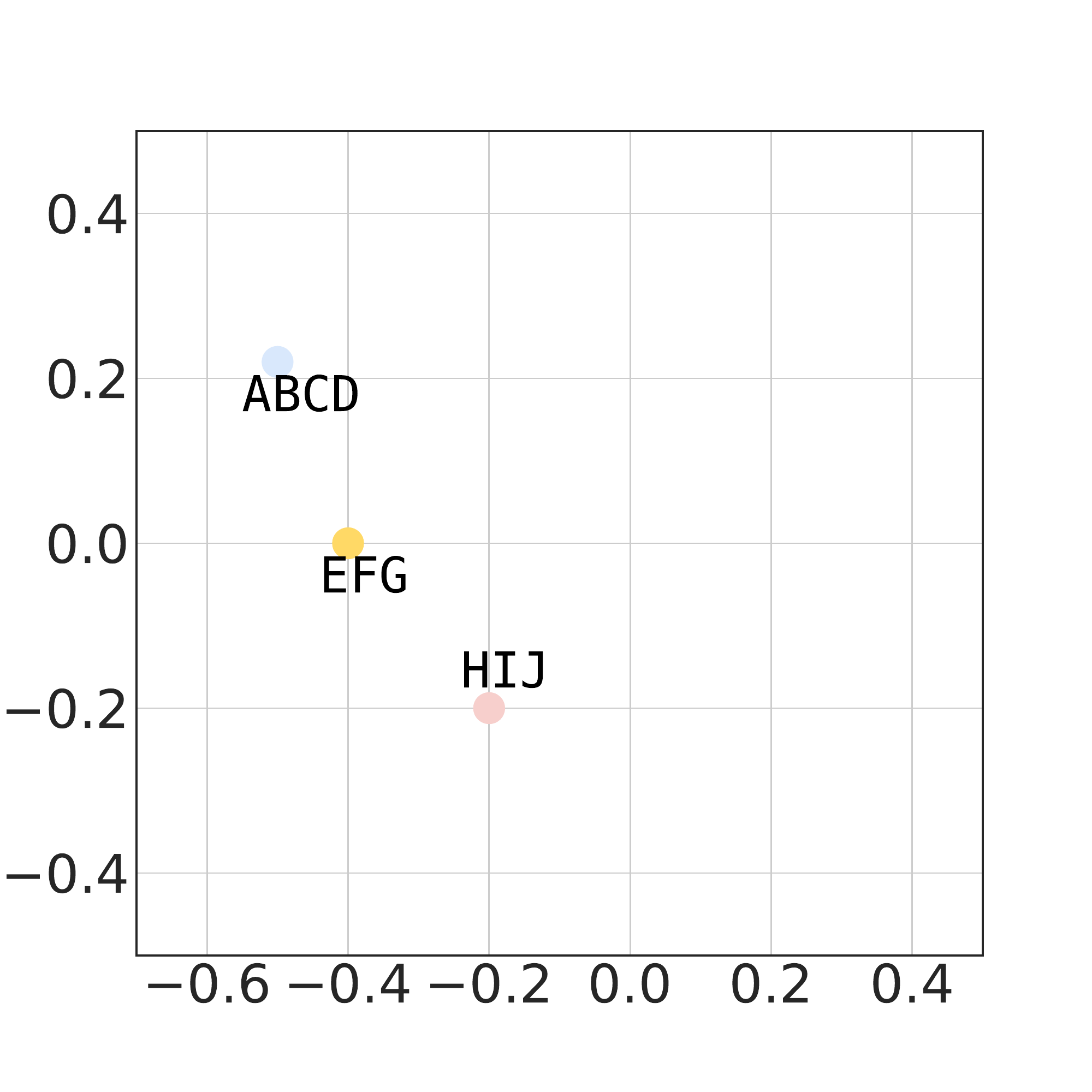}
         \caption{Substructure Embedding}
         \label{fig:substructureembedding}
     \end{subfigure}
     \hfill
     \begin{subfigure}[b]{0.35\textwidth}
         \centering
         \includegraphics[width=\linewidth]{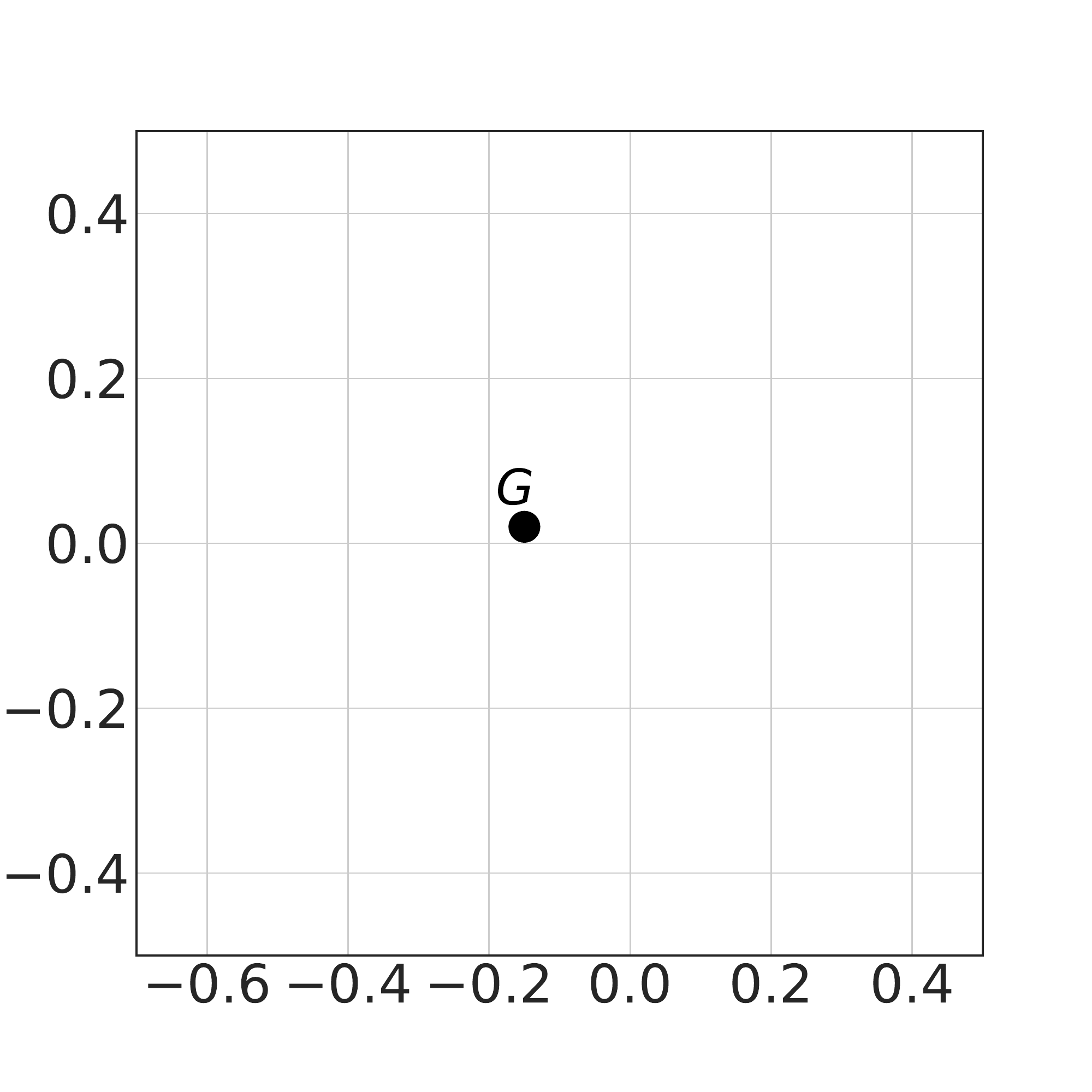}
         \caption{Whole-Graph Embedding}
         \label{fig:wholegraphembedding}
     \end{subfigure}
    \end{minipage}
     \caption{A toy example of embedding a graph into 2D space taking into account different granularities. (a) Sample network used as a reference for graph embedding, where the different node colors depict different substructures, and different edge colors depict intra-substructure and inter-substructure connections. (b-e)~Different static graph embedding outputs, as described by Cai et al.~\cite{Cai2018Comprehensive}, where $d = 2$. Note that the colors refer to the substructures and connections displayed in~(a).}
     \label{fig:staticgraphembedding}
\end{figure}

The central idea behind graph embedding lies in learning a mapping that embeds nodes, edges, subgraphs, or even entire graphs, in a low-dimensional vector space, where the embedding dimension is expected to be much lower than the total number of nodes in the network. More specifically, given a graph $G = (V, E)$, and a predefined embedding dimension $d$, such that $d \ll |V|$, the problem of graph embedding is to map $G$ into a $d$-dimensional space, in which graph properties are preserved as much as possible, i.e. topology and similarity measures~\cite{Hamilton2017Representation, Cai2018Comprehensive, Goyal2018Graph}. Based on the output of the graph embedding, four categories may be defined: (i) node embedding, where vector embeddings are learned for each node; (ii) edge embedding, where edges are mapped into the embedding space; (iii) substructure embedding, in which subgraphs (i.e. clusters, communities, graphlets, ...) are represented in the vector space; and (iv) whole-graph embedding, i.e. an entire graph is mapped into a single vector.~\cite{Cai2018Comprehensive} (see Fig.~\ref{fig:staticgraphembedding}).

Static graphs embedding taxonomies have been proposed in the past few years~\cite{Goyal2018Graph, Cai2018Comprehensive}. Graph embedding based on \textbf{matrix factorization} represents some graph similarity in the form of a matrix and factorizes this matrix to obtain a node embedding. The problem of graph embedding is treated as a structure-preserving dimensionality reduction problem, which assumes the input data lie in a low dimensional manifold. Approaches based on \textbf{deep learning} apply deep neural architectures on graphs, including autoencoders (AEs), convolutional neural networks (CNNs), and variational autoencoders (VAEs). \textbf{Random walk approaches} generate node sequences from a graph to create contexts for each node, then applying techniques from natural language processing for learning embeddings. They try to preserve higher-order proximity between nodes by maximizing the probability of occurrence of subsequent nodes in fixed-length random walks, using neural language models, such as SkipGram. Cai et al.~\cite{Cai2018Comprehensive} further suggest other paradigms, such as \textbf{edge reconstruction based optimization}, which learns representations that directly optimize either edge reconstruction probability or edge reconstruction loss; \textbf{graph kernel-based methods}, which decompose the graph into atomic substructures~(as graphlets and subtrees) and build a vector using these features; and \textbf{generative models}, which specify the joint distribution of the input features and the class labels conditioned on a set of parameters.

Several further discussions about static graph embeddings were presented in other surveys and reviews~\cite{Hamilton2017Representation,Cui2018Survey,Zhang2018Network}, along with some works concerning knowledge graph embedding, specifying their analysis to tasks including knowledge graph completion and relation extraction~\cite{Wang2017Knowledge,Nickel2015Review}. Moreover, Goyal developed GEM~\cite{Goyal2018Graph}, an open-source Python library that provides a framework for graph embedding implementation, and Grattarola and Alippi have presented Spektral~\cite{grattarola2020graph}, another open-source Python library for building graph neural networks with Keras API and TensorFlow 2, handling tasks such as node classification, link prediction, and graph generation.

\subsection{Dynamic Graphs}
\label{subsection:dynamicgraphs}

There are different mathematical formulations to describe interactions between nodes over the lifetime of a system~\cite{Casteigts2012Time, Wehmuth2015Unifying, Latapy2018}. A possible definition is to describe a dynamic graph by a mathematical structure $\mathcal{G} = (\mathcal{V},\mathcal{E},\mathcal{T})$, where $\mathcal{V} = \{V(t)\}_{t \in \mathcal{T}}\,$ is a collection of node sets over time, $\mathcal{E} = \{E(t)\}_{t \in \mathcal{T}}\,$ is a collection of edge sets over time, and $\mathcal{T}$ is the time span. For each $t \in \mathcal{T}$, it is possible to define a graph snapshot $G(t) = (V(t),E(t))$, i.e. a static graph representing a fixed timestamp $t$ of the dynamic graph. Adjacency matrix $A(t)$, weight matrix $W(t)$ and similarity matrix $S(t)$ are now time-dependent, and can be calculated for each snapshot $G(t)$, as well as node types $\mathcal{L}^{n}(t)$ and edge types $\mathcal{L}^{e}(t)$.

Casteigts et al.~\cite{Casteigts2012Time} alternatively defined a dynamic graph as $\mathcal{G} = (V,E,\mathcal{T},\rho_{v},\rho_{e})$, where $V$ is a node set containing every node that is present in the network at any given time $t \in \mathcal{T}$, $E$ is an edge set defined similarly, and further defining a node presence function $\rho_{v}: V \times \mathcal{T} \rightarrow \{0,1\}$, indicating whether a given node $v \in V$ is available at a given time $t \in \mathcal{T}$, and an edge presence function $\rho_{e}: E \times \mathcal{T} \rightarrow \{0,1\}$, specifying if a given edge $e \in E$ exists at a timestamp $t \in \mathcal{T}$.

\begin{figure}[tbh]
    \centering
    \includegraphics[scale=0.45]{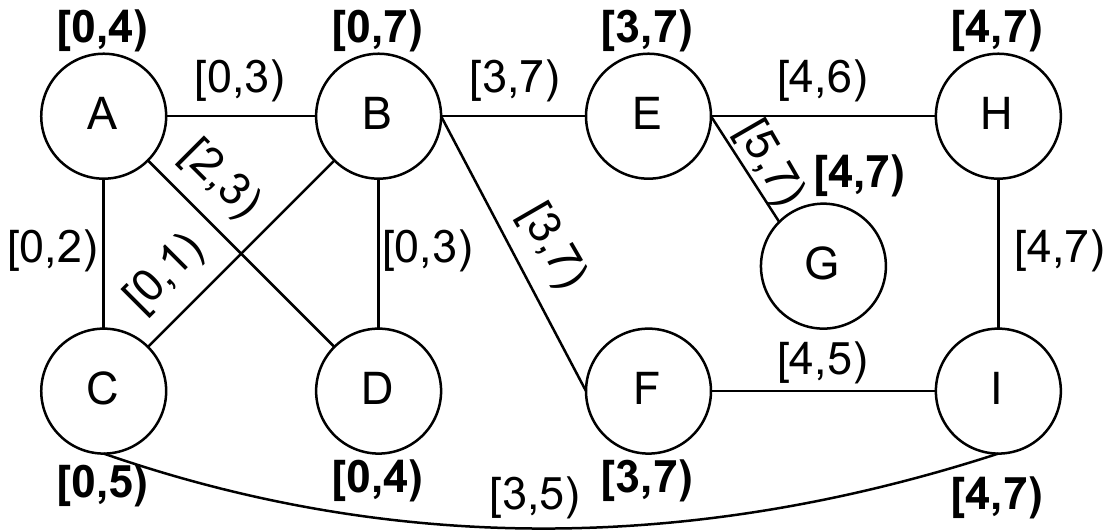}
    \caption{A representation of a small dynamic network, showing edge presence (continuous intervals above edges) and node presence (bold continuous intervals next to nodes) intervals.}
    \label{fig:tvg}
\end{figure}

Figure~\ref{fig:tvg} depicts a dynamic network as a time-varying graph, containing 9 nodes along lifetime $\mathcal{T} = [0,7)$. It is noteworthy that, at the beginning of the network lifetime, only nodes A, B, C, and D are present, as well as links (A,B), (A,C), (B,C), and (B,D). As time passes, new nodes and edges arrive, even as nodes and edges are removed from the system. In the end, we have nodes B, E, F, G, H, and I, and links (B,E), (B,F), (E,G), and (H,I).

The formulations we described above are sufficient for understanding the dynamic graph embedding methods we review in this paper. It is important to mention that dynamic graphs can present even more complex temporal patterns, such as latency~\cite{Casteigts2012Time} (i.e. nodes/edges not arising instantaneously in the network, instead of taking a finite time interval to be established) and spatial-temporal edges~\cite{Wehmuth2015Unifying} (i.e. a node at a given timestamp connected to another node at another timestamp). In Sec.~\ref{sec:conc}, we propose future directions for embedding dynamic graphs concerning these properties.

\subsection{Dynamic Graph Modeling}
\label{subsection:dynamicgraphmodels}

One of the first aspects to be considered when modeling a dynamic network is to define its life span $\mathcal{T}$. Two different approaches may be adopted to model the system's time domain: \textbf{discrete-time approaches}, where $\mathcal{T}$ is a discrete set, hence the evolution of a dynamic graph can be described by a sequence of static graphs, with a fixed timestamp; and \textbf{continuous-time approach}, where $\mathcal{T}$ is a continuous set, therefore the evolution is modeled at a finer temporal granularity to encompass different events in real time~\cite{Trivedi2018Representation}. Computationally, dynamic graph models assuming discrete-time domain are easier to manipulate. Most of the existing embedding methods are based on this approach~\cite{Goyal2018Dyngem}. However, some authors have proposed to model more sophisticated phenomena, such as stochastic events, to leverage applications such as event time prediction (i.e. predict when an edge or a node is created or removed from a network)~\cite{Trivedi2018Representation, dai2016recurrent}. Therefore, these approaches must rely on continuous-time lifespan to capture temporal evolution at an appropriate granularity. We briefly discuss some dynamic graph models covered by embedding methods presented in this survey.

\subsubsection{Graph Snapshots}

This model represents a dynamic graph as a list of static graphs, i.e. $\mathcal{G} = \{{G}(t_{0}), ..., G(t_{N_{S}-1})\}$, where $G(t_{k}) = (V(t_{k}), E(t_{k}))$ is a static graph with timestamp $t_{k}$ ($k \in \{0,...,N_{S}-1\})$, $N_{S}$ is the number of snapshots, $V(t_{k})$ is the node set at timestamp $t_{k}$ and $E(t_{k})$ is the edge set including all edges within the period $[t_{k},t_{k+1})$~\cite{Chen2019Dynamic}. Most of the methods for embedding dynamic graphs manage this model as the input, either by adopting directly a sequence of successive state sub-graphs that represent the network in a discrete way as time passes~\cite{Zhou2018Dynamic,Goyal2018Dyngem,Goyal2019dyngraph2vec}, or splitting the time domain into non-overlapping windows of fixed duration, establishing a static graph for each window~\cite{Ferreira2019Modeling, Mitrovic2019dyn2vec}.

\subsubsection{Difference Network Models}

In many real problems, the number of edges inserted or removed at any given time is much smaller than the total number of edges. i.e. the topological evolution is sparse~\cite{Du2018Dynamic, Mahdavi2019Dynamic}. These models representing these network changes take as input an initial graph $G_{t_{0}}$, and a list of adjacency matrix changes $\Delta \mathcal{A} = \{\Delta A(t_{1}), ..., \Delta A(t_{N_{R}-1})\}$, where $\Delta A(t_{k}) = A(t_{k}) - A(t_{k-1})$, and $N_{R}$ is the total number of recorded timestamps. This definition may be extended for other similarity matrices~\cite{Zhang2018Timers, Li2017Attributed}, and difference networks may be divided into a link formation network (concerning positive values of adjacency matrix change) and a link dissolution network (regarding negative values of adjacency matrix change)~\cite{Hisano2018Semi, Mitrovic2019dyn2vec}. Note that these models do not handle nodes being added or removed from the network, as they rely on matrices with a fixed dimensionality.

\subsubsection{Continuous-Time Network Models}

Continuous-time approaches may include timestamped edges (edges with the information about the time they were created, or the time intervals concerning their existence in the network)~\cite{Nguyen2018Continuous} and link streams (a list of node interactions over time)~\cite{Latapy2018}. Events in the network, such as the creation and removal of nodes and edges, may occur in any time $t \in \mathcal{T}$, and maybe instantaneous (i.e. much faster than the typical temporal granularity of the system) or may be assigned with a latency~\cite{Casteigts2012Time}. Several dynamic graph embedding methods rely on continuous-time networks, either by modeling timestamped edges as stochastic point processes~\cite{dai2016recurrent} or leveraging link streams~\cite{Torricelli2020weg2vec}. Furthermore, node arrival and removal may be included by these network models, as proposed in the literature by stream graphs~\cite{Latapy2018} and appeared in some embedding techniques~\cite{Trivedi2018Representation,Liu2019Real}.

\subsection{Temporal Point Processes on Graphs} \label{subsec:temporalpointprocess}

As discussed above, nodes and edges network churn may be modeled by stochastic events in a continuous-time domain, normally as stochastic point processes, random processes whose realization is comprised of discrete events in a continuous time. A \textbf{temporal point process} is a point process that can be represented as a counting process $N(t)$, recording the number of events up to time $t$, thus being useful for modeling sequential asynchronous discrete events occurring in continuous time~\cite{Kazemi2020Representation}. The \textbf{conditional intensity function} $\lambda(t)$ characterizes a temporal point process such that $\lambda(t)\,\Delta t$ is the conditional probability of observing an event in the tiny window $[t,t+\Delta t)$ given the network history, i.e., all events before $t$, and only one event can happen in this tiny interval $\Delta t$. Similarly, a \textbf{survival function} $S(t)$ determines the conditional probability that no event happens during a time window $[t,t+\Delta t)$ given the network history, and the \textbf{conditional density} $f(t) = \lambda(t) S(t)$ for an event that occurs at time $t$ is further defined as well.

The functional form of the intensity $\lambda(t)$ is often designed to capture the phenomena of interests, some of them include Poisson Process, Hawkes Process, Self-Correcting Process, Power Law, or Rayleigh Process~\cite{Yan2019Modeling}. Many dynamic graph embedding techniques consider that interactions between nodes are stochastic processes whose probabilities depend on the topological structure of the network, and node features (if applicable) at each timestamp~\cite{Trivedi2018Representation, Trivedi2017Know, Zuo2018Embedding, Knyazev2019Learning, Dai2016Deep, Wu2020Modeling}.

\subsection{Dynamic Graph Embedding Input}

\begin{figure}[htp]
    \centering
    \includegraphics[width=0.5\linewidth]{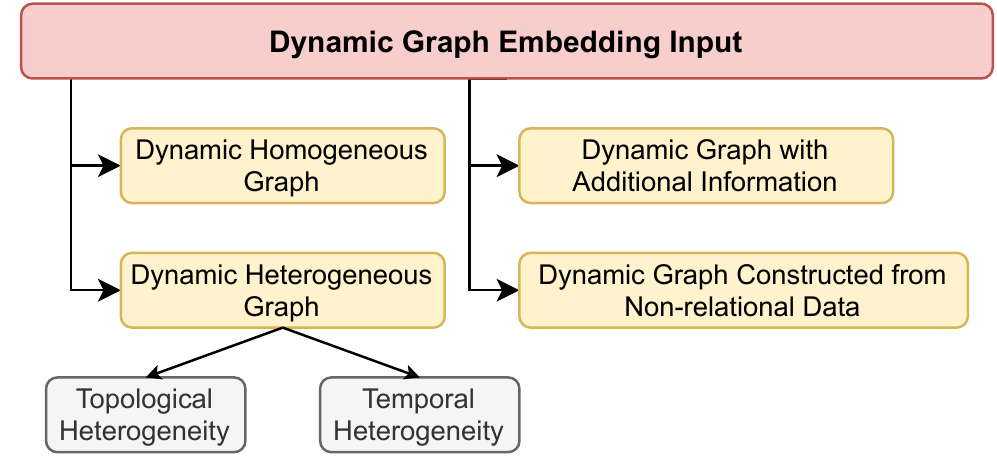}
    \caption{The proposed taxonomy for dynamic graph embedding input, an extension of Cai et al.~\cite{Cai2018Comprehensive} to encompass dynamic networks and to consider topological heterogeneity (similar to static networks) and temporal heterogeneity (i.e. timestamps having labels).}
    \label{fig:input}
\end{figure}

In addition to the dynamic network modeling, discussed in Sec.~\ref{subsection:dynamicgraphmodels}, dynamic graphs can be (i) homogeneous, in which only topological information over time is available, (ii) heterogeneous, in which either nodes, edges (topological heterogeneity) or timestamps (temporal heterogeneity) are assigned with labels, (iii) attributed (or with additional information), where nodes and edges may hold several different features, and (iv) constructed from non-relational data (see Fig.~\ref{fig:input}). This proposed taxonomy extends Cai et al.~\cite{Cai2018Comprehensive}, who encompassed static graph embedding input considering static networks, without leveraging dynamic aspects neither handling the difference between topological and temporal heterogeneity. In the following, we discuss each dynamic graph embedding input shown in Figure~\ref{fig:input}.

\subsubsection{Dynamic Homogeneous Graph}

Undirected and unweighted homogeneous graphs are widely used as dynamic graph embedding inputs due to their simplicity and to handle only basic structural information over time~\cite{Saha2018Models}. Several embedding methods, however, are proposed to handle weighted~\cite{Lei2019GCNGAN, Wu2019T} and directed dynamic graphs~\cite{Goyal2018Dyngem}.

\subsubsection{Dynamic Heterogeneous Graph}

Embedding methods to handle topological heterogeneity, i.e. nodes or edges having labels, are usually concerned with node and edge classification~\cite{Yin2019DHNE, Pareja2020EvolveGCN}. Nevertheless, graph snapshots may have labels, characterizing different global behavior~\cite{Taheri2019Learning, Taheri2019Predictive} and describing another type of heterogeneity, which we have named \textbf{temporal heterogeneity}.

\subsubsection{Dynamic Graph with Additional Information}

Additional attributes may be assigned to nodes, such as a set of numerical or categorical features. It is possible to define a time-dependent node feature matrix $F(t) \in \mathbb{R}^{N \times f}$, where $f$ is the number of additional node features, and learn representations leveraging these features in addition to their topological structure~\cite{Li2017Attributed, Wei2019Lifelong}. Although an edge feature matrix would be defined as well, its usage is much less common.

\subsubsection{Dynamic Graph Constructed from Non-Relational Data}

Non-relational time series data can be transformed into a dynamic graph by defining a similarity measure between two data instances, and constructing a similarity matrix $S(t)$ afterward. Several papers use this step as an intermediate to learn vector representations from this constructed graph to support some task-driven application, such as traffic forecasting~\cite{Yu2017Spatio}, predicting bike-sharing demand~\cite{Li2019Learning}, predicting social events~\cite{Deng2019Learning} and missing label classification on videos~\cite{Yuan2017Temporal}. 

\subsection{Problem Formulation and Output for Dynamic Graph Embedding}
\label{subsection:dynamic_graph_embedding}

It is important to mathematically formulate dynamic graph embedding to understand its outputs. Given a dynamic graph $\mathcal{G} = (\mathcal{V}, \mathcal{E}, \mathcal{T})$, where $\mathcal{V} = \{V(t)\}_{t \in \mathcal{T}}$ and $\mathcal{E} = \{E(t)\}_{t \in \mathcal{T}}$, and an embedding dimension $d$, the problem of embedding dynamic graph is regarded as learning how to map $\mathcal{G}$ into a $d$-dimensional vector space over time, in which both topological information and temporal dependencies of the network are captured, either by learning representations able to reconstruct the dynamic graph $\mathcal{G}$, to predict the behavior of the network at timestamps outside the lifespan $\mathcal{T}$, or to directly handle a task-driven application such as node classification. When the graph topology evolves, two possible interpretations are possible for the evolution of embeddings: (i)~the vector representations move along the embedding space, making it possible to trace the trajectory of each node; or (ii)~the embedding space itself evolves in time, thus being possible to learn mappings between embedding spaces in consecutive timestamps~\cite{Saha2018Models}.

The time-domain of vector representations and the network does not need to be identical, i.e., $\mathbb{T} \ne \mathcal{T}$. For instance, a dynamic graph may have daily information about interactions between users in a social network, but the network analytics and inference are more interested in capturing weekly or even monthly features. Hence, even though the network life span is given by daily timestamps, vector representations are extracted for a coarser temporal granularity.

\begin{figure}[htp]
    \centering
    \includegraphics[width=0.45\linewidth]{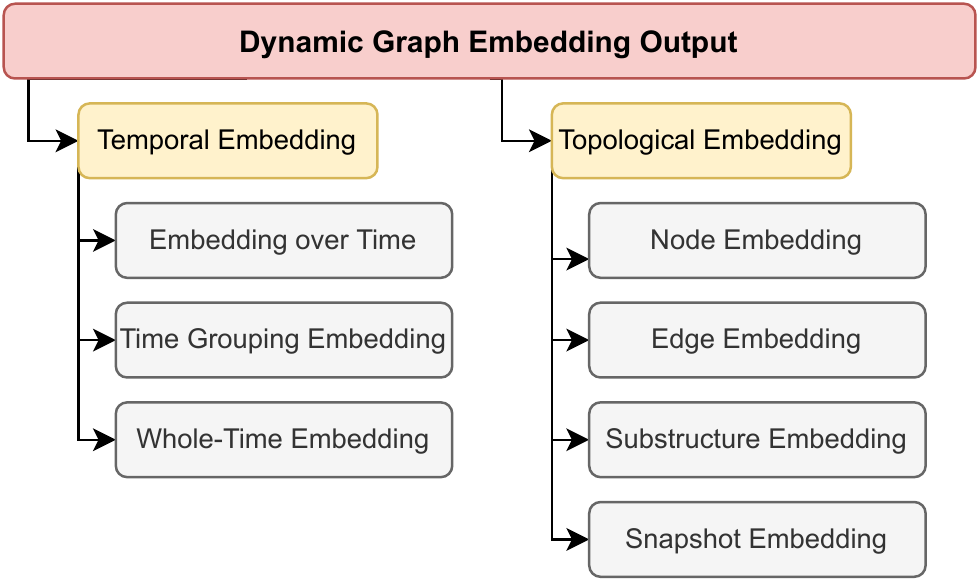}
    \caption{Dynamic graph embedding output taxonomy proposed in this survey.}
    \label{fig:output}
\end{figure}

Therefore, to define different dynamic graph embedding outputs, it is important to separate between (i) \textbf{topological embedding}, which is similar to the definitions for static graph embedding~\cite{Cai2018Comprehensive} and concerns node embedding, edge embedding, substructure embedding, and graph snapshot embedding, all of them over time, and (ii) \textbf{temporal embedding}, regarding the relation between network temporal granularity given by $\mathcal{T}$ and embedding temporal granularity given by $\mathbb{T}$. The complete classification we propose is shown in Figure~\ref{fig:output} and is further discussed in the following topics.

\subsubsection{Temporal Embedding}

\begin{figure}[htp]
     \centering
     \begin{subfigure}[b]{0.6\textwidth}
         \centering
         \includegraphics[width=\textwidth]{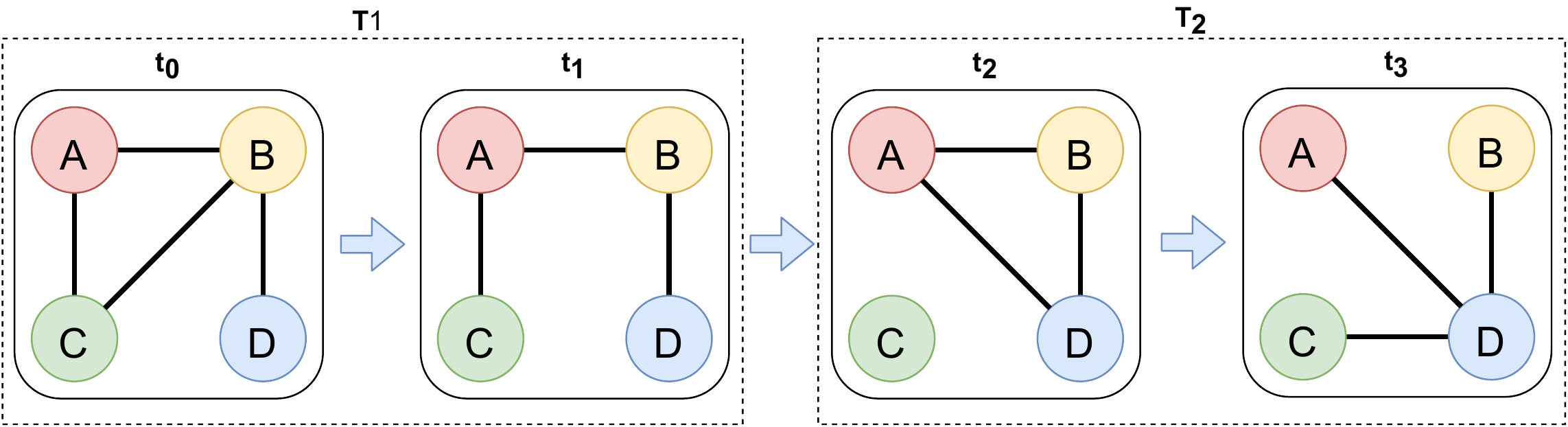}
         \caption{Dynamic graph $\mathcal{G}$}
         \label{fig:temporalembedding1}
     \end{subfigure}
     \hfill\\
     \begin{subfigure}[t]{0.25\textwidth}
        \centering
         \includegraphics[width=\textwidth]{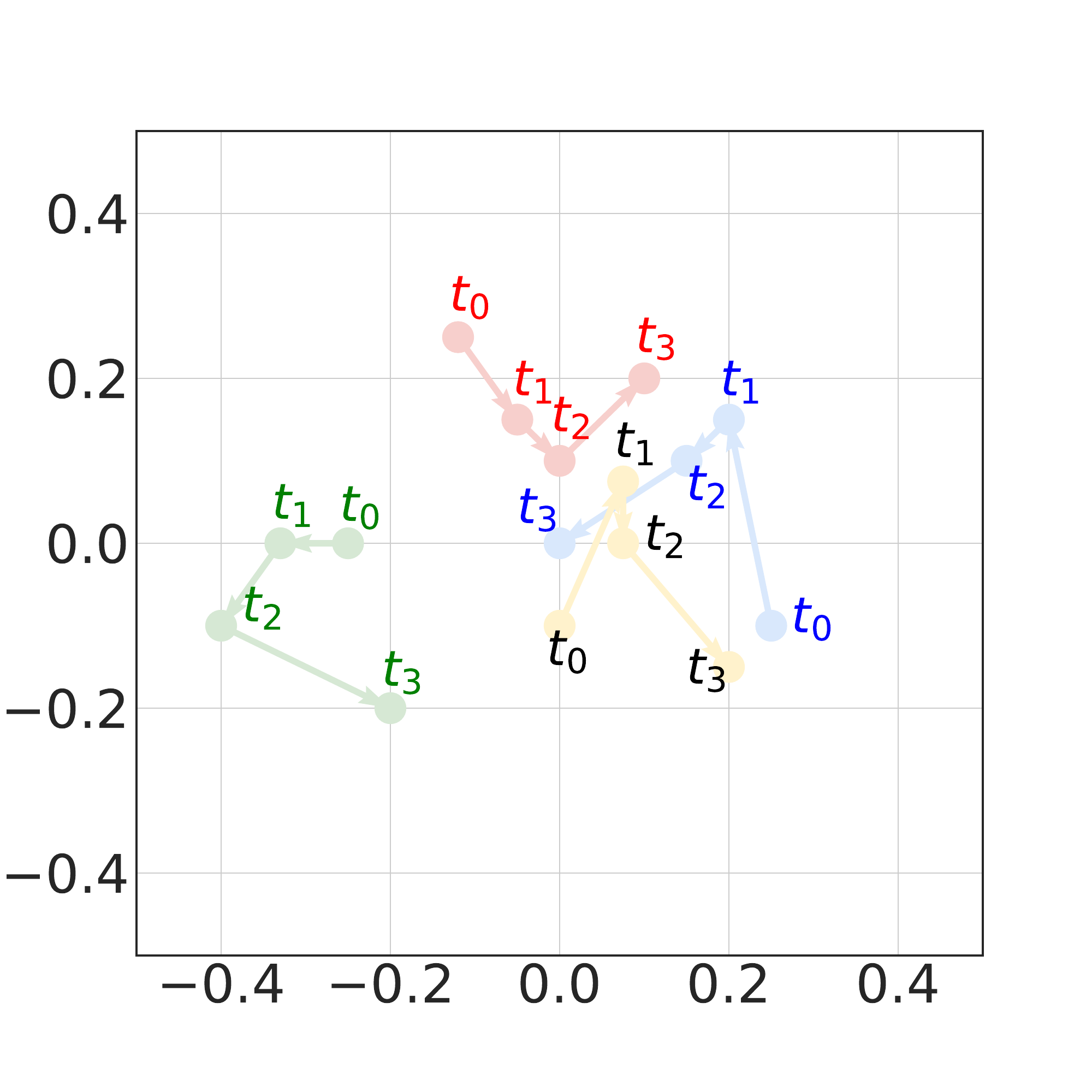}
         \caption{Embedding dynamic graph}
         \label{fig:embeddingovertime}
     \end{subfigure}
     \begin{subfigure}[t]{0.25\textwidth}
        \centering
         \includegraphics[width=\textwidth]{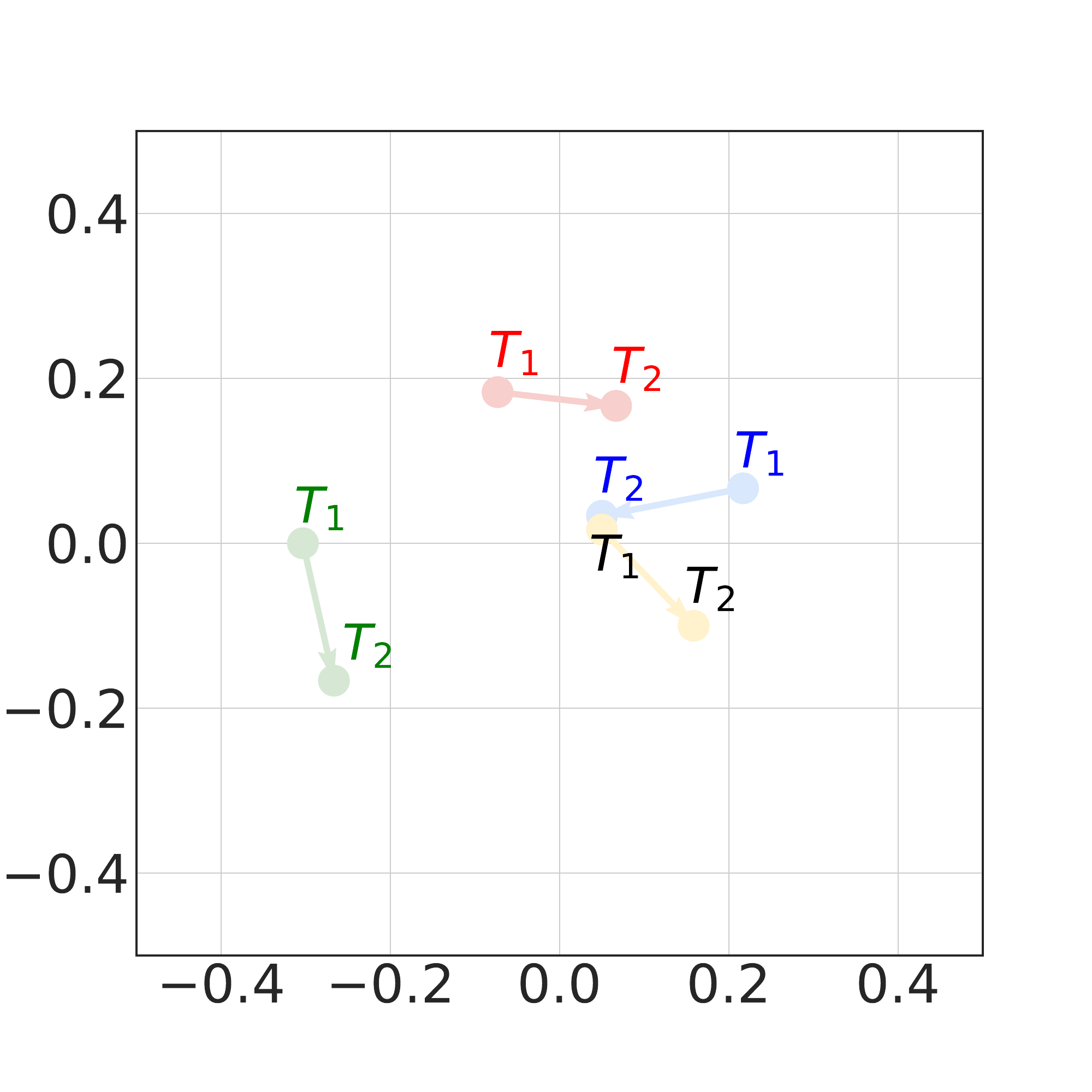}
         \caption{Time-grouping embedding}
         \label{fig:timegroupingembedding}
     \end{subfigure}
     \begin{subfigure}[t]{0.25\textwidth}
        \centering
         \includegraphics[width=\textwidth]{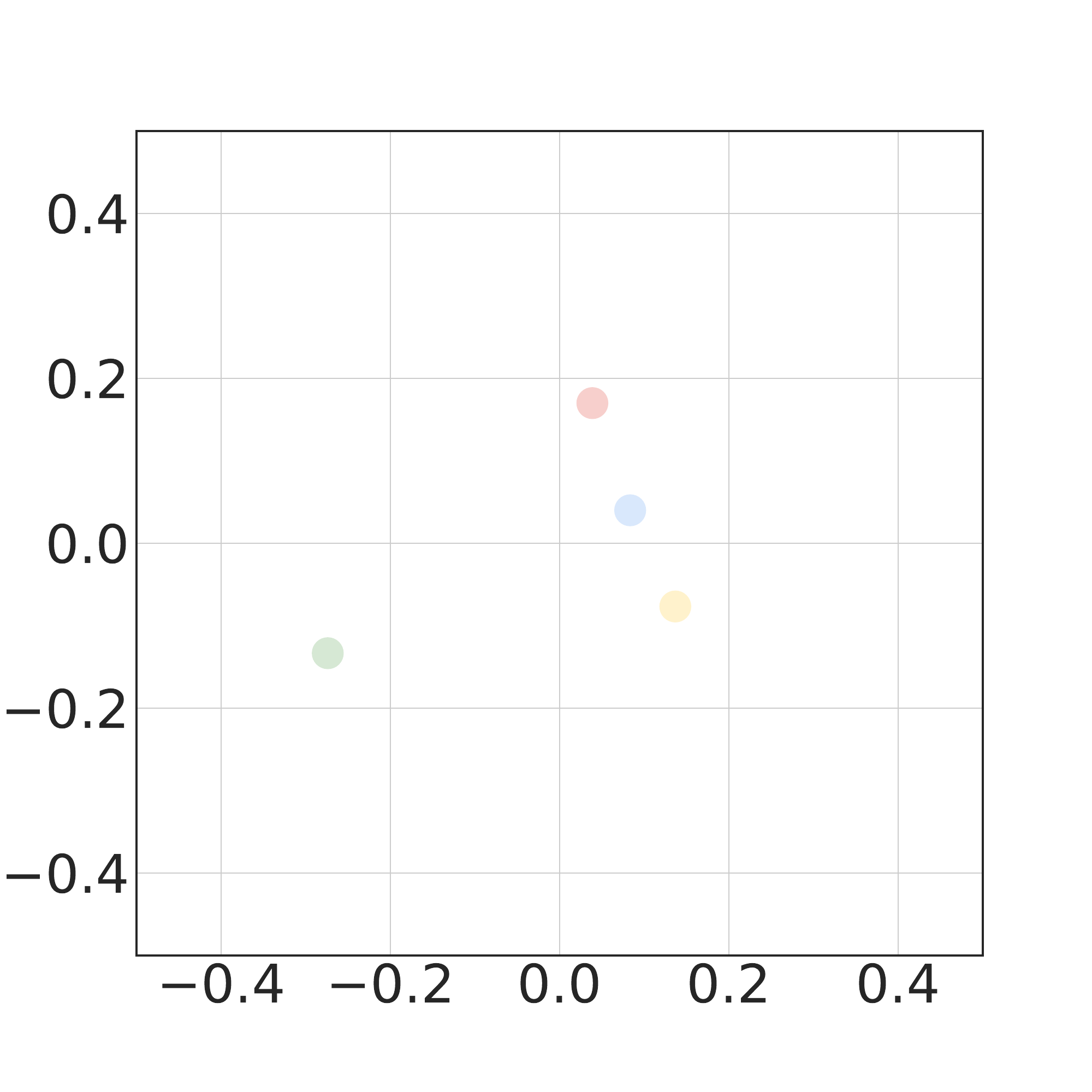}
         \caption{Whole-time embedding}
         \label{fig:wholetimeembeddig}
     \end{subfigure}
     \hfill
        \caption{A toy example of embedding each node of a dynamic graph into 2D space taking into account different temporal granularities. (a) Sample network used as a reference for dynamic node embedding, where the snapshot model is considered, different node colors are used to distinguish each of them, and the four timestamps $t_{0}$ to $t_{3}$ are grouped into two time aggregates $T_{1}$ (holding $t_{0}$ and $t_{1}$) and $T_{2}$ (which holds $t_{2}$ and $t_{3}$). (b) In embedding over time, each node is mapped into the low-dimensional space for each timestamp, thus describing trajectories in embedding space. These trajectories are illustrated by arrows, indicating the time flow. In this case, $\mathcal{T} = \mathbb{T}_{a} = \{t_{0},t_{1},t_{2},t_{3}\}$. (c) Although time granularity of the network is described by $\mathcal{T}$, the graph is mapped into a different time granularity $\mathbb{T}_{b} = \{T_{1},T_{2}\}$, where each new timestamp aggregates temporal information about the original timestamps. The representations were defined from embedding over time handcrafted design, and   (d) In the whole-time aggregation embedding, $\mathbb{T}_{c}$ is a unitary set, and every timestamp is aggregated into a single representation for each node.}
        \label{fig:temporalembedding}
\end{figure}

Temporal embedding concerns the relationship between the \textbf{input temporal domain} $\mathcal{T}$ and the \textbf{output temporal domain} $\mathbb{T}$. Defining $T(\mathcal{G})$ as a set compressing a topological property of the dynamic graph (i.e. node set, edge set, induced subgraph set or entire graph, as described later in Section~\ref{subsubsec:topological}), one may distinguish between three different classifications for temporal embedding, as shown in Figure~\ref{fig:output} and further illustrated in Figure~\ref{fig:temporalembedding}.

\begin{definition}
An \textbf{embedding over time} is a mapping $\mu_{t}: T(\mathcal{G}) \times \mathcal{T} \to \mathbb{R}^{d} \times \mathcal{T}$ (i.e. $\mathbb{T} = \mathcal{T}$). Therefore, each node/edge/substructure/graph at time $t \in \mathcal{T}$ is represented as a vector in a low dimensional space over $t$.
\end{definition}

In this type of temporal embedding, the mapping $\mu_{t}$ is a bijection in time, allowing to distinguish, in the vector space, each time instant for each mapped entity of the network~(Figure~\ref{fig:embeddingovertime} shows an example).

\begin{definition}
A \textbf{time-grouping embedding} is a mapping $\mu_{g}: T(\mathcal{G}) \times \mathcal{T} \to \mathbb{R}^{d} \times \mathbb{T}$, where $\mathbb{T}$ is a discrete set composed of elements that aggregate timestamps or time intervals of $\mathcal{T}$. Therefore, instead of representing each node/edge/substructure/graph at every time $t \in \mathcal{T}$, they are represented at every time aggregate $t' \in \mathbb{T}$.
\end{definition}

Time aggregates are useful for representing networks whose desired time information is of a different granularity than that available in the data. As discussed, vector representations may be required each week from data obtained daily. Figure~\ref{fig:timegroupingembedding} contains another example, where two timestamps have been aggregated into a single one.

\begin{definition}
A \textbf{whole-time embedding} is a mapping $\mu_{w}: T(\mathcal{G}) \times \mathcal{T} \to \mathbb{R}^{d}$, i.e. $\mathbb{T}$ is a unitary set indicating that every timestamp is aggregated into a single point. Therefore, each node/edge/substructure/graph at every time $t \in \mathcal{T}$ is represented as a single vector in $\mathbb{R}^{d}$.
\end{definition}

Temporal aggregation can be performed: (i)~as a step before embedding, aggregating temporal data before applying an embedding method; or (ii)~after representation learning over the dynamic graph, usually by performing operations in the vector space (e.g. weighted averages or non-linear functions) to obtain lower temporal granularity representations (i.e. embedding daily network data to extract weekly or even monthly network representations).

\subsubsection{Topological Embedding}
\label{subsubsec:topological}

Embedding topological properties of a dynamic graph is similar to Fig.~\ref{fig:staticgraphembedding}, and the main difference is the coupling between the temporal embedding discussed above and each graph structure, as presented in the right branch of Fig.~\ref{fig:output}. WLoG, we are considering $\mathcal{V} = V \times \mathcal{T}$ and $\mathcal{E} = E \times \mathcal{T}$ to simplify notations.

\begin{definition}
A \textbf{dynamic node embedding} is a mapping $\nu_{n}: V \times \mathcal{T} \to \mathbb{R}^{d} \times \mathbb{T}$. Therefore, each node at time $t \in \mathbb{T}$ is represented as a vector in a low dimensional space.
\end{definition}

Node embedding over time is useful for several applications such as time-dependent node classification, network clustering evolution, and link prediction. It also allows one to track node trajectories in the embedding space and extract information about node behavior and roles in the network.~\cite{Rossi2013Modeling, Ferreira2019Modeling}.

\begin{definition}
A \textbf{dynamic edge embedding} is a mapping $\nu_{e}: E \times \mathcal{T} \to \mathbb{R}^{d} \times \mathbb{T}$. Therefore, each edge at time $t \in \mathbb{T}$ is represented as a vector in a low dimensional space.
\end{definition}

In addition to edge embedding, a few methods map both edges and nodes, in particular for embedding dynamic knowledge graphs or user-item interaction graphs~\cite{Dai2016Deep,Goel2019Diachronic}.

\begin{definition}
A \textbf{dynamic substructure embedding} is a mapping $\nu_{h}: S(\mathcal{G}) \times \mathcal{T} \to \mathbb{R}^{d} \times \mathbb{T}$, where $S(\mathcal{G})$ is a set of induced subgraphs in $G$ at each time $t \in \mathcal{T}$. Therefore, each substructure at time $t \in \mathcal{T}$ is represented as a vector in a low dimensional space.
\end{definition}

Several dynamic graph embedding methods generalize traditional node or link prediction tasks to consider joint prediction over larger $k$-node induced subgraphs~\cite{Meng2018Subgraph} and graphlets~\cite{Rahman2016Link, Dave2019Triangle}.

\begin{definition}
A \textbf{snapshot embedding} is a mapping $\nu_{\mathcal{G}}: \mathcal{G} \times \mathcal{T} \to \mathbb{R}^{d} \times \mathbb{T}$, where $\mathcal{G} = \{G_{t}\}_{t \in \mathcal{T}}$. Hence, each graph snapshot at time $t \in \mathcal{T}$ is represented as a vector in low dimensional space.
\end{definition}

Snapshot embeddings are useful for tracking network behavior over time, when the topological structure of the network is related to some emerging property or global interpretation of node interactions~\cite{Taheri2019Predictive, Taheri2019Learning}.

\subsection{Dynamic Behaviors}

Embedding methods can also be identified according to the type of time dynamics they capture. Most methods can capture the evolution of connectivity between network nodes, i.e. addition and removal of edges. However, several works generalize the method to include adding and removing nodes in the network. In addition, there are methods capable of capturing other temporal properties of networks, including varying edge weights, changing node and edge classification, evolving node, and edge attributes, and dynamic processes in the network~(such as diffusion cascades).

\begin{figure}[htp]
    \centering
    \includegraphics[width=0.47\linewidth]{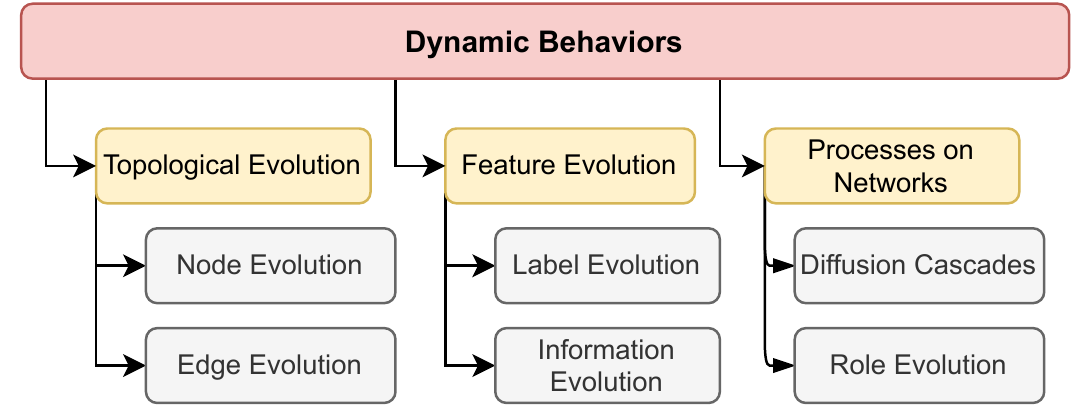}
    \caption{Several dynamical properties that may be captured by embedding methods, divided by topological evolution (concerning changes in the node and edge set), feature evolution (related to node or edge label or other additional information changes over time) and processes on the network (regarding diffusion and role evolution of nodes in a network).}
    \label{fig:dyncharacteristics}
\end{figure}

We mapped three groups of dynamical behaviors on networks: \textbf{topological evolution} (nodes and edges varying over time), \textbf{feature evolution} (node and edge features changing over time), and \textbf{processes on networks} (a time-dependent process taking place on the network), as shown in  Figure~\ref{fig:dyncharacteristics}.

\subsubsection{Topological Evolution}

Topological evolution means that nodes and edges may vary over time, being added/removed from the network. \textbf{Node evolution} is characterized by changes on the node set $V$, whereas the \textbf{Edge evolution} concerns connections between nodes that may be continuously formed or broken. Several dynamic graph embedding methods require that the number of nodes in a network does not change over time, since they handle adjacency or similarity matrices with fixed dimensionality~\cite{Saha2018Models}. Moreover, approaches considering edge creation may be captured concerning a trend through historical records of interaction between a pair of nodes (i.e. if two nodes have recently made multiple connections, they are more likely to make future connections), and ternary closure (if a third node is a common neighbor of two unconnected nodes, there is a greater tendency to close the triangle and form a clique)~\cite{Meng2018Subgraph}.

\subsubsection{Feature Evolution}

In many problems involving heterogeneous graphs, it is assumed nodes have fixed labels. However, it is also possible to observe a \textbf{label evolution}: In a citation network, an author may have as its main research area a different topic compared to previous years. Such changes are linked in some way to the topological evolution of the network. In node classification tasks, each node in a graph has a class label, hence it is possible to predict the class label for the nodes in a graph $G(t_{k})$ using previous graphs $G(t_{0}),...,G(t_{k-1})$~\cite{Mahdavi2019Dynamic}. 

Even more, in several real-world networks, nodes and edges may have rich attributes (i.e. additional information) that are changing over time in addition to the network structure and the topology may influence attribute modification. Therefore, embedding methods may also capture \textbf{information evolution}~\cite{Wei2019Lifelong, Goel2019Diachronic}. Furthermore, edge weights may also change over time in weighted networks, and their changes may be handled by embedding methods~\cite{Yu2017Temporally}.

\subsubsection{Processes on Networks}

It is possible to analyze dynamic processes on the network, such as information diffusion or disease spreading, and other emergent properties, such as node roles changing over time. Two of the most basic and widely-studied diffusion models include~\cite{kempe2003maximizing}: (i)~linear threshold model, where each node $v_{i}$ is influenced by each neighbor $v_{j} \in \mathcal{N}_{v_{i}}$ according to a sum of weights $b_{ij}$, and if this sum is higher than a random choice of a threshold $\theta_{v_{i}}$ at time $t_{k}$, the node $v_{j}$ is activated at time $t_{k+1}$; and (ii)~independent cascade model, where an active node $v_{i}$ influences its neighborhood $\mathcal{N}_{v_{i}}$ with a probability parameter $p_{ij}$ (for each node $v_{j} \in \mathcal{N}_{v_{i}})$ independently of the network history. In this context, the problem of modeling diffusion by independent cascades comes down to learning probability distributions characterizing the hidden influence between users, to discover the main communication channels of the network. A group of nodes sharing similar roles in the network can be regarded as a set of nodes that are more structurally similar to nodes inside the set than outside, whereas communities are sets of nodes with more connections inside the set than outside. Hence the dynamic role evolution aims to automatically discover groups of nodes (representing common patterns of behavior) based on their latent features given by its representations~\cite{Rossi2013Modeling}.

\subsection{Stability and Temporal Smoothness}
\label{subsection:stability}

A successful dynamic graph embedding algorithm should create stable embeddings over time. In other words, an embedding should be able to learn similar representations at consecutive timestamps if the underlying graph changes only slightly. More specifically, given the dynamic graph $\mathcal{G}$, if the graph snapshot $G(t_{k+1})$ is similar to $G(t_{k})$ (for instance, adjacency matrices $A(t_{k+1})$ and $A(t_{k})$), the embedding matrix $Z(t_{k+1})$ is expected to be similar to $Z(t_{k})$. Goyal et al.~\cite{Goyal2018Dyngem} propose the stability constant $K_{A}(\nu)$ as a metric to evaluate the stability of a dynamic graph embedding function $\nu$ in terms of the adjacency matrix $A$ over time. More specifically, the authors consider the stability of any embedding at a given timestamp as the ratio of the Frobenius norm of the difference between embedding matrices and the Frobenius norm of the difference between adjacency matrices, at consecutive timestamps. Then, the stability constant is the maximum difference between stabilities calculated along the entire life span $\mathcal{T}$ of the network, and the authors claim that a dynamic embedding $\nu$ is stable as long as the stability constant is small. It is possible to further extend these proposals to include any similarity matrix $S$, and to take the limit of two consecutive timestamps $t_{k+1} - t_{k} = \Delta t \to 0$ in order to analyze continuity aspects of embeddings (see future directions in Sec.~\ref{sec:conc}).

Although the above discussion is related to the global behavior of embedding, most of the proposed embedding methods seek local stability, i.e., for each node in the network. If the local topological structure around a node $v$ has undergone few changes, it is expected that the representations $z_{v}(t_{k+1})$ and $z_{v}(t_{k})$ are similar, assuming a \textbf{temporal smoothness} in the embedding space~\cite{Mahdavi2019Dynamic, Zhou2019Dynamic, Sankar2018Dynamic}.

\section{Techniques for the Embedding of Dynamic Graphs}
\label{sec:tech}

\begin{figure}[htp]
    \centering
    \includegraphics[width=0.7\linewidth]{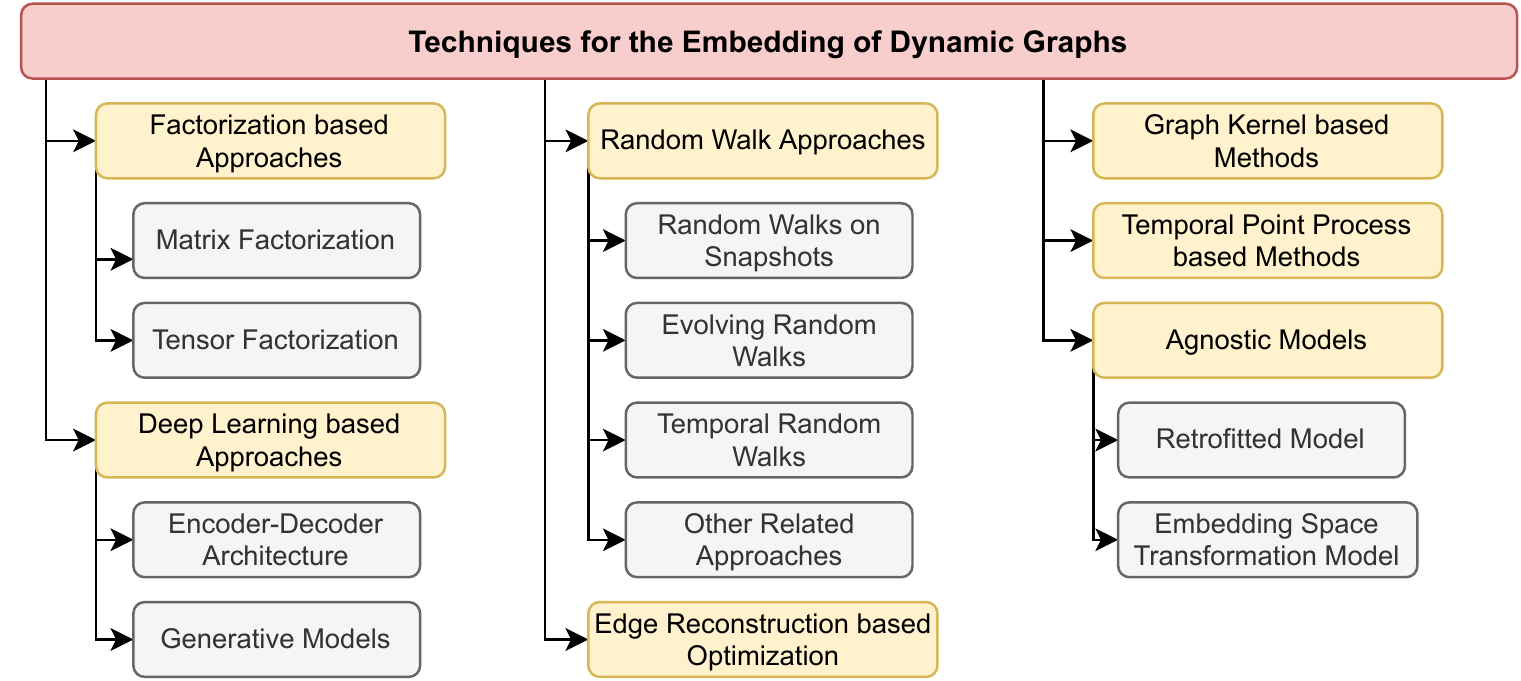}
    \caption{Proposed taxonomy for dynamic graph embedding techniques, organized by algorithmic approaches.}
    \label{fig:techniques}
\end{figure}

In this section, we propose a taxonomy of dynamic graph embedding techniques, summarized in Figure~\ref{fig:techniques}. This taxonomy has been inspired by previous static graph embedding classifications~\cite{Goyal2018Graph,Cai2018Comprehensive}. Classifications for dynamic graphs can be seen as extensions of static graph methods: (i)~\textbf{matrix factorization approaches}; (ii)~\textbf{deep learning approaches}; (iii)~\textbf{random walk-based methods} (which may be understood as node sequence sampling methods); (iv)~\textbf{optimization based on edge reconstruction}, which also leverages \textbf{temporal smoothness}, and (v) \textbf{graph kernel methods}. Here we include \textbf{tensor factorization approaches}, fusing them with matrix factorization approaches and defining general \textbf{factorization based approaches}, and introduce two novel paradigms: (i)~\textbf{temporal point process based} methods, which handles similarity matrix changes as stochastic processes; and (ii)~\textbf{agnostic models}, which learn embeddings over time independent of the approach used for each graph snapshot in dynamic networks.

Earlier approaches to map dynamic networks into vector spaces proposed learning independent vector representations of each snapshot employing static graph embedding methods. However, since representation learning assumes that the probability mass of the data concentrates in manifolds that have much smaller dimensionality than the original space where the data lives, the evolution of the network may cause two relevant impacts on the embedding space: (i)~the embedding vectors move on the manifold; and (ii)~the manifold itself evolves in time~\cite{Saha2018Models}. Therefore, integrating spatial topology of network and the temporal network evolution into feature vectors encompassing these temporal correlations enhances the performance of prediction, classification and many other temporal network analysis problems~\cite{Yu2017Link}.

Next, we present each classification proposed by our taxonomy, discussing its insights, how it leverages temporal dependence in addition to topological structure, and describing several methods following each approach.Concrete examples of applications using the methods covered here are presented only in the following section.

\subsection{Factorization-based methods}

Factorization-based approaches generate node embeddings over time by finding low-rank decompositions of time-dependent similarity measures. These similarity measures can be represented by: (i)~sequence of matrices over time; or (ii)~three-way tensors, where the first two dimensions are related to the similarity between nodes, and the third dimension is the temporal slice. The matrix representation relies on capturing the temporal evolution between representations in adjacent timestamps, and the embedding learns how the network changes over time, separating the topological contribution from the temporal contribution. The tensor representation, on the other hand, couples topology and temporal evolution in a structure to be factorized in a unified way. Consequently, it is valid to separate the factorization approaches into \textbf{matrix factorization approaches} and \textbf{tensor factorization approaches}.

\subsubsection{Matrix factorization approaches} \label{subsub:matrixfactorization}

\begin{figure}[htp]
    \centering
    \includegraphics[width=0.45\linewidth]{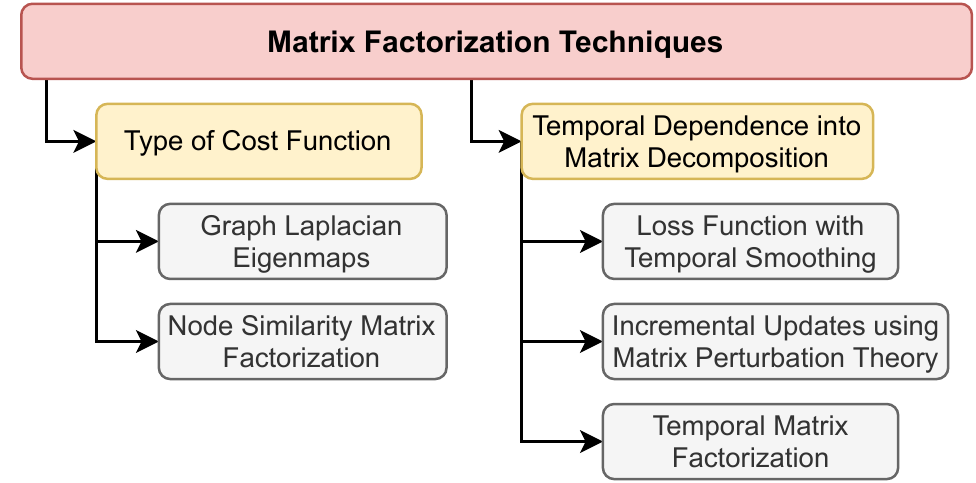}
    \caption{Classification of matrix factorization techniques for dynamic graph embedding, taking into account the type of cost function (similar to Cai et al.~\cite{Cai2018Comprehensive}) and how temporal dependence is leveraged into matrix decomposition.}
    \label{fig:matrixfactorization}
\end{figure}

Matrix factorization approaches express the evolutionary structure of networks in the form of matrices, therefore leveraging the time-dependent structural correlation among pairs of nodes. Dynamic graph embedding based on matrix factorization may be classified, as in static graphs, according to which type of loss function the approach minimizes. While most approaches factorize the similarity matrix and define an inner-product function between node embeddings to approximate the proximity measure~\cite{Rossi2013Modeling, Yu2017Temporally, Ferreira2019Modeling, Zhang2018Timers, Zhu2016Scalable}, graph Laplacian Eigenmaps can be used to reconstruct time-dependent adjacency matrix from an eigendecomposition of Laplacian matrix~\cite{Li2017Attributed}. The novelty in the embedding of dynamic networks consists in the way of propagating the factorization over time, maintaining the stability of the representations while inserting temporal dependence into matrix decomposition. It is possible to distinguish three paradigms based on the modeling for the connections between different timestamps: (i)~adding temporal smoothing directly to the loss function, thus ensuring the stability of embeddings over time and focusing on node trajectories; (ii)~updating the similarity matrix by using matrix perturbation theory, assuming that temporal evolution changes network topology slightly; and (iii)~defining a temporal matrix factorization, decomposing the similarity matrix into a constant term and a time-dependent term. Figure~\ref{fig:matrixfactorization} shows these two classifications.

\begin{compactitem}
    \item \textbf{Jointly optimizing loss function and temporal smoothing:} The methods based on the above insight perform the matrix factorization of each snapshot, in addition to bound the representations through a loss function term that is minimized when, for each node, the embeddings at two consecutive timestamps are similar. In other words, these approaches assume temporal smoothness in the embedding space, and the optimal embedding should jointly reconstruct the similarity matrix over time while being continuous. Given the loss function $\mathcal{L}_{0}(t)$ minimized by a matrix factorization approach on static graphs at each timestamp $t \in \mathcal{T}$, the jointly similarity reconstruction and temporal smoothing loss function for a dynamic graph can be defined as:
    \begin{equation}
        \mathcal{L} = \sum_{t \in \mathcal{T}} \mathcal{L}_{0}(t) + \tau \sum_{t \in \mathcal{T}}\sum_{t' \in \mathcal{T} | (t' - t) \le \Delta t} \mathcal{L}_{\Delta t} (t', t),
    \end{equation}
    where $\Delta t$ is the time interval between consecutive timestamps in $\mathcal{T}$, $\mathcal{L}_{\Delta t}(t', t)$ is a loss function for each pair of consecutive timestamps ($t$ and $t'$) implying temporal smoothness. $\tau > 0$ regulates the temporal smoothness term contribution. Some approaches following this paradigm include Ferreira et al.~\cite{Ferreira2019Modeling} and Zhu et al.~\cite{Zhu2016Scalable}.
    
    One can implement two generalizations of the loss function: (i) the contribution of each timestamp may not be homogeneous, since the future in general depends more directly on the recent past than on the distant past, weighing the loss function $\mathcal{L}_{0}$ with a function $f(t)$ that is close to $1$ when $t \approx \tau$ (where $\tau$ is the last timestamp, i.e. $t_{N_{S}-1}$ for snapshot model), and close to $0$ if $t \ll \tau$~(as an example, $f(t) = e^{\tau - t}$); and (ii) the dataset may contain non-homogeneous time intervals, i.e., a function $g(\Delta t)$ may be defined in order to leverage different values of $\Delta t$, and the smoothness of embeddings at timestamps $t$ and $t + \Delta t$ may be less important when $\Delta t$ is large.

    \item \textbf{Incremental updates on embeddings:} In these methods, the initial graph snapshot $G_{0}$ gives the initial similarity matrix $S(0)$ using a standard matrix decomposition. Afterwards, the embeddings for subsequent timestamp are updated by assuming that $||S(\Delta t) - S(0)|| < \epsilon$ for some small $\epsilon$, i.e. the similarity matrix $S(\Delta t)$ is a perturbation of the initial matrix $S(0)$. Therefore, it is possible to update a low dimensional representation of the nodes using first-order matrix perturbation theory in symmetric matrices iteratively~\cite{Stewart90MatrixPerturbation}. Li et al.~\cite{Li2017Attributed} apply matrix perturbation theory for Laplacian Eigenmaps, whereas Zhang et al.~\cite{Zhang2018Timers} propose TIMERS to update SVD decompositions incrementally. 
    
    It is noteworthy that these approaches accumulate errors due to the perturbative approach made, and it may not be very effective if the network evolves intensely over time. Nevertheless, since for many real networks the temporal evolution is quite sparse (i.e. the number of added or removed edges is much lower than the total number of edges), these methods present promising results in dynamic link prediction, node clustering and node classification. Moreover, these strategies may reduce error accumulation over time, setting a restart time, i.e. a time interval at which the algorithm recalculates the factorization instead of the incremental update~\cite{Zhang2018Timers}.

    \item \textbf{Temporal Matrix Factorization}: In these methods, the time-dependent similarity matrix $S_{t}$ is decomposed by a temporal rank-$k$ matrix factorization model as follows~\cite{Yu2017Temporally}:
    \begin{equation}
    S(t) = h(U \times V(t)^{T}),
    \end{equation}
    where both $U$ and $V(t)$ are $|V| \times k$ matrices, $U$ is a constant matrix, $V(t)$ is a time-dependent matrix and $h(\cdot)$ is an element-wise function. For undirected networks, $S(t)$ is symmetric and it is possible to (i)~average $U V(t)^{T}$ and its transpose as the prediction of $S(t)$; or (ii)~factorize $S(t)$ as the product of a time-dependent matrix $V(t)$ and its transpose~\cite{Yu2017Temporally}. Therefore, this method learns two types of embedding: (i)~a constant term embedding, given by the rows of $U$ and represents persistent properties between pairs of nodes; and (ii)~a time-varying embedding, given by the rows of $V(t)$ and represents changes in topology over time. A non-linearity may be inserted by the function $h$, e.g. a logistic function to interpret the reconstruction $U \times V(t)^{T}$ as a probability measure for the similarity.
    
    The main challenge to handle this approach is to describe the time-dependent matrices $V(t)$ for each timestamp. The dynamic behavioral mixed-membership model~(DBMM) proposed by Rossi~et~al.~\cite{Rossi2013Modeling} was the first  to employ this factorization, and proposed: (i)~a transition matrix $T$ in order to bound $V(t+\Delta t) \approx V(t) T$, (ii)~a stacked transition model, which bounds training examples from $l$ previous timestamps, and (iii)~a summary transition model, defining $V(t)$ at a specific timestamp as a linear combination of time-dependent matrices at previous timestamps. Yu et al.~\cite{Yu2017Temporally}, who developed the formal description of the temporal matrix factorization approach described above, represented $V(t)$ as a polynomial function over time of order $p$, i.e. $V(t) = \sum_{i=0}^{p} W^{(i)} t^{i}$, where $\{W^{(i)}\}_{i=0}^{p}$ are $|V| \times k$ matrices which need to be learnt from the model along with $U$. The LIST model~\cite{Yu2017Link} leverages a temporal matrix factorization to learn the feature vector of each node by simultaneously optimizing the temporal smoothness constraint and network propagation constraint, ensuring that two vertices that are connected are likely to share similar features.
\end{compactitem}

Table~\ref{tab:matrixfactorization} compares the approaches based on matrix factorization discussed in this section. Some combinations of the matrix factorization approaches were not yet explored, such as graph Laplacian eigenmaps including a temporal smoothing term and a Laplacian temporal matrix factorization.

\begin{table}[H]
\caption{Matrix factorization based Dynamic Graph Embedding.}
\scriptsize
\vspace{-3mm}
\label{tab:matrixfactorization}
\centering
\tabcolsep=0.05cm
\renewcommand{\arraystretch}{1.1}
\begin{tabular}{|c|c|c|c|c|c|c|c|c|}
\hline
\multicolumn{2}{|c|}{} & \multicolumn{7}{|c|}{\textbf{Matrix Factorization based Dynamic Graph Embedding}} \\ \cline{3-9}
 \multicolumn{2}{|c|}{}& BCGD~\cite{Zhu2016Scalable} & \cite{Ferreira2019Modeling} & DANE~\cite{Li2017Attributed} & TIMERS~\cite{Zhang2018Timers} & DBMM~\cite{Rossi2013Modeling} & TMF~\cite{Yu2017Temporally} & LIST~\cite{Yu2017Link} \\ \hline
\multicolumn{1}{|c|}{\multirow{2}{*}{Cost Function Type}} & Laplacian Eigenmaps &  &  & \checkmark &  &  &  & \\ \cline{2-9} 
\multicolumn{1}{|c|}{} & Similarity Matrix Factorization & \checkmark & \checkmark &  & \checkmark & \checkmark & \checkmark & \checkmark \\ \hline
\multirow{3}{*}{Temporal Dependence} & Temporal Smoothing & \checkmark & \checkmark &  &  &  & & \\ \cline{2-9} 
 & Incremental Updates &  &  & \checkmark & \checkmark &  & & \\ \cline{2-9} 
 & Temporal Matrix Factorization &  &  &  &  & \checkmark & \checkmark & \checkmark \\ \hline
\end{tabular}
\end{table}

\subsubsection{Tensor Factorization}

Tensors are higher-order generalizations of vectors and matrices, represented as $X \in \mathbb{R}^{I_{1} \times I_{2} \times ... \times I_{N}}$,  where the order of $X$ is $N > 2$~\cite{Acar2008Unsupervised}. Dynamic networks are usually expressed as three-way tensors, i.e. $N = 3$. Several tensor factorization methods that can extract latent structure in the data have been proposed, including CANDECOMP/PARAFAC (CP) family, Tucker family, and alternative models~\cite{Acar2008Unsupervised}.

For dynamic networks, CP decomposition is the most popular approach, since it learns both node embeddings and temporal embeddings with computational efficiency. Dunlavy et al.~\cite{Dunlavy2011Temporal} used the temporal profiles computed by CP as a basis for predicting the scores in future timestamps, using a forecasting method for time-series data with periodic patterns. Rafailidis and Nanopoulos~\cite{Rafailidis2014Modeling} represent continuous user-item interactions over time using a three-way tensor by proposing a measure of user-preference dynamics~(UPD) that captures the rate at which the current preferences of each user have been shifted, and generate recommendations based on CP tensor factorization.

Even though models based on Tucker decomposition achieve good performance, they require a lot of computational power, which may explain the reason every tensor factorization approach for dynamic graphs relies on CP decomposition~\cite{Fang2015Personalized}. The alternative models discussed by Acar et al.~\cite{Acar2008Unsupervised}, including Multilinear Engine~(ME), STATIS, and multiblock multiway models are also yet to be explored by dynamic graph embedding methods.

\subsection{Approaches based on Deep Learning} \label{subsec:deeplearning}

Deep Learning has shown a remarkable performance in a wide variety of research fields, including computer vision and language modeling, and it also benefits several applications related to representation learning and other tasks over graphs. Many successful static graph embedding methods have been proposed in the last few years, from Graph Neural Networks~\cite{Scarselli2008Graph} and Convolutional Neural Networks on Graphs~\cite{Niepert2016Learning}, to autoencoders, such as Structural Deep Network Embedding (SDNE)~\cite{Wang2016Structural}.

Different architectures based on neural networks are used both to extract the topological properties of a network and to capture temporal dependencies therein. Some preliminary works developed a fully-connected neural networks, either using them as an autoencoder or as a part of the decoding process~\cite{Goyal2018Dyngem, Goyal2019dyngraph2vec,Chen2018GC, Wei2019Lifelong}. Architectures based on recurrent neural networks~(RNNs), including long short-term memory units~(LSTMs)~\cite{Hochreiter1997Long} and gated recurrent units~(GRUs)~\cite{Cho2014Learning}, leverage a sequence of graphs, or their representations, in order to learn or enhance embeddings taking into consideration temporal correlation, and store information over time to handle more complex correlations beyond consecutive timestamps~\cite{Goyal2019dyngraph2vec,Chen2019lstm,Li2015Gated}. Attention mechanisms~\cite{Bahdanau2014Neural} are used to further improve understanding of the most relevant time points for each representation~\cite{Sankar2018Dynamic, Xu2019Spatio}. Convolutional neural networks~(CNNs)~\cite{Krizhevsky2012Imagenet} for graphs have been widely adopted in order to handle topological properties, including Graph Convolutional Networks~(GCNs)~\cite{Kipf2016Semi}. Iterative propagation procedures have been also employed to learn the graph topology, as in Graph Neural Networks~(GNNs)~\cite{Scarselli2008Graph}, GraphSAGE~\cite{Hamilton2017Inductive}, and Gated Graph Neural Networks~(GGNNs), the former also being able to learn the reachability across the nodes in a graph using GRUs~\cite{Li2015Gated}. These approaches have been also explored by several dynamic graph embedding methods~\cite{Chen2018GC, Gao2019dyngraph2seq}. Even further, instead of using RNNs to explore the characteristics of the network over time, many succeeding models also employ convolutional networks~(for instance, 1D CNNs) to leverage temporal dependencies~\cite{Bonner2018Temporal, Deng2019Learning, Zhao2019T, Xiong2019Dyngraphgan}.

Neural networks may also learn an approximation of the network distribution by using generative models, such as the variational autoencoders (VAEs)~\cite{Kingma2013Auto} and generative adversarial networks (GANs)~\cite{Goodfellow2014Generative}. These approaches learn low-dimensional latent representations of the training data that store information about the type of output the model needs to generate, using a generative network to capture the data distribution and a recognition network~(or a discriminator network) to estimate the probability that a sample came from the data distribution. Variational graph autoencoder~(VGAE)~\cite{Kipf2016Variational} and Graph-GAN~\cite{Wang2018Graphgan} employ these approaches to static graphs. These generative models are used in dynamic graphs to learn these data distributions over time~\cite{Bonner2018Temporal, Mahdavi2019Dynamic, Bonner2019Temporal, Zhao2019Large, Hajiramezanali2019Variational, Xiong2019Dyngraphgan, Lei2019GCNGAN}.

\begin{figure}[htp]
    \centering
    \includegraphics[width=0.45\linewidth]{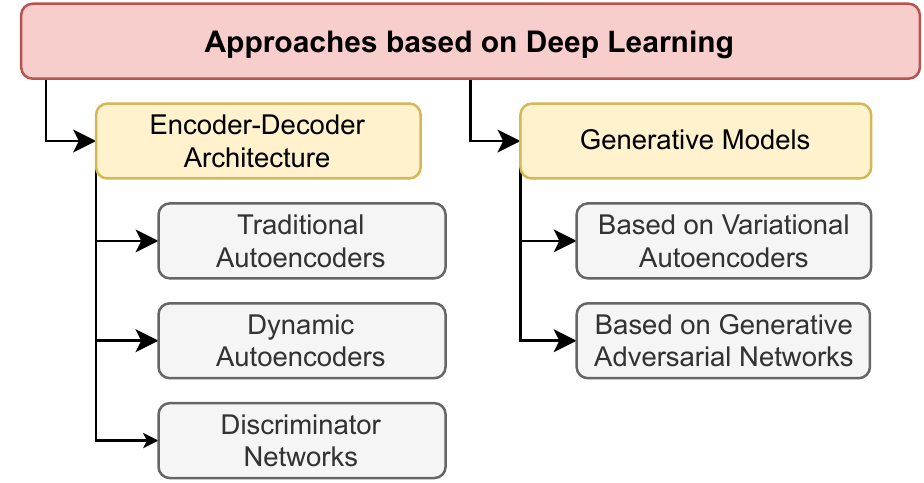}
    \caption{Classification of deep learning techniques for dynamic graph embedding, dividing into encoder-decoder perspective and generative models.}
    \label{fig:deeplearning}
\end{figure}

In order to define a taxonomy that categorizes every method with minimal overlap between different categories, we classify the approaches according to 
the general architecture of the neural network. We have identified two main general architectures: (i)~encoder-decoder perspective, which holds the majority of works concerning learning representations and decoding them for an application~(i.e. network reconstruction, link prediction, or node/graph classification); and (ii)~encoder, sampling, and decoder perspective, which contains the generative models and handles representations as probability distributions~(see Figure~\ref{fig:deeplearning} for the complete proposed taxonomy for deep learning based approaches). In the following, we further detail each of these methods and also present the most important existing works for each of them.

\subsubsection{Encoder-Decoder Architecture}

The encoder-decoder architecture consists of an encoder, which maps the data into a low-dimensional representation, and a decoder, that aims to either (i)~reconstruct the original data; or (ii)~solve an application-driven problem, such as a binary or multi-class classification problem. In the first case, the network is called an autoencoder, since the encoding seeks to be as lossless as possible. In the second case, the encoding is lossy and the original data cannot be fully recovered. 

Concerning dynamic graphs, an encoder receives graph snapshots as input and the decoder may exhibit three distinct outputs: (i)~traditional autoencoders, whose representations reconstruct each graph snapshot, hence following the lossless encoding paradigm; (ii)~dynamic autoencoders, whose representations do not reconstruct each graph snapshot whereas instead they reconstruct a snapshot in a future timestamp, therefore predicting the network structure; and (iii)~discriminator networks, whose embeddings do not reconstruct the network topology at all, but are intended to learn node labels, node clusters or a global network property, and the loss functions are usually application-driven.

\begin{compactitem}
    \item \textbf{Traditional Autoencoders}: These architectures are applied for each graph snapshot, similar to the SDNE in static graphs~\cite{Wang2016Structural}. DynGEM~\cite{Goyal2018Dyngem} builds a fully-connected autoencoder for each graph snapshot, using a transfer learning paradigm to share parameters between two consecutive autoencoders and a strategy that allows the autoencoder network to widen its layers and inserts new layers in order to handle a growing number of nodes in the graph. LDANE~\cite{Wei2019Lifelong} follows a similar strategy, and handles node attributes by adding a margin-based ranking loss term in the loss function that ensures the embeddings of two similar nodes are closer than the embedding of two non-similar nodes.
    
    Models employing recurrent neural networks and their hidden states to encode dynamic graph structure have also been proposed. For instance, Taheri et al.~\cite{Taheri2019Learning} developed DyGGNN, which leverages Gated Graph Neural Networks~(GGNNs) to capture graph topology and couples it with an LSTM encoder to handle graph dynamics, and with an LSTM decoder to reconstruct the structure of the dynamic graph at each timestamp. Other approaches include DySAT~\cite{Sankar2018Dynamic} and DGNN~\cite{Ma2020Streaming}. 
    
    Table~\ref{tab:traditionalautoencoders} summarizes the deep learning approach employed by each of these methods.
    
    \begin{table*}[h]
    \caption{Traditional Autoencoders for Dynamic Graph Embedding.}
    \scriptsize
    \vspace{-3mm}
    \label{tab:traditionalautoencoders}
    \centering
    \tabcolsep=0.05cm
    \renewcommand{\arraystretch}{1.1}
    \begin{tabular} {  |c|c| } \hline
    \textbf{Algorithm}& \textbf{Deep Learning Model} \\ \hline
    DynGEM~\cite{Goyal2018Dyngem} & Fully-Connected Autoencoder (based on SDNE~\cite{Wang2016Structural}) \\ \hline
    LDANE~\cite{Wei2019Lifelong} & Similar to DynGEM, also handles node attributes (margin-based ranking loss term)  \\ \hline
    DyGGNN~\cite{Taheri2019Learning} & Gated Graph Neural Networks and LSTMs \\ \hline
    DySat~\cite{Sankar2018Dynamic} & Structural and Temporal Self-Attention \\ \hline
    DGNN~\cite{Ma2020Streaming} & Attentive LSTMs \\ \hline
    \end{tabular}
    \vspace{-1mm}
    \end{table*}
    
    \item \textbf{Dynamic Autoencoders}: The input of these methods based on dynamic autoencoders is the historical records of the network. The output is the reconstructed graph at a future time. Bonner et al.~\cite{Bonner2018Temporal} regarded this approach as the temporal graph offset reconstruction problem, i.e. creating temporal graph embeddings that recreate a future timestamp of the graph.
    
    Several embedding methods developed a recurrent architecture to capture the dynamics of the network in order to predict its future state, such as Goyal et al.~\cite{Goyal2019dyngraph2vec}, which propose three different strategies for taking a set of graph snapshots, from autoencoders (dyngraph2vecAE) to LSTM networks (dyngraph2vecRNN), and the combination of both of them (dyngraph2vecAERNN). Chen et al.~\cite{Chen2019lstm} propose an encoder-LSTM-decoder (E-LSTM-D), which resembles dyngraph2vecAERNN, but uses rectified linear units~(ReLUs) as the activation function for each encoder/decoder layer, and adds a regularization term to prevent overfitting. AdaNN~\cite{Xu2019Adaptive} employs a triple attention module, leveraging topology, node attributes and temporal attention to further feed them into two connected GRUs and concatenating them into a joint state vector. TRRN~\cite{Xu2021Transformer} adopts memories to enhance temporal capacity, applying multi-head self-attention and learning contextualized representations feeding different factors (including node features and topological features) and updated memories into LSTMs.
    
    Graph convolution combined with recurrent units have been exploited for building dynamic autoencoders. GC-LSTM~\cite{Chen2018GC} uses convolutions to extract topological features while coupling with an LSTM in order to learn temporal features of the dynamic network. EvolveGCN~\cite{Pareja2020EvolveGCN} proposes two versions that follow a similar approach: (i)~EvolveGCN-H, where the GCN parameters are hidden states of GRUs that take node embeddings as input; and (ii)~EvolveGCN-O, where the GCN parameters are input/output of an LSTM unit. T-GCN~\cite{Zhao2019T} uses GCNs to learn topological structures, then passing these features to GRUs in order to extract temporal dependencies. 
    
    Table~\ref{tab:dynamicautoencoders} summarizes the deep learning approach employed by each of the methods based on dynamic autoencoders.
    
    \begin{table*}[h]
    \scriptsize
    \caption{Dynamic Autoencoders for Dynamic Graph Embedding.}
    \vspace{-3mm}
    \label{tab:dynamicautoencoders}
    \centering
    \tabcolsep=0.05cm
    \renewcommand{\arraystretch}{1.1}
    \begin{tabular} {  |c|c| } \hline
    \textbf{Algorithm}& \textbf{Deep Learning Model} \\ \hline
    TO-GAE~\cite{Bonner2018Temporal} & GCNs over time \\ \hline
    dyngraph2vecAE~\cite{Goyal2019dyngraph2vec} & Fully-Connected (FC) Autoencoders \\ \hline
    dyngraph2vecRNN~\cite{Goyal2019dyngraph2vec} & Sparsely Connected LSTMs \\ \hline
    dyngraph2vecAERNN~\cite{Goyal2019dyngraph2vec} & FC Encoder, LSTMs and FC Decoder \\ \hline
    E-LSTM-D~\cite{Chen2019lstm} & FC Encoder, LSTMs and FC Decoder \\ \hline
    AdaNN~\cite{Xu2019Adaptive} & Spatial, Attribute-Topology and Temporal Attention, and GRUs \\ \hline
    TRRN~\cite{Xu2021Transformer} & FC Encoder, Transformer-Style Self-Attention and LSTMs \\ \hline
    GC-LSTM~\cite{Chen2018GC} & Graph Convolutions and LSTMs \\ \hline
    EvolveGCN-H~\cite{Pareja2020EvolveGCN} & GCNs and GRUs \\ \hline
    EvolveGCN-O~\cite{Pareja2020EvolveGCN} & GCNs and LSTMs \\ \hline
    T-GCN~\cite{Zhao2019T} & GCNs and GRUs \\ \hline
    \end{tabular}
    \vspace{-1mm}
    \end{table*}
    
    \item \textbf{Discriminator Networks:} This approach considers that a neural network must learn application-driven representations, such as properly classifying nodes or graphs/subgraphs over time~\cite{Meng2018Subgraph}, extracting network properties~\cite{Gao2019dyngraph2seq}, or predicting a specific global feature~\cite{Yu2017Spatio}. Several approaches combine GCNs and recurrent networks, such as DynGraph2Seq~\cite{Gao2019dyngraph2seq},     TSGNet~\cite{Park2019Exploiting} and NAAM~\cite{Shrestha2019Learning}. Topological features may also be extracted by other techniques as an alternative to GCNs, as Xu et al. showed by implementing STAR~\cite{Xu2019Spatio}. Other approaches use convolutions for both spatial and temporal feature extraction, including Spatio-Temporal Graph Convolutional Network~(STGCN)~\cite{Yu2017Spatio}.
    
    These discriminator networks are commonly employed to embed graphs constructed from non-relational data. For instance, DynamicGCN~\cite{Deng2019Learning} extracts and learns graph representations from historical event documents, encoding the input data into a sequence of graphs with node embeddings, and developing a graph convolutional network model to predict the occurrence of certain type of events. TD-Graph LSTM~\cite{Yuan2017Temporal}, in the other hand, is applied to action-driven video object detection, passing each frame through a spatial convolutional network in order to detect similar regions in consecutive frames, and construct a temporal graph structure by connecting semantically similar regions. LSTM units take the spatial visual features as the input states, incorporating temporal motion patterns for participating objects in the action while minimizing an action-driven object categorization loss. Li et al.~\cite{Li2019Learning} propose a spatial-temporal graph embedding model called STG2Vec, which includes a temporal attention and incorporates multi-source information to fed a collaborative temporal modeling based on LSTMs.
    
    Table~\ref{tab:discriminatornetworks} lists the approaches described above and summarizes the deep learning approach employed by each of them, along with the specific task each method attempts to solve.
    
    \begin{table*}[h]
    \scriptsize
    \caption{Discriminator Networks for Dynamic Graph Embedding.}
    \vspace{-3mm}
    \label{tab:discriminatornetworks}
    \centering
    \tabcolsep=0.05cm
    \renewcommand{\arraystretch}{1.1}
    \begin{tabular} {  |c|c|c| } \hline
    \textbf{Algorithm}& \textbf{Deep Learning Model} & \textbf{Task} \\ \hline
    DynGraph2Seq~\cite{Gao2019dyngraph2seq} & GCNs, LSTMs and Hierarchical Attention & Sequence of Target Health Stages \\ \hline
    TSGNet~\cite{Park2019Exploiting} & GCNs and LSTMs & Node Classification \\ \hline
    NAAM~\cite{Shrestha2019Learning} & GCNs, LSTMs (or BiLSTMs) and Temporal Attention & Forecasting User Interactions \\ \hline
    STAR~\cite{Xu2019Spatio} & Spatio-Temporal Attention and GRUs & Node Classification \\ \hline
    STGCN~\cite{Yu2017Spatio} & GCNs and Gated Convolutional Neural Networks & Traffic Forecasting \\ \hline
    DynamicGCN~\cite{Deng2019Learning} & GCNs with Updates from Previous Timestamps & Event Prediction \\ \hline
    TD-Graph LSTM~\cite{Yuan2017Temporal} & CNNs and LSTMs & Missing Label Classification \\ \hline
    STG2Vec~\cite{Li2019Learning} & Temporal Attention and LSTMs & Bike-sharing Demand \\ \hline
    \end{tabular}
    \vspace{-1mm}
    \end{table*}
    
\end{compactitem}

\subsubsection{Generative Models}

Generative algorithms attempt to predict features given a certain label, i.e. to learn the data distribution patterns, in contrast to discriminative models, which attempt to learn representations and classify each input data. Therefore, these approaches have the power to synthesize data in addition to compress and to learn embeddings. Regarding generative neural networks for dynamic graph embedding, two groups of approaches arise: (i)~methods based on variational autoencoders~(VAEs), which encodes an input as a distribution over the latent space; and (ii)~methods based on generative adversarial networks~(GANs), which train both a generator network to synthesize graphs and a discriminator network to distinguish between true graphs and generated ones.

\begin{compactitem}
    \item \textbf{Based on Variational Autoencoders:} The encoder for a variational autoencoder takes a data point and produces a distribution, usually parameterized as a multivariate Gaussian. In this case, the encoder predicts the mean and standard deviation of the Gaussian distribution, and the lower-dimensional embedding is sampled from this distribution. The decoder is a variational approximation, which takes an embedding and produces an output~\cite{Kingma2013Auto}. 
   
    Every variational autoencoder for dynamic graphs is inspired by VGAE~\cite{Kipf2016Variational} to address the encoding of each snapshot, differing on how to handle the graph evolution. These methods include (i) Dyn-VGAE~\cite{Mahdavi2019Dynamic}, which addresses a temporal smoothness loss term, (ii) TO-GVAE~\cite{Bonner2018Temporal}, which applies VGAE over time to reconstruct subsequent snapshots, and (iii) VGRNN~\cite{Hajiramezanali2019Variational}, which adopts VGAE whose prior distribution parameters are based on the hidden states in previous timestamps. 
    
    There are several proposals to enhance the encoder model. For instance, Bonner et al.~\cite{Bonner2019Temporal} propose a Temporal Neighbourhood Aggregation (TNA) block to comprise a GCN with a GRU in the encoder, controlling the combination of topological and temporal learning via a final linear layer. Zhao et al.~\cite{Zhao2019Large} have developed a framework called BurstGraph, which splits the adjacency matrix of a graph into a standard adjacency matrix and a burst adjacency matrix to pick up unexpected behavior within a time duration. A variational autoencoder is employed using GraphSAGE~\cite{Hamilton2017Inductive} as the encoder, and two decoders: a standard decoder, which learns representations $Z^{v}$, and a bursty decoder, which learns sparse embeddings~$Z^b$.

    \item \textbf{Based on Generative Adversarial Networks:} 
    Generative adversarial networks (GANs) are algorithmic architectures that use two neural networks, pitting one against the other by designing a game-theoretical minimax game to combine generative and discriminative models. This approach has been applied for graph representation learning by a framework called GraphGAN~\cite{Wang2018Graphgan}. While a generator network attemps to approximate the true graph connectivity distribution, the discriminator network aims to discriminate the connectivity of each node pair. Therefore, the generator network  tries to deceive the discriminator network, whereas the discriminator network improves itself to distinguish better and better between true edges and generated edges.
    
    Inspired by these recent approach for graph representation learning, some methods arise for dynamic graph embedding. DynGraphGAN~\cite{Xiong2019Dyngraphgan} designs the discriminator network with the following components: (i)~a GCN to encode neighborhood features of nodes; and (ii)~CNNs to learn temporal graph evolution along the time dimension. The generator network implements a sigmoid function of the inner product of two node's embeddings at a timestamp~$t$ to estimate the probability distribution of an edge connecting these nodes at time~$t$.  GCN-GAN~\cite{Lei2019GCNGAN} displays an architecture whose generative network consists of a GCN layer, an LSTM layer and a fully-connected output layer, and the discriminator network is a fully-connected feedforward neural network. 
    
    Each of the GAN-based methods previously mentioned are trained with a single graph. Therefore, the trained model is capable of generating artificial snapshots following a similar structure and dynamic of the original graph used during training. Also, note that the resulting model is limited to creating snapshots with the same number of nodes as the graph used in training, given that number of parameters of the model is proportional to the number of nodes in the TVG. This dependency is observed, for example, for the last layer of the generative model of the GCN-GAN model, which builds an adjacency matrix with dimensions $N\times N$ from an embedding vector (outputted by the LSTM layer).
\end{compactitem}

Table~\ref{tab:generative} summarizes the generative model and architectures employed by  the methods discussed in this section.

\begin{table*}[h]
\caption{Generative Models for Dynamic Graph Embedding.}
\scriptsize
\vspace{-3mm}
\label{tab:generative}
\centering
\tabcolsep=0.05cm
\renewcommand{\arraystretch}{1.1}
\begin{tabular} {  |c|c|c|c| } \hline
\textbf{Algorithm}& \textbf{VAE}& \textbf{GAN} & \textbf{Deep Learning Models} \\ \hline
TO-GVAE~\cite{Bonner2018Temporal} & \checkmark & & GCNs  \\ \hline
Dyn-VGAE~\cite{Mahdavi2019Dynamic} & \checkmark & & Original VAE with Temporal Smoothness \\ \hline
TNA~\cite{Bonner2019Temporal} & \checkmark & &  GCNs and GRUs   \\ \hline
\cite{Zhao2019Large} & \checkmark & & GraphSAGE \cite{Hamilton2017Inductive} and RNNs  \\ \hline
VGRNN~\cite{Hajiramezanali2019Variational} & \checkmark & & GCNs and LSTMs  \\ \hline
DynGraphGAN~\cite{Xiong2019Dyngraphgan} & & \checkmark & GCNs and CNNs  \\ \hline
GCN-GAN~\cite{Lei2019GCNGAN} & & \checkmark & GCNs and LSTMs \\\hline
\end{tabular}
\vspace{-1mm}
\end{table*}

\subsection{Random Walk Approaches}

Another class of methods for graph embedding functions relies on random walks. Multiple random walks of fixed length $L$ are considered sentences, generating a context for each node and trying to extract higher-order dependencies without adjacency matrices. 
 
The node sequence matrix is therefore generated and factorized, usually, by applying a neural network architecture, the most popular being the Skip-Gram~\cite{Mikolov2013Efficient, Mikolov2013Distributed}, to produce low dimensional vector representations for each node while maintaining their proximity in the new embedded space.

Random walks applied to dynamic graphs must generate time-dependent contexts $C(t)$ in addition to sequences that capture topological dependencies. Then, the methods based on random walks are separated according to how they include the temporal aspect into the calculation: (i)~random walk on snapshots, where a time-dependent node sequence matrix is generated by applying random walks starting on each node at each snapshot, and further optimizing a joint problem that takes into account the temporal dependency; (ii)~evolving random walks, where the node sequences are generated for the initial time (first snapshot), then the method incrementally updates node representation by updating random walks starting on nodes affected by topological evolution; and (iii)~temporal random walks, which define time-dependent context matrices by allowing random walks across consecutive timestamps and considering time ordering restriction. Also, other node sequence sampling methods besides random walks may be applied to generate contexts, including neighborhood aggregation. In the following, we further detail each of these methods.

\subsubsection{Random Walks on Snapshots} \label{subsub:rws}

This approach performs random walks on each snapshot of a dynamic graph, obtaining vector representations by optimizing a joint problem taking into account temporal dependencies. It is important to note that methods following this approach generate contexts whose temporal connection between two matrices at consecutive time points is not modeled by random walks. Instead, the temporal dependency is defined later in the generation of the embeddings, taking into account the temporal smoothness. For instance, embeddings may be learned for each graph snapshot independently using methods including node2vec and DeepWalk, and afterward, the representations may be combined using operations, from simple vector concatenation to dynamic embeddings and orthogonal transformations to align embedding vectors at consecutive timestamps.

De Winter et al.~\cite{De2018Combining} and Dyn2Vec~\cite{Mitrovic2019dyn2vec} apply vector concatenation to node representations over time. The former applies node2vec for each snapshot, whereas the latter employs a DeepWalk variant, whose probability of choosing a certain edge depends on the normalized edge weight. Chen et al.~\cite{Chen2019Dynamic} initialize node embeddings by using a Gaussian prior with a diagonal covariance, and learn representations over time using dynamic Bernoulli embeddings, considering the rows of the node sequence matrix as the context for each node. tNodeEmbed~\cite{Singer2019Node} preserves static network neighborhoods of nodes in a $d$-dimensional feature space by using Orthogonal Procrustes, and optimizes a LSTM for specific tasks~(i.e., link prediction and multi-label node classification). DynSEM~\cite{Zhou2019Dynamic} train node embeddings for each timestamp using node2vec, align node embeddings into a common space using Orthogonal Procrustes, and optimize a joint loss function taking into account temporal smoothness.

Table~\ref{tab:randomwalk} lists the methods described above and points out both static embedding method applied for each snapshot and how they handle temporal dependencies.

\begin{table*}[h]
\scriptsize
\caption{Random Walks on Snapshots.}
\vspace{-3mm}
\label{tab:randomwalk}
\centering
\tabcolsep=0.05cm
\renewcommand{\arraystretch}{1.1}
\begin{tabular} {  |c|c|c| } \hline
\textbf{Algorithm}& \textbf{Static Embedding Method}& \textbf{Temporal Dependence Handling} \\ \hline
(Static) \cite{De2018Combining}  & node2vec & Vector Concatenation \\ \hline
Dyn2Vec \cite{Mitrovic2019dyn2vec} & DeepWalk variant & Vector Concatenation \\ \hline
\cite{Chen2019Dynamic} & Gaussian Initialization & Dynamic Bernoulli Embeddings  \\ \hline
tNodeEmbed \cite{Singer2019Node} & node2vec & Orthogonal Procrustes and LSTM \\ \hline
DynSEM \cite{Zhou2019Dynamic} & node2vec & Orthogonal Procrustes and Temporal Smoothing Loss Function \\ \hline
\end{tabular}
\vspace{-1mm}
\end{table*}

\subsubsection{Evolving Random Walks}
\label{subsub:erw}

Generating random walks for every timestamp is an expensive time-consuming process. Several approaches first generate embeddings for the initial timestamp by using a static random walk approach, then incrementally update node representations taking into account that, in general, only a few nodes are influenced by topological evolution. Dynnode2vec~\cite{Mahdavi2018dynnode2vec} follows this approach by sampling node sequences for only evolving nodes instead of generating random walks for all nodes in a given timestamp and afterward feeding these sequences as an input to a dynamic Skip-Gram model~\cite{Bamler2017Dynamic}, which is initialized at the first snapshot and used for weight initialization of the Skip-Gram of subsequent timestamps.

Other approaches include EvoNRL~\cite{Heidari2018EvoNRL}, which initially employs node2vec, stores the random walks in memory, and updates the set of random walks when a single new edge arrives in the network. EvoNRL uses a similar dynamic Skip-Gram model previously mentioned~\cite{Bamler2017Dynamic}. Sajjad~et~al.~\cite{Sajjad2019Efficient} follow the same Skip-Gram implementation over time and proposes random walk update algorithms aiming to be statistically indistinguishable from a set of random walks generated from scratch on the new graph. NetWalk~\cite{Yu2018NetWalk} also proposes a network embedding algorithm inspired by the Skip-Gram architecture (which the authors call Clique Embedding). It uses a deep autoencoder neural network to learn vector representations through a stream of random walks while minimizes the pairwise distance among all nodes in each walk. Evolving random walks are used in order to mitigate the computational cost of performing a full random walk in every snapshot. The former is not expected to be as precise as the latter, as shown in some approaches~\cite{Sajjad2019Efficient}. But the small loss in accuracy is compensated by the huge computational efficiency shown by these methods.

Table~\ref{tab:evolvingwalk} lists the methods based on evolving random walks and points out both static embedding methods applied for each snapshot and how they update random walks and vector representations.

\begin{table*}[h]
\caption{Methods based on Evolving Random Walks.}
\scriptsize
\vspace{-3mm}
\label{tab:evolvingwalk}
\centering
\tabcolsep=0.05cm
\renewcommand{\arraystretch}{1.1}
\begin{tabular} {  |c|c|c| } \hline
\textbf{Algorithm}& \textbf{Static Embedding Method} & \textbf{Update Method} \\ \hline
dynnode2vec~\cite{Mahdavi2018dynnode2vec} & node2vec & Dynamic Skip-Gram Model \\ \hline
EvoNRL~\cite{Heidari2018EvoNRL} & node2vec & Skip-Gram Model over Time \\ \hline
\cite{Sajjad2019Efficient} & DeepWalk with Unbiased Random Walk Updates & Skip-Gram Model over Time \\ \hline
NetWalk~\cite{Yu2018NetWalk} & Clique Embedding (AutoEncoder) & Vertex Reservoir and Walk Updating \\ \hline
\end{tabular}
\vspace{-1mm}
\end{table*}

\subsubsection{Temporal Random Walk Methods}

The methods described in Sections~\ref{subsub:rws} and \ref{subsub:erw} consider that random walks and their updates are made to each snapshot separately. However, one way to include the time dependency directly in a sequence of nodes generated by random walks is to build a method to create a corpus of walks over time, respecting the temporal flux. In the literature, these walks are regarded as temporal walks~\cite{Nguyen2018Continuous, Casteigts2012Time}.

It is possible to generalize the Skip-Gram architecture to handle continuous-time dynamic networks, as described by Nguyen et al.~\cite{Nguyen2018Continuous}. In particular, the authors propose a general framework called CTDNE for learning time-preserving embeddings and propose several methods to select the subsequent nodes from a starting node, thus performing a temporal random walk: (i)~an unbiased temporal neighbor selection; (ii)~a biased selection, which may be based on temporal exponentially-weighted decay~(i.e., older timestamps have an exponentially lower contribution to the selection) or on temporal linearly-weighted decay. 

Several approaches follow CTDNE's paradigm, including Wu et al.~\cite{Wu2019T}, which developed T-EDGE to encompass weighted networks, and De Winter et al.~\cite{De2018Combining}, proposing a continuous-time version of node2vec. STWalk2 proposed by Pandhre et al.~\cite{Pandhre2018stwalk}, on the other hand, generates a spatial walk for each snapshot and a temporal walk, employing a Skip-Gram network to combine the two learned embeddings to get node representations. LSTM-node2vec~\cite{Khoshraftar2019Dynamic} trains an LSTM autoencoder with node sequences generated by temporal random walks, and afterward initialize node2vec with the input layer weights of the trained LSTM encoder for each snapshot at time~$t$.

Diffusion prediction problems are related to temporal random walks, but the exact timestamp of diffusion is not necessarily known. instead only the temporal ordering is defined~(i.e., typically one does not know exactly when some information have passed from a node to another, but the source and the target of the diffusion process is known). Models following this objective include (i) DeepCas~\cite{Li2017DeepCas}, which uses GRUs and attention mechanisms to predict the future size of the cascade, (ii) DAN~\cite{wang2018attention}, which outputs the probability distribution of the next infected node leveraging feed-forward neural networks and attention mechanism, and (iii) Topo-LSTM~\cite{Wang2017Topological}, employing LSTMs to handle temporal dependence over diffusion. Moreover, Yang et al.~\cite{Yang2019Multi} implement GRUs, GCNs and GraphSAGE to predict next affected nodes, and a reinforcement learning framework to predict cascade size.

Temporal Random Walks represents a more natural way to deal with dynamic continuous graphs~\cite{Nguyen2018Continuous,De2018Combining}, since it does not require any time discretization of the graph into snapshots. It is also the ideal approach for diffusion problems, as we have shown.
Table~\ref{tab:temporalrandomwalk} lists the temporal random walk methods and points out both static embedding method applied for each snapshot and how they update random walks and vector representations.

\begin{table*}[t]
\scriptsize
\caption{Temporal Random Walk Methods.}
\vspace{-3mm}
\label{tab:temporalrandomwalk}
\centering
\tabcolsep=0.05cm
\renewcommand{\arraystretch}{1.1}
\begin{tabular} {  |c|c|c| } \hline
\textbf{Algorithm}& \textbf Neural Network Model & Comments \\ \hline
CTDNE~\cite{Nguyen2018Continuous} & Skip-Gram model & Defined temporal random walk embedding methods \\ \hline
T-EDGE~\cite{Wu2019T} & Skip-Gram model & Encompasses weighted dynamic networks \\ \hline
\cite{De2018Combining} (Dynamic) & node2vec & Continuous version of the first random walks on snapshots \\ \hline
STWalk2~\cite{Pandhre2018stwalk} & Skip-Gram model & Separates temporal random walks and spatial random Walks \\ \hline
LSTM-node2vec~\cite{Khoshraftar2019Dynamic} & node2vec and LSTMs & Handles both temporal sequences and static sequences \\ \hline
DeepCas~\cite{Li2017DeepCas} & DeepWalk, GRUs and Attention Mechanism & Diffusion cascades \\ \hline
DAN~\cite{wang2018attention} & Feed-forward neural network and attention mechanism & Diffusion cascades\\ \hline
\cite{Yang2019Multi} & GCNs and GraphSAGE & Diffusion cascades \\ \hline
Topo-LSTM~\cite{Wang2017Topological} & LSTMs & Diffusion cascades \\ \hline
\end{tabular}
\vspace{-1mm}
\end{table*}

\subsubsection{Other Node Sequence Sampling Methods}

Some techniques exhibit steps or insights given by two different random walk based approaches. For instance, several models create a graph containing the nodes at a given time $t$ and their neighbors at the same timestamp $t$ and the previous timestamps in a defined time window, and employ random walk procedures leveraging temporal ordering~\cite{Pandhre2018stwalk,Sato2019DyANE}. In particular, DHNE~\cite{Yin2019DHNE} gives exponential-decaying weights for edges connecting nodes and past neighbors. These approaches share elements from random walks over snapshots and temporal random walks. Furthermore, StreamWalk algorithm introduced by Beres et al.~\cite{Beres2019Node} compresses both evolving random walks and temporal random walks by updating the weight for walks to handle more recent edges.

Although most node sequence sampling methods are based on random walks, some techniques rely on other ways to aggregate the neighborhood. Liu et al.~\cite{Liu2019Towards} develop a spatial-temporal neural attention mechanism to estimate the co-occurrence matrix and guide the embedding algorithm to focus on the context information with higher importance. Dynamic Knowledge Graph Embedding~(DKGE)~\cite{Wu2019Efficiently}, on the other hand, applies an attentive GCN~(AGCN) to learn contextual subgraph embeddings over knowledge graphs, integrating them with knowledge embedding of entities and relations to build the joint representation of each object in the graph. The temporal evolution is leveraged by an online learning strategy that learns knowledge embeddings and contextual element embeddings of emerging entities and relations, as well as knowledge embeddings of existing entities and relations with changed contexts~(i.e. whose induced subgraphs are changed). Torricelli et al.~\cite{Torricelli2020weg2vec} introduce weg2vec, which takes a dynamic network and project it into a weighted link stream (the authors called weighted event graph), sampling neighborhoods for events (i.e. the edges of the original dynamic graph) from the link stream (i.e. to create a graph that connects edges concerning involved nodes, co-occurrence, and event time difference), and inputting sequences of connected events to a Skip-Gram model.

Table~\ref{tab:othernodesequencesampling} lists the methods described in this section and shows comments about their peculiarities, i.e., how they leverage temporal dependence or how they choose node context to apply the Skip-Gram model.

\begin{table*}[t]
\caption{Other Node Sequence Sampling Methods.}
\scriptsize
\vspace{-3mm}
\label{tab:othernodesequencesampling}
\centering
\tabcolsep=0.05cm
\renewcommand{\arraystretch}{1.1}
\begin{tabular} {  |c|c| } \hline
\textbf{Algorithm}& Comments \\ \hline
DHNE~\cite{Yin2019DHNE} & Historical-current graphs \\ \hline
STWalk1~\cite{Pandhre2018stwalk} & Similar to historical-current graphs \\ \hline
DyAne~\cite{Sato2019DyANE} & Supra-adjacency representation \\ \hline
StreamWalk~\cite{Beres2019Node} & Temporal random walks updated for affected nodes \\ \hline
DKGE~\cite{Wu2019Efficiently} & Attentive GCNs over subgraphs, and online learning \\ \hline
\cite{Liu2019Towards} & Co-ocurrence matrix using spatial-temporal neural attention model\\ \hline
weg2vec~\cite{Torricelli2020weg2vec} & Weighted event graphs \\ \hline
\end{tabular}
\vspace{-1mm}
\end{table*}

\subsection{Edge Reconstruction based Optimization with Temporal Smoothing} \label{subsec:edgereconstruction}

Following the taxonomic approach proposed by Cai et al.~\cite{Cai2018Comprehensive} for static graphs, some methods for dynamic graphs have been identified whose approach is similar to techniques that directly optimize an objective function based on edge reconstruction. In addition to either maximizing edge reconstruction probability or minimizing edge reconstruction loss, these approaches also preserve temporal smoothness. It is noteworthy that these methods may be understood as reconstructing temporal edges between a node $v$ at a given time $t_{i}$ and the same node at the subsequent timestamp, therefore justifying this category even more adequately for embedding methods.

DynamicTriad~\cite{Zhou2018Dynamic} is a representative method of this category, aiming to preserve both structural information and evolution patterns of a network by modeling how a closed triad (i.e. three vertices connected) develops from an open triad (i.e. three vertices where two of them are not connected). The authors define the probability that an open triad $(v,u,w)$ (where $v$ and $u$ are not connected) evolves into a closed triad, and the probability of the edge $(v,u)$ will not be created, joining these probabilities into a distance-based loss function. Moreover, the model supposes that highly connected nodes should be embedded closely in the low-dimensional vector space and imposes this condition by a margin-based rank loss function, and finally considers temporal smoothness at consecutive time stamps.

Other approaches, solely based on a distance loss function include DNE~\cite{Du2018Dynamic}, and Liu et al.~\cite{Liu2019Real}. Time-Aware KB Embedding~\cite{Jiang2016Encoding} learns node embeddings by modeling relationships as translation operators in the low-dimensional vector space~\cite{Bordes2013Translating} and optimizes a joint margin-based ranking loss function concerning both temporal order score function (the temporal encoding) and translation embeddings (the topological encoding). Table~\ref{tab:edgereconstruction} lists the methods described above, and the loss function each technique aims to minimize, pointing how they handle temporal dependence.
\begin{table*}[t]
\scriptsize
\caption{Edge Reconstruction based Optimization.}
\vspace{-3mm}
\label{tab:edgereconstruction}
\centering
\tabcolsep=0.05cm
\renewcommand{\arraystretch}{1.1}
\begin{tabular} {  |c|c|c| } \hline
\textbf{Algorithm}& \textbf{Temporal Dependence Handling} & \textbf{Loss Function} \\ \hline
DNE~\cite{Du2018Dynamic} & Delta of Theoretical Optimal Solution~\cite{Levy2014Neural} and Temporal Smoothness & Based on LINE~\cite{Tang2015Line}\\ \hline
DynamicTriad \cite{Zhou2018Dynamic} & Temporal Smoothness & Triadic Closure and Social Homophily \\ \hline
\cite{Liu2019Real} & Temporal Smoothness & Based on Laplacian Eigenmaps  \\ \hline
Time-Aware KB Embedding~\cite{Jiang2016Encoding} & Temporal Order Score Function & Joint Margin-Based Rank \\ \hline
\end{tabular}
\vspace{-1mm}
\end{table*}

Note that the temporal smoothing given by a distance-based loss is similar to both matrix factorization problems and skip-gram based models. Indeed, there is a general view that demonstrates the relationship between network embedding approaches, matrix factorization, and Skip-Gram models~\cite{Liu2019General}. Even further, Liu et al.~\cite{Liu2019General} provide a fundamental connection from an optimization perspective, which is the fundamental idea of edge reconstruction based methods. In this survey, these approaches are separated in the taxonomy to follow more strictly algorithmic properties rather than theoretical aspects of loss functions.

\subsection{Methods based on Graph Kernel} \label{subsec:graphkernels}

As presented by Cai et al.~\cite{Cai2018Comprehensive} for static graphs, a few methods handle elementary substructures that are decomposed from a whole graph structure. They incorporate topological attributes built in the network processing step, including graphlet transitions count~\cite{Rahman2016Link}, graphlet frequencies over time~\cite{Dave2019Triangle} and adjacency matrix summation~\cite{Rahman2018dylink2vec}, to learn representations capable of reconstructing such elaborate attributes using a shallow approach of an autoencoder. Hence, since these substructures are used as a topological building block of a static network, dynamic graph embedding takes into account the transitions between different elementary structures. In addition, Béres et al.~\cite{Beres2019Node} developed an online second-order similarity (SecondOrder) that learns neighborhood similarity by Min-Hash fingerprinting, modifying the embedding vector whenever a neighbor of $v_{i}$ gets more similar to $v_{j}$ after adding the edge $(v_{i},v_{j})$ into the network, which may be regarded as a graph kernel based approach.

\subsection{Methods based on Temporal Point Process}

Several dynamic graph embedding techniques, consider that interactions between nodes are stochastic processes whose probabilities depend on the topological structure of the network, on node features, and the network history. For these methods, it is assumed that an event influences a given node and, consequently, it can interact with other nodes in the network, if they are susceptible to the influence of the current node. Therefore, there is a probability that the event will propagate based on the mathematical definitions presented in Section~\ref{subsec:temporalpointprocess}, such as the conditional intensity function.

Main methods in this category include DyRep~\cite{Trivedi2018Representation}, handling a continuous-time deep model of a temporal point process using a conditional intensity function modeling the occurrence of an event $p$ with time scale $k$ between nodes $v_{i}$ and $v_{j}$, and DeepCoEvolve~\cite{dai2016recurrent}, modeling the user-item interaction as a multidimensional temporal point process. Other approaches include KnowEvolve~\cite{Trivedi2017Know}, modeling a fact in a knowledge graph as a temporal point process, M$^{2}$DNE~\cite{Lu2019Temporal}, capturing edge evolution by a temporal point process with an attentive mechanism, in addition to a general dynamics equation concerning the linking rate, and HTNE~\cite{Zuo2018Embedding} with an attention mechanism for neighborhood formation sequence of a node as a counting process. Furthermore, Knyazev et al.~\cite{Knyazev2019Learning} extend DyRep~\cite{Trivedi2018Representation}, replacing the original encoder with a procedure that, given an event between nodes $v_{i}$ and $v_{j}$: (i)~calculates representations of all nodes at time $t_{k-1}$; (ii)~returns an edge embedding for all pair of nodes; (iii)~updates the embedding of node $v_{j}$ based on all edges connected to it; and (iv)~updates the edge embedding between nodes $v_{i}$ and $v_{j}$.

MHDNE~\cite{Yin2019mhdne} models the edge formation process as two temporal sequences with historical edge information and network evolution information therein, respectively. In particular, the network evolution is based on open triangles and triadic closure problem~\cite{Zhou2018Dynamic}, and the intensity function for a new edge creation at time $t$ is given by a Hawkes process, leveraging a term dependent on node embeddings, a time decay function on an exponential form, and the distance between node's neighborhood. Wu et al.~\cite{Wu2020Modeling} propose a Graph Biased Temporal Point Process (GBTPP), which aims to compute the probability of an event propagating to nodes $v_{j} \in \mathcal{N}(v_{i})$ in the future timestamp $t_{k+1}$ given event propagation history and node $v_{i}$, which is influenced by the event at time $t_{k}$.

\subsection{Agnostic Models}

The models described so far use some algorithmic paradigm as a basis for their development. A few approaches in the literature are independent of how the vector representations in each timestamp are obtained. They commit to learning the connections between representations at consecutive time points, or even within a time window. Because of this property of independence of the algorithmic procedure, we classify them as Agnostic Models. Two different paradigms may be followed: (i)~the retrofitted model, where vector representations are learned for the initial graph by using any of the state-of-the-art static network embeddings, then the dynamics are captured by retrofitting the initial embedding with the subsequent graph snapshots; and (ii)~the embedding space transformation method, where the representations of each graph snapshot are calculated by any static method separately, then a transformation function that connects an embedding at time $t$ to the embedding at the next timestamp is learned. In the following, we detail these two paradigms.

\subsubsection{Retrofitted Model}

The retrofitted model~\cite{Saha2018Models} is based on the local temporal smoothness assumption, considering the vertex-centric evolution of the network. Therefore, for the first timestamp, the model employs an existing static embedding method to learn vector representations, but for subsequent timestamps, the vectors are updated by a local update method. In retrofitting, embeddings at previous timestamp $z_{i}(t-1)$ are revised by using the embedding of its neighborhood available from the graph snapshot at time $t$, hence the resulting vector $z_{i}(t)$ is similar to both the prior vector $z_{i}(t-1)$ and the vectors of its adjacent nodes in timestamp $t$. This update rule is carried out until convergence for each time step. It is noteworthy that, except for the first timestamp, the retrofitted model does not learn embeddings directly from data, instead it presumes a temporal smoothing criteria to guarantee node representations over time.

\subsubsection{Embedding Space Transformation Model}

The second agnostic approach assumes that the network time evolution is a global process attempting to fulfill the global temporal smoothness objective by considering the temporal evolution of a network as a transformation over the node embedding vectors of successive timestamps. Once the transformation operator is learned, it can map the latent representation from a known snapshot to the next unobserved snapshot.

Saha et al.~\cite{Saha2018Models} introduce two paradigms for embedding space evolution: (i) \textbf{homogeneous transformation}, where the transformation is assumed to be the same across any two successive timestamps; and (ii) \textbf{heterogeneous transformation}, which refrains from the uniformity assumption i.e. every pair of timestamps has a different transformation procedure. These two paradigms are further discussed in the following:
\begin{compactitem}
    \item \textbf{Homogeneous Transformation:} This transformation is shared across timestamps, and Saha et al.~\cite{Saha2018Models} have proposed a linear transformation in order to learn these mappings between embedding spaces.
    \item \textbf{Heterogeneous Transformation:} Methods based on heterogeneous transformation learn a different projection matrix for each pair of network snapshots. Saha et al.~\cite{Saha2018Models} employ a linear heterogeneous transformation model, learning $N_{S}-1$ different transformation matrices and obtaining a final transformation matrix by combining these different matrices. 
\end{compactitem}

\subsection{Other Dynamic Graph Embedding Approaches}

At last, some methods do not fit into any of the discussed categories, either by presenting some specific methodology and different from all approaches, or by combining different techniques without having one in particular as the main one. It is also interesting to note that some methods aggregate temporal information of the dynamic network into a static graph (i.e. a network containing all interactions and vertices that were present from the beginning of a network until the final timestamp of analysis)~\cite{Dunlavy2011Temporal, Hisano2018Semi}. Therefore, these works use static embedding methods over a network that encompasses temporal information stored at its edges.

\section{Dynamic Graph Embedding Applications}
\label{sec:apps}

\begin{figure}[htp]
    \centering
    \includegraphics[width=0.7\linewidth]{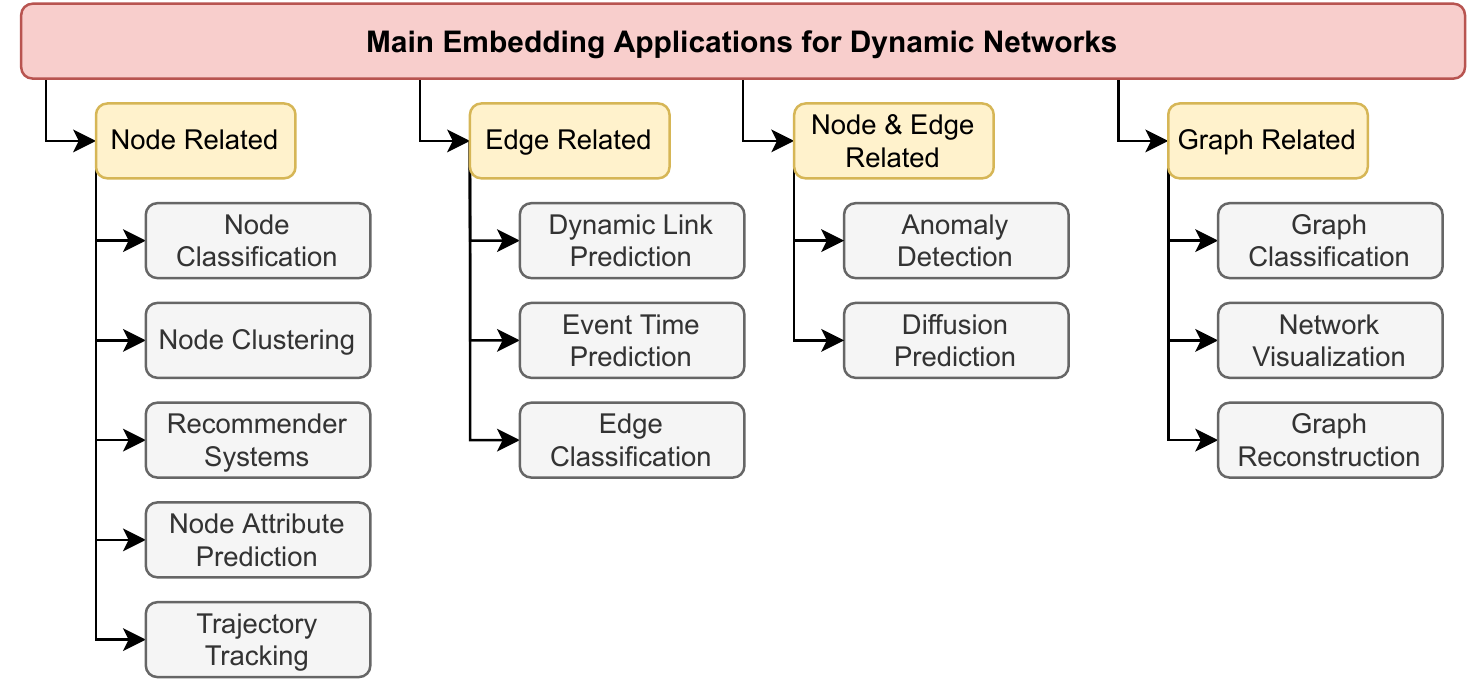}
    \caption{Every embedding application for dynamic networks covered in this survey, organized by which elements of the network they are oriented to or focused on.}
    \label{fig:applications}
\end{figure}

In this section, we provide an overview of different network applications that are typically improved by the embedding methods for dynamic graphs presented so far. Applications of dynamic graph embeddings for network mining can be divided into which elements of the network they are oriented to or focused on: (i)~\textbf{node related}, including node classification, recommendation systems, and trajectory analysis; (ii)~\textbf{edge related}, including link prediction and event time prediction; and (iii)~\textbf{graph related}, including graph classification over time, network visualization and diffusion prediction. The complete list of tasks discussed in this survey is shown in Figure~\ref{fig:applications}. 

\subsection{Node-Related Applications}

Node embeddings are used for various purposes on network analysis, with applications already known in static graphs, but handled over time, including \textbf{node classification}, \textbf{node clustering}, and \textbf{recommendation systems}, up to novel applications, specific within the dynamic scenario, such as the \textbf{node feature prediction} and \textbf{trajectory tracking}.

\begin{compactitem}
    \item \textbf{Node Classification:} The node classification problem focuses on the assignment of a class label to each node in a graph based on the rules learned from the labelled nodes. In a dynamic network, it is possible to (i)~classify nodes whose labels are unknown in a given timestamp~$t$, considering behavior and labels of the other nodes in the network; or (ii)~to predict the classification of a node in the future, given that node labels can vary over time~\cite{Rossi2018Relational}. Approaches concerning node classification apply a classifier on labeled node embeddings for training, including: (i)~linear layer~\cite{Yuan2017Temporal}; (ii)~SVM~\cite{Wei2019Lifelong, Taheri2019Learning, Yin2019DHNE, Pei2016Node, Yin2019mhdne, Rahman2016Link, Rahman2018dylink2vec}; (iii)~logistic regression~\cite{Li2017Attributed, Mahdavi2019Dynamic, Zuo2018Embedding, Lu2019Temporal, De2018Combining, Mahdavi2019Dynamic, Khoshraftar2019Dynamic, Liu2019Towards, Sajjad2019Efficient, Sato2019DyANE, Du2018Dynamic, Liu2019Real, Pei2016Node}; (iv)~softmax~\cite{Pareja2020EvolveGCN, Park2019Exploiting, Ma2020Streaming, Singer2019Node, Xu2019Spatio, Goyal2018Dyngem}; and (v)~random forests and gradient boosting techniques~\cite{De2018Combining, Rahman2016Link, Rahman2018dylink2vec}.

    \item \textbf{Node Clustering:} While the classification task is supervised, clustering similar nodes is an unsupervised task, aiming to group similar vertices when information about labels is unavailable. An important challenge in this task is to ensure that the embeddings of similar nodes are close to each other in the vector space while being able to capture possible node transitions between different clusters over time, as in DynGem~\cite{Goyal2018Dyngem} and dyngraph2vec variations~\cite{Goyal2019dyngraph2vec}. Clusters in the embedding space can also represent different behavior patterns of nodes over time, which is defined by Rossi et al.~\cite{Rossi2013Modeling} as an analysis of the role evolution of each node in the network.

    \item \textbf{Recommender Systems:} A dynamic network consisting of users, items, and timestamped interactions between users and items may be explored by embedding methods to recommend items to users according to their interests over time. As discussed by Kazemi et al.~a\cite{Kazemi2020Representation}, recommendations may suggest items not exactly similar to user's interest in order to attract the user to a novel interest, and recommendations may arouse a future desire for items of certain type even if the user does not display any immediate interest.

   \item \textbf{Node Attribute Prediction:} 
It is important to predict time-varying attributes of network nodes. Formally, given a time-varying graph $\mathcal{G} = \{G(t_{0}),...,G(t_{N_{S}-1})\}$ with additional node attributes $X(t) \in \mathbb{R}^{f}$ (where $f$ is the number of these attributes) as the training data, this task aims to estimate the real-valued variable $X(t_{N_{S}}) \in \mathbb{R}^{f}$ at time $t_{N_{S}}$ for each node in the graph~\cite{Rossi2018Relational}. This problem is also known as relational time series regression, and node embeddings over time can serve as input to time series prediction models, such as ARIMA or recurrent neural networks, to enhance prediction by leveraging topological information along with attribute evolution.

    \item \textbf{Trajectory Tracking:} From the representations obtained by embedding over time, it is possible to make an analysis of the trajectories of each entity. Ferreira et al.~\cite{Ferreira2019Modeling} capture ideological changes of two diverse party systems~(Brazil and the United States), as expressed by members’ voting behavior, by mapping the network into a temporal latent ideological space. Therefore, the authors have tracked individual members overtime in the low-dimensional vector space, analyzing how vector representations of individual members change and then measure ideological shifts over time. Although this task shares similarities with the node clustering problem, the focus of trajectory tracking is not on the groups themselves, but on the transitions between them.

\end{compactitem}

\subsection{Edge-Related Applications} \label{subsec:edgerelated}

Edge-related tasks comprise the most commonly explored problem by the techniques presented in this survey: \textbf{dynamic link prediction}. However, a novel application, when compared with static methods, appears in the literature: \textbf{event time prediction}, whose focus is on the detection of the time instant when a new edge should appear. Finally, the \textbf{edge classification} may also be treated on heterogeneous dynamic graphs where edges have labels.

\begin{compactitem}
    \item \textbf{Dynamic Link Prediction:} Link prediction in dynamic networks is more complex than in static networks, since it comprises two different tasks: (i)~\textbf{temporal link prediction}, (the prediction of new edges)~\cite{junuthula2016evaluating}, where, given a sequence of snapshots of an evolving network $\mathcal{G} = \{G(t_{0}), ..., G(t_{N_{S} -1})\}$, aims to predict the links in $G(t_{N_{S}})$ (where $N_{S}$ is the total number of snapshots), i.e. to construct a function $f(v,u)$ that predicts whether an edge $(v,u)$ exists between nodes $v$ and $u$ at time $t_{N_{S}}$~\cite{Saha2018Models}; and (ii)~\textbf{link completion} (prediction of previously observed edges)~\cite{junuthula2016evaluating}, which consists of finding the missing links along the evolving network. Most approaches consider link prediction as a classification task, where labels are (i)~existence; or (ii)~non-existence of an edge. Junuthula et al.~\cite{junuthula2016evaluating} provides a deeper discussion regarding evaluation of link prediction on dynamic networks, and Yu et al generalizes temporal link prediction to include: (i)~prediction of link weights, considering henceforth the aforementioned definition as a particular case when the link weight is restricted to 0 or 1, and (ii)~prediction at timestamp $N(S)+\alpha$, where $\alpha \ge 0$, and the classical definition above regards the special case $\alpha = 0$.

    \item \textbf{Event Time Prediction:} In dynamic networks it is valid to question at which timestamp a given interaction can occur, configuring the event time prediction task. Methods based on temporal point process have a mathematical formulation to predict the next time point $t_{k}$ for an event given a pair of nodes $v$ and $u$ and network history (i.e. a list of interactions over time), including DyRep~\cite{Trivedi2017Know} and DeepCoevolve~\cite{dai2016recurrent}. 

    \item \textbf{Edge Classification:} When the edges of a network can belong to certain classes, the problem of classifying the edges shares similarities with the node classification task. For instance, interactions between nodes may be associated with a trustworthy rating (i.e. users who trust others, or not), or sentiment analysis (i.e. a post on a social network). Therefore, predicting the label of an edge $(v_{i}, v_{j})$ over time may be done concatenating node embeddings or operating over them to obtain edge embeddings, and applying a classifier afterward~\cite{Pareja2020EvolveGCN}. 

\end{compactitem}

\subsection{Node- and Edge-Related Tasks} \label{subsec:nodeedgerelated}

Some tasks can be applied to both nodes and edges in a graph, depending on how the problem is formulated. In this context, two tasks have been identified so far: (i)~\textbf{anomaly detection}, which can be related to a node with anomalous behavior or to an unwanted or unexpected edge; and (ii)~\textbf{diffusion prediction}, which can either identify which nodes will be affected by a diffusion process or detect edges most likely to diffuse information.
\begin{compactitem}
    \item \textbf{Anomaly Detection:} Anomaly detection is an important application for detecting malicious activity in networks. These anomalies can be detected by unexpected changes in the vector representation of nodes and edges, suggesting that some non-expected activity is arising at some timestamp. Goyal et al.~\cite{Goyal2018Dyngem} propose the definition of $\delta_{t} = \norm{Z(t_{k+1}) - Z(t_{k})}_{F}$ as the change in embedding between time $t_{k}$ and $t_{k+1}$, and a threshold to consider the node behavior as anomalous when its embeddings change above this value. This threshold value can be calibrated diferently to each specific problem, and can be tuned to identify certain types of anomalies based on node behavior changes, where each type of behavior is encoded differently. A similar approach is presented by Rossi et al.~\cite{Rossi2013Modeling}, which defines several groups of behaviors for the nodes of the network based on node embeddings (i.e. similar to node clustering), hence the authors claim to detect anomalous behaviors from abrupt node transitions between different clusters.  Khoshraftar et al.~\cite{Khoshraftar2019Dynamic} formulate the anomaly detection as a classification task, where edges may belong to anomalous or normal class labels, therefore using node embeddings to compute edge representations and then using their method (LSTM-Node2vec) to classify the edges. 
    
    \item \textbf{Diffusion Prediction:} Diffusion problems solved by embeddings methods may be categorized as (i)~a sequence prediction problem (i.e. a microscopic diffusion problem), willing to predict the future affected node given the previously affected ones~\cite{wang2018attention, Yang2019Multi, Wang2017Topological,Lamprier2018Variational, Zhang2018Cosine}; or (ii)~a regression problem, which predicts the future numerical properties of the network (e.g. the total number of infected nodes in a macroscopic diffusion problem)~\cite{Li2017DeepCas,Yang2019Multi}.  

\end{compactitem}

\subsection{Graph-Related Applications}

Problems related to the whole graph usually analyze the network globally, therefore dealing with tasks that are not centered on vertices or edges. The main examples verified in the literature are: \textbf{graph classification}, \textbf{network visualization}, and \textbf{graph reconstruction}.
\begin{compactitem}
 
\item \textbf{Graph Classification:} Classifying the whole graph over time into one class from a set of predefined categories $\mathcal{L}^{G}$ is a relevant problem when the topological structure and possible attributes of each node in a network configure a global classifiable behavior~\cite{Taheri2019Predictive, Taheri2019Learning}. By obtaining a whole graph embedding over time (either by aggregating node embeddings, or using graph kernels, as in Section~\ref{subsec:graphkernels}), it is possible to use the same classifiers presented in the node classification task to either perform classifications at known timestamps or to predict the classification of the graph in the future.

\item \textbf{Network Visualization:} It is also possible to visualize dynamic graphs in 2D or 3D space by applying dimensionality reduction techniques that preserve the embedding structure, such as t-SNE~\cite{Maaten2008Visualizing}, to node embeddings. Goyal et al.~\cite{Goyal2018Dyngem} state that, to avoid visualization instability (i.e. embedding instability over time), t-SNE needs to be initialized with an identical random state for all timestamps. 

\item \textbf{Graph Reconstruction:} The learned vector representations may reconstruct the dynamic graph through operations in the vector space that decode similarity information between pairs of nodes, such as a dot product or pairwise distance to estimate the adjacency matrix or the weight matrix. Goyal et al.~\cite{Goyal2018Dyngem} propose a methodology to rank pair of nodes according to their corresponding reconstructed similarity, then define the reconstruction precision as the ratio of real links in the top $k$ pair of nodes. 

\end{compactitem}

\section{Conclusion and Future Directions}
\label{sec:conc}

In this survey, we conducted a comprehensive review of the literature in embedding methods for dynamic graphs. We defined the problem of embeddings for dynamic graphs inspired by previous surveys on static graph embeddings, whereas introducing some important concepts to consider scenarios for dynamic graphs. We proposed a taxonomy to classify the problem settings for dynamic graph embedding problems, broadening the design presented in the static scenario by introducing fundamental time aspects in embeddings, such as the different dynamic graph models that can be embedded, in addition to embedding outputs that aggregate temporal information or track representation trajectories. We also proposed a taxonomy for the different embedding paradigms for dynamic graphs, classifying them according to the methodology they use, topological-temporal properties that preserve, and assumptions made for the method to be valid. After that, we summarized the applications that the embedding of dynamic graphs enables.

\begin{compactitem}
    \item \textbf{Development and Expansion of Libraries and Frameworks for Dynamic Graph Embedding:} With the increasing number of dynamic graph embedding techniques, it is interesting to invest in developing and expanding a framework capable of unifying the different algorithms, applications, and standard benchmark datasets. Goyal et al.~\cite{Goyal2018DynamicGEM} propose DynamicGEM, an open-source Python library consisting of state-of-the-art algorithms for dynamic graph embeddings, focusing on node embeddings. The library contains the evaluation framework for graph reconstruction, static and temporal link prediction, node classification, and temporal visualization, with various metrics implemented to evaluate the state-of-the-art methods, and examples of evolving networks from various domains. Since DynamicGEM provides a template to add new algorithms, it would be interesting to invest in further developing the package so as to incrementally insert new techniques. In addition, expanding the framework to include embeddings of other types (such as edges, hybrids, and whole graphs), as well as methods for continuous-time dynamic graphs. Building a reference package of embedding methods for dynamic graphs~(be it DynamicGEM or other) would benefit the community interested in this topic. 
    
    \item \textbf{Mathematical Analysis and Different Stability Metrics:} To define useful and accurate stability metrics is a future research field in the area, with a more theoretical focus. Goyal et al.~\cite{Goyal2018Dyngem} suggest a metric considering the adjacency matrix and snapshot models. Nevertheless, it is necessary to explore these metrics in more detail, by testing them on real-world networks and checking the behavior of embeddings for different methods and problems. In addition, a more sophisticated mathematical analysis may promote a research field direction to improve understanding of the relationship between representations and the evolution of dynamic graphs. 

    \item \textbf{Temporal Multiscale Evolution Embedding:} In real networks, the phenomena of temporal evolution may be associated with different scales (e.g. daily, weekly, monthly, and yearly phenomena). It would be interesting to investigate embedding techniques capable of efficiently capturing these peculiarities in a methodology that deals with temporal multiscale evolution. Trivedi et al.~\cite{Trivedi2018Representation} make an important step forward in this direction by considering two different timescales: for dynamics on the network and dynamics of the network.
    
    \item \textbf{Dynamic Hypergraph Embedding:} The existing models for dynamic graphs only consider edges connecting two nodes in the graph, therefore not being able to expand to hypergraphs, where sets of nodes form connections without necessarily having binary ordered relations. Although there are a few works in hypergraph embedding \cite{Feng2019Hypergraph} and even fewer that are extended to dynamic hypergraphs \cite{Zhang2018DynamicHypergraph}, a promising area of research is to develop new methods for dynamic hypergraphs, as well as extending some of the existing dynamic graph embedding methods, allowing them to handle hyperedges.
    
    \item \textbf{Capturing Latencies and Spatial-Temporal Edge Patterns:} Although the methods described in the survey capture dynamic behaviors such as topological evolution, feature evolution and processes on the network, more sophisticated temporal models take into account that nodes and edges are not created or removed instantaneously in the network~\cite{Casteigts2012Time}. Thus, latency is an important feature of dynamic networks, as it carries information about the affinity between nodes or a node within the network, and no method described above has a methodology for dealing with such factors. In addition, spatio-temporal edges make embedding of dynamic graphs more complex than extracting information from snapshots, or from timestamped edges, as temporal correlations are more complex between non-consecutive timestamps~\cite{Wehmuth2015Unifying}.
    
    \item \textbf{Generalization of Graph Embedding to Higher-Order Dimensional Networks:} With the diversity of models for dynamic graphs, it is noted that there is no consensus to define a more general model and, consequently, an embedding that can be generalized to as many dynamic graphs as possible. Keeping this in mind, there are graph generalization models, including MAGs~\cite{Wehmuth2015Unifying} and Stream Graphs \cite{Latapy2018}, and a future direction of research may be the application of embedding methods in such models. Even more, it is interesting to suggest extending embedding for higher-order graphs, allowing not only the capture of temporal and topological properties but also of multilayer structures at a different time and connectivity scales.
    
    \item \textbf{Generation of Property-Preserving Network Evolution in Embedding Space:} Several complex network properties, such as pathways, degree distribution, and scale invariance, may change over time, and finding out which patterns of temporal evolution conserve or change these properties is challenging. Cheng et al.~\cite{Cheng2016Graph} propose a structure-preserving model reduction procedure  developed for linear network systems, whereas Rossi et al.~\cite{Rossi2013Modeling} propose a matrix factorization to discover roles of certain vertices in a network and study possible changes in these roles over time. One possible direction for future research is to use representations in low-dimensional spaces to study and generate evolutions in a network that are capable of preserving certain properties of interest. One possible way to pursue this idea is to explore generative models, such as variational autoencoders and GANs, and use learned distributions of input data to generate new networks that are similar to training, and thereby capture patterns associated with their characteristics.

\end{compactitem}

\section*{Acknowledgment}

This work has been partially supported by CAPES, CNPq, FAPEMIG, and FAPERJ. Moreover, this paper is dedicated to the memory of our dear co-worker Artur Ziviani, who passed away while this paper was being peer-reviewed. Artur was a brilliant researcher and dedicated advisor.

\bibliographystyle{unsrt}  
\bibliography{reference}

\begin{thebibliography}{100}

\bibitem{Barabasi2016Network}
Albert-L{\'a}szl{\'o} Barab{\'a}si et~al.
\newblock {\em Network {S}cience}.
\newblock Cambridge University Press, 2016.

\bibitem{Hamilton2017Representation}
William~L Hamilton, Rex Ying, and Jure Leskovec.
\newblock Representation {L}earning on {G}raphs: Methods and {A}pplications.
\newblock {\em IEEE Data Engineering Bulletin}, 2017.

\bibitem{Cai2018Comprehensive}
Hongyun Cai, Vincent~W Zheng, and Kevin Chen-Chuan Chang.
\newblock A {C}omprehensive {S}urvey of {G}raph {E}mbedding: Problems,
  {T}echniques, and {A}pplications.
\newblock {\em IEEE Trans Knowl Data Eng}, 30(9):1616--1637, 2018.

\bibitem{Goyal2018Graph}
Palash Goyal and Emilio Ferrara.
\newblock Graph {E}mbedding {T}echniques, {A}pplications, and {P}erformance: A
  {S}urvey.
\newblock {\em Knowledge-Based Systems}, 151:78--94, 2018.

\bibitem{Dunlavy2011Temporal}
Daniel~M Dunlavy, Tamara~G Kolda, and Evrim Acar.
\newblock Temporal {L}ink {P}rediction using {M}atrix and {T}ensor
  {F}actorizations.
\newblock {\em ACM Transactions on Knowledge Discovery from Data (TKDD)},
  5(2):10, 2011.

\bibitem{Casteigts2012Time}
Arnaud Casteigts, Paola Flocchini, Walter Quattrociocchi, and Nicola Santoro.
\newblock Time-{V}arying {G}raphs and {D}ynamic {N}etworks.
\newblock {\em International Journal of Parallel, Emergent and Distributed
  Systems}, 27(5):387--408, 2012.

\bibitem{Li2017Attributed}
Jundong Li, Harsh Dani, Xia Hu, Jiliang Tang, Yi~Chang, and Huan Liu.
\newblock Attributed {N}etwork {E}mbedding for {L}earning in a {D}ynamic
  {E}nvironment.
\newblock In {\em Proceedings of the 2017 ACM on Conference on Information and
  Knowledge Management}, pages 387--396. ACM, 2017.

\bibitem{Trivedi2018Representation}
Rakshit Trivedi, Mehrdad Farajtabar, Prasenjeet Biswal, and Hongyuan Zha.
\newblock Dy{R}ep: Learning {R}epresentations over {D}ynamic {G}raphs.
\newblock In {\em International Conference on Learning Representations}, 2019.

\bibitem{Zhu2016Scalable}
Linhong Zhu, Dong Guo, Junming Yin, Greg Ver~Steeg, and Aram Galstyan.
\newblock Scalable {T}emporal {L}atent {S}pace {I}nference for {L}ink
  {P}rediction in {D}ynamic {S}ocial {N}etworks.
\newblock {\em IEEE Transactions on Knowledge and Data Engineering},
  28(10):2765--2777, 2016.

\bibitem{Goyal2018Dyngem}
Palash Goyal, Nitin Kamra, Xinran He, and Yan Liu.
\newblock Dyn{G}{E}{M}: Deep {E}mbedding {M}ethod for {D}ynamic {G}raphs.
\newblock {\em arXiv Preprint arXiv:1805.11273}, 2018.

\bibitem{Li2017DeepCas}
Cheng Li, Jiaqi Ma, Xiaoxiao Guo, and Qiaozhu Mei.
\newblock Deep{C}as: An {E}nd-to-end {P}redictor of {I}nformation {C}ascades.
\newblock In {\em Proceedings of the 26th International Conference on World
  Wide Web}, pages 577--586. International World Wide Web Conferences Steering
  Committee, 2017.

\bibitem{Kazemi2020Representation}
Seyed~Mehran Kazemi, Rishab Goel, Kshitij Jain, Ivan Kobyzev, Akshay Sethi,
  Peter Forsyth, and Pascal Poupart.
\newblock Representation {L}earning for {D}ynamic {G}raphs: A {S}urvey.
\newblock {\em Journal of Machine Learning Research}, 21(70):1--73, 2020.

\bibitem{Xie2020Survey}
Yu~Xie, Chunyi Li, Bin Yu, Chen Zhang, and Zhouhua Tang.
\newblock A {S}urvey on {D}ynamic {N}etwork {E}mbedding.
\newblock {\em arXiv preprint arXiv:2006.08093}, 2020.

\bibitem{Skarding2021Foundations}
Joakim Skarding, Bogdan Gabrys, and Katarzyna Musial.
\newblock Foundations and {M}odelling of {D}ynamic {N}etworks using {D}ynamic
  {G}raph {N}eural {N}etworks: A {S}urvey.
\newblock {\em IEEE Access}, 2021.

\bibitem{Katz1953New}
Leo Katz.
\newblock A {N}ew {S}tatus {I}ndex {D}erived from {S}ociometric {A}nalysis.
\newblock {\em Psychometrika}, 18(1):39--43, 1953.

\bibitem{Adamic2003Friends}
Lada~A Adamic and Eytan Adar.
\newblock Friends and {N}eighbors on the {W}eb.
\newblock {\em Social networks}, 25(3):211--230, 2003.

\bibitem{Cui2018Survey}
Peng Cui, Xiao Wang, Jian Pei, and Wenwu Zhu.
\newblock A {S}urvey on {N}etwork {E}mbedding.
\newblock {\em IEEE Transactions on Knowledge and Data Engineering}, 2018.

\bibitem{Zhang2018Network}
Daokun Zhang, Jie Yin, Xingquan Zhu, and Chengqi Zhang.
\newblock Network {R}epresentation {L}earning: A {S}urvey.
\newblock {\em IEEE Transactions on Big Data}, 2018.

\bibitem{Wang2017Knowledge}
Quan Wang, Zhendong Mao, Bin Wang, and Li~Guo.
\newblock Knowledge {G}raph {E}mbedding: A {S}urvey of {A}pproaches and
  {A}pplications.
\newblock {\em IEEE Transactions on Knowledge and Data Engineering},
  29(12):2724--2743, 2017.

\bibitem{Nickel2015Review}
Maximilian Nickel, Kevin Murphy, Volker Tresp, and Evgeniy Gabrilovich.
\newblock A {R}eview of {R}elational {M}achine {L}earning for {K}nowledge
  {G}raphs.
\newblock {\em Proceedings of the IEEE}, 104(1):11--33, 2015.

\bibitem{grattarola2020graph}
Daniele Grattarola and Cesare Alippi.
\newblock Graph neural networks in tensorflow and keras with spektral.
\newblock {\em arXiv preprint arXiv:2006.12138}, 2020.

\bibitem{Wehmuth2015Unifying}
Klaus Wehmuth, Artur Ziviani, and Eric Fleury.
\newblock A {U}nifying {M}odel for {R}epresenting {T}ime-{V}arying {G}raphs.
\newblock In {\em 2015 IEEE International Conference on Data Science and
  Advanced Analytics (DSAA)}, pages 1--10. IEEE, 2015.

\bibitem{Latapy2018}
Matthieu Latapy, Tiphaine Viard, and Cl\'{e}mence Magnien.
\newblock Stream graphs and link streams for the modeling of interactions over
  time.
\newblock {\em Social Network Analysis and Mining (SNAM)}, 8(61), December
  2018.

\bibitem{dai2016recurrent}
Hanjun Dai, Yichen Wang, Rakshit Trivedi, and Le~Song.
\newblock Recurrent coevolutionary latent feature processes for continuous-time
  recommendation.
\newblock In {\em Proceedings of the 1st Workshop on Deep Learning for
  Recommender Systems}, pages 29--34, 2016.

\bibitem{Chen2019Dynamic}
Chuanchang Chen, Yubo Tao, and Hai Lin.
\newblock Dynamic {N}etwork {E}mbeddings for {N}etwork {E}volution {A}nalysis.
\newblock {\em arXiv preprint arXiv:1906.09860}, 2019.

\bibitem{Zhou2018Dynamic}
Lekui Zhou, Yang Yang, Xiang Ren, Fei Wu, and Yueting Zhuang.
\newblock Dynamic {N}etwork {E}mbedding by {M}odeling {T}riadic {C}losure
  {P}rocess.
\newblock In {\em Thirty-Second AAAI Conference on Artificial Intelligence},
  2018.

\bibitem{Goyal2019dyngraph2vec}
Palash Goyal, Sujit~Rokka Chhetri, and Arquimedes Canedo.
\newblock dyngraph2vec: Capturing {N}etwork {D}ynamics using {D}ynamic {G}raph
  {R}epresentation {L}earning.
\newblock {\em Knowledge-Based Systems}, 2019.

\bibitem{Ferreira2019Modeling}
Carlos Henrique~Gomes Ferreira, Fabricio~Murai Ferreira, Breno de~Sousa~Matos,
  and Jussara~Marques de~Almeida.
\newblock Modeling {D}ynamic {I}deological {B}ehavior in {P}olitical
  {N}etworks.
\newblock {\em The Journal of Web Science}, 7, 2019.

\bibitem{Mitrovic2019dyn2vec}
Sandra Mitrovic and Jochen De~Weerdt.
\newblock Dyn2vec: Exploiting {D}ynamic {B}ehaviour using {D}ifference
  {N}etworks-{B}ased {N}ode {E}mbeddings for {C}lassification.
\newblock In {\em Proceedings of the International Conference on Data Science},
  pages 194--200. CSREA Press, 2019.

\bibitem{Du2018Dynamic}
Lun Du, Yun Wang, Guojie Song, Zhicong Lu, and Junshan Wang.
\newblock Dynamic {N}etwork {E}mbedding: An {E}xtended {A}pproach for
  {S}kip-gram based {N}etwork {E}mbedding.
\newblock In {\em IJCAI}, pages 2086--2092, 2018.

\bibitem{Mahdavi2019Dynamic}
Sedigheh Mahdavi, Shima Khoshraftar, and Aijun An.
\newblock Dynamic {J}oint {V}ariational {G}raph {A}utoencoders.
\newblock In {\em Joint European Conference on Machine Learning and Knowledge
  Discovery in Databases}, pages 385--401. Springer, 2019.

\bibitem{Zhang2018Timers}
Ziwei Zhang, Peng Cui, Jian Pei, Xiao Wang, and Wenwu Zhu.
\newblock {T}{I}{M}{E}{R}{S}: Error-{B}ounded {S}{V}{D} {R}estart on {D}ynamic
  {N}etworks.
\newblock In {\em Thirty-Second AAAI Conference on Artificial Intelligence},
  2018.

\bibitem{Hisano2018Semi}
Ryohei Hisano.
\newblock Semi-{S}upervised {G}raph {E}mbedding {A}pproach to {D}ynamic {L}ink
  {P}rediction.
\newblock In {\em International Workshop on Complex Networks}, pages 109--121.
  Springer, 2018.

\bibitem{Nguyen2018Continuous}
Giang~Hoang Nguyen, John~Boaz Lee, Ryan~A Rossi, Nesreen~K Ahmed, Eunyee Koh,
  and Sungchul Kim.
\newblock Continuous-{T}ime {D}ynamic {N}etwork {E}mbeddings.
\newblock In {\em Companion Proceedings of the The Web Conference 2018}, pages
  969--976. International World Wide Web Conferences Steering Committee, 2018.

\bibitem{Torricelli2020weg2vec}
Maddalena Torricelli, M{\'a}rton Karsai, and Laetitia Gauvin.
\newblock weg2vec: Event {E}mbedding for {T}emporal {N}etworks.
\newblock {\em Scientific Reports}, 10(1):1--11, 2020.

\bibitem{Liu2019Real}
Xi~Liu, Ping-Chun Hsieh, Nick Duffield, Rui Chen, Muhe Xie, and Xidao Wen.
\newblock Real-{T}ime {S}treaming {G}raph {E}mbedding {T}hrough {L}ocal
  {A}ctions.
\newblock In {\em Companion Proceedings of The 2019 World Wide Web Conference},
  pages 285--293. ACM, 2019.

\bibitem{Yan2019Modeling}
Junchi Yan, Hongteng Xu, and Liangda Li.
\newblock Modeling and {A}pplications for {T}emporal {P}oint {P}rocesses.
\newblock In {\em Proceedings of the 25th ACM SIGKDD International Conference
  on Knowledge Discovery \& Data Mining}, pages 3227--3228. ACM, 2019.

\bibitem{Trivedi2017Know}
Rakshit Trivedi, Hanjun Dai, Yichen Wang, and Le~Song.
\newblock Know-{E}volve: Deep {T}emporal {R}easoning for {D}ynamic {K}nowledge
  {G}raphs.
\newblock In {\em Proceedings of the 34th International Conference on Machine
  Learning-Volume 70}, pages 3462--3471. JMLR. org, 2017.

\bibitem{Zuo2018Embedding}
Yuan Zuo, Guannan Liu, Hao Lin, Jia Guo, Xiaoqian Hu, and Junjie Wu.
\newblock Embedding {T}emporal {N}etwork via {N}eighborhood {F}ormation.
\newblock In {\em Proceedings of the 24th ACM SIGKDD International Conference
  on Knowledge Discovery \& Data Mining}, pages 2857--2866. ACM, 2018.

\bibitem{Knyazev2019Learning}
Boris Knyazev, Carolyn Augusta, and Graham~W Taylor.
\newblock Learning {T}emporal {A}ttention in {D}ynamic {G}raphs with {B}ilinear
  {I}nteractions.
\newblock {\em arXiv preprint arXiv:1909.10367}, 2019.

\bibitem{Dai2016Deep}
Hanjun Dai, Yichen Wang, Rakshit Trivedi, and Le~Song.
\newblock Deep {C}oevolutionary {N}etwork: Embedding {U}ser and {I}tem
  {F}eatures for {R}ecommendation.
\newblock {\em arXiv preprint arXiv:1609.03675}, 2016.

\bibitem{Wu2020Modeling}
Weichang Wu, Huanxi Liu, Xiaohu Zhang, Yu~Liu, and Hongyuan Zha.
\newblock Modeling {E}vent {P}ropagation via {G}raph {B}iased {T}emporal
  {P}oint {P}rocess.
\newblock {\em IEEE Transactions on Neural Networks and Learning Systems},
  2020.

\bibitem{Saha2018Models}
Tanay~Kumar Saha, Thomas Williams, Mohammad~Al Hasan, Shafiq Joty, and
  Nicholas~K Varberg.
\newblock Models for {C}apturing {T}emporal {S}moothness in {E}volving
  {N}etworks for {L}earning {L}atent {R}epresentation of {N}odes.
\newblock {\em arXiv preprint arXiv:1804.05816}, 2018.

\bibitem{Lei2019GCNGAN}
Kai Lei, Meng Qin, Bo~Bai, Gong Zhang, and Min Yang.
\newblock G{C}{N}-{G}{A}{N}: A {N}on-{L}inear {T}emporal {L}ink {P}rediction
  {M}odel for {W}eighted {D}ynamic {N}etworks.
\newblock In {\em IEEE INFOCOM 2019-IEEE Conference on Computer
  Communications}, pages 388--396. IEEE, 2019.

\bibitem{Wu2019T}
Jiajing Wu, Dan Lin, Zibin Zheng, and Qi~Yuan.
\newblock T-{E}{D}{G}{E}: Temporal w{E}ighted {M}ulti{D}i{G}raph {E}mbedding
  for {E}thereum {T}ransaction {N}etwork {A}nalysis.
\newblock {\em arXiv preprint arXiv:1905.08038}, 2019.

\bibitem{Yin2019DHNE}
Ying Yin, Li-Xin Ji, Jian-Peng Zhang, and Yu-Long Pei.
\newblock D{H}{N}{E}: Network {R}epresentation {L}earning {M}ethod for
  {D}ynamic {H}eterogeneous {N}etworks.
\newblock {\em IEEE Access}, 7:134782--134792, 2019.

\bibitem{Pareja2020EvolveGCN}
Aldo Pareja, Giacomo Domeniconi, Jie Chen, Tengfei Ma, Toyotaro Suzumura,
  Hiroki Kanezashi, Tim Kaler, Tao~B. Schardl, and Charles~E. Leiserson.
\newblock {EvolveGCN}: Evolving graph convolutional networks for dynamic
  graphs.
\newblock In {\em Proceedings of the Thirty-Fourth AAAI Conference on
  Artificial Intelligence}, 2020.

\bibitem{Taheri2019Learning}
Aynaz Taheri, Kevin Gimpel, and Tanya Berger-Wolf.
\newblock Learning to {R}epresent the {E}volution of {D}ynamic {G}raphs with
  {R}ecurrent {M}odels.
\newblock In {\em Companion Proceedings of The 2019 World Wide Web Conference},
  pages 301--307, 2019.

\bibitem{Taheri2019Predictive}
Aynaz Taheri and Tanya Berger-Wolf.
\newblock Predictive {T}emporal {E}mbedding of {D}ynamic {G}raphs.
\newblock In {\em Proceedings of the 2019 IEEE/ACM International Conference on
  Advances in Social Networks Analysis and Mining}, pages 57--64, 2019.

\bibitem{Wei2019Lifelong}
Hao Wei, Guyu Hu, Wei Bai, Shiming Xia, and Zhisong Pan.
\newblock Lifelong {R}epresentation {L}earning in {D}ynamic {A}ttributed
  {N}etworks.
\newblock {\em Neurocomputing}, 358:1--9, 2019.

\bibitem{Yu2017Spatio}
Bing Yu, Haoteng Yin, and Zhanxing Zhu.
\newblock Spatio-{T}emporal {G}raph {C}onvolutional {N}etworks: A {D}eep
  {L}earning {F}ramework for {T}raffic {F}orecasting.
\newblock {\em arXiv preprint arXiv:1709.04875}, 2017.

\bibitem{Li2019Learning}
Youru Li, Zhenfeng Zhu, Deqiang Kong, Meixiang Xu, and Yao Zhao.
\newblock Learning {H}eterogeneous {S}patial-{T}emporal {R}epresentation for
  {B}ike-{S}haring {D}emand {P}rediction.
\newblock In {\em Proceedings of the AAAI Conference on Artificial
  Intelligence}, volume~33, pages 1004--1011, 2019.

\bibitem{Deng2019Learning}
Songgaojun Deng, Huzefa Rangwala, and Yue Ning.
\newblock Learning {D}ynamic {C}ontext {G}raphs for {P}redicting {S}ocial
  {E}vents.
\newblock In {\em Proceedings of the 25th ACM SIGKDD International Conference
  on Knowledge Discovery \& Data Mining}, pages 1007--1016. ACM, 2019.

\bibitem{Yuan2017Temporal}
Yuan Yuan, Xiaodan Liang, Xiaolong Wang, Dit-Yan Yeung, and Abhinav Gupta.
\newblock Temporal {D}ynamic {G}raph {L}{S}{T}{M} for {A}ction-{D}riven {V}ideo
  {O}bject {D}etection.
\newblock In {\em Proceedings of the IEEE International Conference on Computer
  Vision}, pages 1801--1810, 2017.

\bibitem{Rossi2013Modeling}
Ryan~A Rossi, Brian Gallagher, Jennifer Neville, and Keith Henderson.
\newblock Modeling {D}ynamic {B}ehavior in {L}arge {E}volving {G}raphs.
\newblock In {\em Proceedings of the Sixth ACM International Conference on Web
  Search and Data Mining}, pages 667--676. ACM, 2013.

\bibitem{Goel2019Diachronic}
Rishab Goel, Seyed~Mehran Kazemi, Marcus Brubaker, and Pascal Poupart.
\newblock Diachronic {E}mbedding for {T}emporal {K}nowledge {G}raph
  {C}ompletion.
\newblock {\em arXiv preprint arXiv:1907.03143}, 2019.

\bibitem{Meng2018Subgraph}
Changping Meng, S~Chandra Mouli, Bruno Ribeiro, and Jennifer Neville.
\newblock Subgraph {P}attern {N}eural {N}etworks for {H}igh-{O}rder {G}raph
  {E}volution {P}rediction.
\newblock In {\em Thirty-Second AAAI Conference on Artificial Intelligence},
  2018.

\bibitem{Rahman2016Link}
Mahmudur Rahman and Mohammad Al~Hasan.
\newblock Link {P}rediction in {D}ynamic {N}etworks using {G}raphlet.
\newblock In {\em Joint European Conference on Machine Learning and Knowledge
  Discovery in Databases}, pages 394--409. Springer, 2016.

\bibitem{Dave2019Triangle}
V~Dave and M~Hasan.
\newblock Triangle {C}ompletion {T}ime {P}rediction using {T}ime-{C}onserving
  {E}mbedding, 2019.

\bibitem{Yu2017Temporally}
Wenchao Yu, Charu~C Aggarwal, and Wei Wang.
\newblock Temporally {F}actorized {N}etwork {M}odeling for {E}volutionary
  {N}etwork {A}nalysis.
\newblock In {\em Proceedings of the Tenth ACM International Conference on Web
  Search and Data Mining}, pages 455--464. ACM, 2017.

\bibitem{kempe2003maximizing}
David Kempe, Jon Kleinberg, and {\'E}va Tardos.
\newblock Maximizing the spread of influence through a social network.
\newblock In {\em Proceedings of the ninth ACM SIGKDD international conference
  on Knowledge discovery and data mining}, pages 137--146, 2003.

\bibitem{Zhou2019Dynamic}
Yujing Zhou, Weile Liu, Yang Pei, Lei Wang, Daren Zha, and Tianshu Fu.
\newblock Dynamic {N}etwork {E}mbedding by {S}emantic {E}volution.
\newblock In {\em 2019 International Joint Conference on Neural Networks
  (IJCNN)}, pages 1--8. IEEE, 2019.

\bibitem{Sankar2018Dynamic}
Aravind Sankar, Yanhong Wu, Liang Gou, Wei Zhang, and Hao Yang.
\newblock Dynamic {G}raph {R}epresentation {L}earning via {S}elf-{A}ttention
  {N}etworks.
\newblock {\em arXiv preprint arXiv:1812.09430}, 2018.

\bibitem{Yu2017Link}
Wenchao Yu, Wei Cheng, Charu~C Aggarwal, Haifeng Chen, and Wei Wang.
\newblock Link {P}rediction with {S}patial and {T}emporal {C}onsistency in
  {D}ynamic {N}etworks.
\newblock In {\em IJCAI}, pages 3343--3349, 2017.

\bibitem{Stewart90MatrixPerturbation}
G.~W. Stewart.
\newblock Matrix {P}erturbation {T}heory, 1990.

\bibitem{Acar2008Unsupervised}
Evrim Acar and B{\"u}lent Yener.
\newblock Unsupervised {M}ultiway {D}ata {A}nalysis: A {L}iterature {S}urvey.
\newblock {\em IEEE Trans Knowl Data Eng}, 21(1):6--20, 2008.

\bibitem{Rafailidis2014Modeling}
Dimitrios Rafailidis and Alexandros Nanopoulos.
\newblock Modeling the {D}ynamics of {U}ser {P}references in {C}oupled {T}ensor
  {F}actorization.
\newblock In {\em Proceedings of the 8th ACM Conference on Recommender
  systems}, pages 321--324. ACM, 2014.

\bibitem{Fang2015Personalized}
Xiaomin Fang, Rong Pan, Guoxiang Cao, Xiuqiang He, and Wenyuan Dai.
\newblock Personalized {T}ag {R}ecommendation through {N}onlinear {T}ensor
  {F}actorization using {G}aussian {K}ernel.
\newblock In {\em Twenty-Ninth AAAI Conference on Artificial Intelligence},
  2015.

\bibitem{Scarselli2008Graph}
Franco Scarselli, Marco Gori, Ah~Chung Tsoi, Markus Hagenbuchner, and Gabriele
  Monfardini.
\newblock The {G}raph {N}eural {N}etwork {M}odel.
\newblock {\em IEEE Transactions on Neural Networks}, 20(1):61--80, 2008.

\bibitem{Niepert2016Learning}
Mathias Niepert, Mohamed Ahmed, and Konstantin Kutzkov.
\newblock Learning {C}onvolutional {N}eural {N}etworks for {G}raphs.
\newblock In {\em International Conference on Machine Learning}, pages
  2014--2023, 2016.

\bibitem{Wang2016Structural}
Daixin Wang, Peng Cui, and Wenwu Zhu.
\newblock Structural {D}eep {N}etwork {E}mbedding.
\newblock In {\em Proceedings of the 22nd ACM SIGKDD International Conference
  on Knowledge Discovery and Data Mining}, pages 1225--1234. ACM, 2016.

\bibitem{Chen2018GC}
Jinyin Chen, Xuanheng Xu, Yangyang Wu, and Haibin Zheng.
\newblock G{C}-{L}{S}{T}{M}: Graph {C}onvolution {E}mbedded {L}{S}{T}{M} for
  {D}ynamic {L}ink {P}rediction.
\newblock {\em arXiv preprint arXiv:1812.04206}, 2018.

\bibitem{Hochreiter1997Long}
Sepp Hochreiter and J{\"u}rgen Schmidhuber.
\newblock Long {S}hort-{T}erm {M}emory.
\newblock {\em Neural Computation}, 9(8):1735--1780, 1997.

\bibitem{Cho2014Learning}
Kyunghyun Cho, Bart Van~Merri{\"e}nboer, Caglar Gulcehre, Dzmitry Bahdanau,
  Fethi Bougares, Holger Schwenk, and Yoshua Bengio.
\newblock Learning {P}hrase {R}epresentations using {R}{N}{N}
  {E}ncoder-{D}ecoder for {S}tatistical {M}achine {T}ranslation.
\newblock {\em arXiv preprint arXiv:1406.1078}, 2014.

\bibitem{Chen2019lstm}
Jinyin Chen, Jian Zhang, Xuanheng Xu, Chenbo Fu, Dan Zhang, Qingpeng Zhang, and
  Qi~Xuan.
\newblock E-{L}{S}{T}{M}-{D}: A {D}eep {L}earning {F}ramework for {D}ynamic
  {N}etwork {L}ink {P}rediction.
\newblock {\em IEEE Transactions on Systems, Man, and Cybernetics: Systems},
  2019.

\bibitem{Li2015Gated}
Yujia Li, Daniel Tarlow, Marc Brockschmidt, and Richard Zemel.
\newblock Gated {G}raph {S}equence {N}eural {N}etworks.
\newblock {\em arXiv preprint arXiv:1511.05493}, 2015.

\bibitem{Bahdanau2014Neural}
Dzmitry Bahdanau, Kyunghyun Cho, and Yoshua Bengio.
\newblock Neural {M}achine {T}ranslation by {J}ointly {L}earning to {A}lign and
  {T}ranslate.
\newblock {\em arXiv preprint arXiv:1409.0473}, 2014.

\bibitem{Xu2019Spatio}
Dongkuan Xu, Wei Cheng, Dongsheng Luo, Xiao Liu, and Xiang Zhang.
\newblock Spatio-{T}emporal {A}ttentive {R}{N}{N} for {N}ode {C}lassification
  in {T}emporal {A}ttributed {G}raphs.
\newblock In {\em Proceedings of the 28th International Joint Conference on
  Artificial Intelligence}, pages 3947--3953. AAAI Press, 2019.

\bibitem{Krizhevsky2012Imagenet}
Alex Krizhevsky, Ilya Sutskever, and Geoffrey~E Hinton.
\newblock Image{N}et {C}lassification with {D}eep {C}onvolutional {N}eural
  {N}etworks.
\newblock In {\em Advances in Neural Information Processing Systems}, pages
  1097--1105, 2012.

\bibitem{Kipf2016Semi}
Thomas~N Kipf and Max Welling.
\newblock Semi-{S}upervised {C}lassification with {G}raph {C}onvolutional
  {N}etworks.
\newblock {\em arXiv preprint arXiv:1609.02907}, 2016.

\bibitem{Hamilton2017Inductive}
Will Hamilton, Zhitao Ying, and Jure Leskovec.
\newblock Inductive {R}epresentation {L}earning on {L}arge {G}raphs.
\newblock In {\em Advances in Neural Information Processing Systems}, pages
  1024--1034, 2017.

\bibitem{Gao2019dyngraph2seq}
Yuyang Gao, Lingfei Wu, Houman Homayoun, and Liang Zhao.
\newblock Dyn{G}raph2{S}eq: Dynamic-{G}raph-to-{S}equence {I}nterpretable
  {L}earning for {H}ealth {S}tage {P}rediction in {O}nline {H}ealth {F}orums.
\newblock In {\em IEEE ICDM}, pages 1042--1047, 2019.

\bibitem{Bonner2018Temporal}
Stephen Bonner, John Brennan, Ibad Kureshi, Georgios Theodoropoulos,
  Andrew~Stephen McGough, and Boguslaw Obara.
\newblock Temporal {G}raph {O}ffset {R}econstruction: Towards {T}emporally
  {R}obust {G}raph {R}epresentation {L}earning.
\newblock In {\em 2018 IEEE International Conference on Big Data (Big Data)},
  pages 3737--3746. IEEE, 2018.

\bibitem{Zhao2019T}
Ling Zhao, Yujiao Song, Chao Zhang, Yu~Liu, Pu~Wang, Tao Lin, Min Deng, and
  Haifeng Li.
\newblock T-{G}{C}{N}: A {T}emporal {G}raph {C}onvolutional {N}etwork for
  {T}raffic {P}rediction.
\newblock {\em IEEE Transactions on Intelligent Transportation Systems}, 2019.

\bibitem{Xiong2019Dyngraphgan}
Yun Xiong, Yao Zhang, Hanjie Fu, Wei Wang, Yangyong Zhu, and S~Yu Philip.
\newblock Dyn{G}raph{G}{A}{N}: Dynamic {G}raph {E}mbedding via {G}enerative
  {A}dversarial {N}etworks.
\newblock In {\em International Conference on Database Systems for Advanced
  Applications}, pages 536--552. Springer, 2019.

\bibitem{Kingma2013Auto}
Diederik~P Kingma and Max Welling.
\newblock Auto-{E}ncoding {V}ariational {B}ayes.
\newblock {\em arXiv preprint arXiv:1312.6114}, 2013.

\bibitem{Goodfellow2014Generative}
Ian Goodfellow, Jean Pouget-Abadie, Mehdi Mirza, Bing Xu, David Warde-Farley,
  Sherjil Ozair, Aaron Courville, and Yoshua Bengio.
\newblock Generative {A}dversarial {N}ets.
\newblock In {\em Advances in Neural Information Processing Systems}, pages
  2672--2680, 2014.

\bibitem{Kipf2016Variational}
Thomas~N Kipf and Max Welling.
\newblock Variational {G}raph {A}uto-{E}ncoders.
\newblock {\em arXiv preprint arXiv:1611.07308}, 2016.

\bibitem{Wang2018Graphgan}
Hongwei Wang, Jia Wang, Jialin Wang, Miao Zhao, Weinan Zhang, Fuzheng Zhang,
  Xing Xie, and Minyi Guo.
\newblock Graph{G}{A}{N}: Graph {R}epresentation {L}earning with {G}enerative
  {A}dversarial {N}ets.
\newblock In {\em Thirty-Second AAAI Conference on Artificial Intelligence},
  2018.

\bibitem{Bonner2019Temporal}
Stephen Bonner, Amir Atapour-Abarghouei, Philip~T Jackson, John Brennan, Ibad
  Kureshi, Georgios Theodoropoulos, Andrew~Stephen McGough, and Boguslaw Obara.
\newblock Temporal {N}eighbourhood {A}ggregation: {P}redicting {F}uture {L}inks
  in {T}emporal {G}raphs via {R}ecurrent {V}ariational {G}raph {C}onvolutions.
\newblock In {\em 2019 IEEE International Conference on Big Data (Big Data)},
  pages 5336--5345. IEEE, 2019.

\bibitem{Zhao2019Large}
Yifeng Zhao, Xiangwei Wang, Hongxia Yang, Le~Song, and Jie Tang.
\newblock Large {S}cale {E}volving {G}raphs with {B}urst {D}etection.
\newblock In {\em 28th International Joint Conference on Artificial
  Intelligence (IJCAI)}, 2019.

\bibitem{Hajiramezanali2019Variational}
Ehsan Hajiramezanali, Arman Hasanzadeh, Krishna Narayanan, Nick Duffield,
  Mingyuan Zhou, and Xiaoning Qian.
\newblock Variational {G}raph {R}ecurrent {N}eural {N}etworks.
\newblock In {\em Advances in Neural Information Processing Systems}, pages
  10700--10710, 2019.

\bibitem{Ma2020Streaming}
Yao Ma, Ziyi Guo, Zhaocun Ren, Jiliang Tang, and Dawei Yin.
\newblock Streaming {G}raph {N}eural {N}etworks.
\newblock In {\em Proceedings of the 43rd International ACM SIGIR Conference on
  Research and Development in Information Retrieval}, pages 719--728, 2020.

\bibitem{Xu2019Adaptive}
Dongkuan Xu, Wei Cheng, Dongsheng Luo, Yameng Gu, Xiao Liu, Jingchao Ni,
  Bo~Zong, Haifeng Chen, and Xiang Zhang.
\newblock Adaptive {N}eural {N}etwork for {N}ode {C}lassification in {D}ynamic
  {N}etworks.
\newblock In {\em 2019 IEEE International Conference on Data Mining (ICDM)},
  pages 1402--1407. IEEE, 2019.

\bibitem{Xu2021Transformer}
Dongkuan Xu, Junjie Liang, Wei Cheng, Hua Wei, Haifeng Chen, and Xiang Zhang.
\newblock Transformer-{S}tyle {R}elational {R}easoning with {D}ynamic {M}emory
  {U}pdating for {T}emporal {N}etwork {M}odeling.
\newblock In {\em Proceedings of the AAAI Conference on Artificial
  Intelligence}, pages 4546--4554, 2021.

\bibitem{Park2019Exploiting}
Hogun Park and Jennifer Neville.
\newblock Exploiting {I}nteraction {L}inks for {N}ode classification with deep
  graph neural networks.
\newblock In {\em Proceedings of the 28th International Joint Conference on
  Artificial Intelligence}, pages 3223--3230. AAAI Press, 2019.

\bibitem{Shrestha2019Learning}
Prasha Shrestha, Suraj Maharjan, Dustin Arendt, and Svitlana Volkova.
\newblock Learning from {D}ynamic {U}ser {I}nteraction {G}raphs to {F}orecast
  {D}iverse {S}ocial {B}ehavior.
\newblock In {\em Proceedings of the 28th ACM International Conference on
  Information and Knowledge Management}, pages 2033--2042. ACM, 2019.

\bibitem{Mikolov2013Efficient}
Tomas Mikolov, Kai Chen, Greg Corrado, and Jeffrey Dean.
\newblock Efficient {E}stimation of {W}ord {R}epresentations in {V}ector
  {S}pace.
\newblock {\em arXiv preprint arXiv:1301.3781}, 2013.

\bibitem{Mikolov2013Distributed}
Tomas Mikolov, Ilya Sutskever, Kai Chen, Greg~S Corrado, and Jeff Dean.
\newblock Distributed {R}epresentations of {W}ords and {P}hrases and their
  {C}ompositionality.
\newblock In {\em Advances in Neural Information Processing Systems}, pages
  3111--3119, 2013.

\bibitem{De2018Combining}
Sam De~Winter, Tim Decuypere, Sandra Mitrovi{\'c}, Bart Baesens, and Jochen
  De~Weerdt.
\newblock Combining {T}emporal {A}spects of {D}ynamic {N}etworks with
  {N}ode2{V}ec for a {M}ore {E}fficient {D}ynamic {L}ink {P}rediction.
\newblock In {\em 2018 IEEE/ACM ASONAM}, pages 1234--1241. IEEE, 2018.

\bibitem{Singer2019Node}
Uriel Singer, Ido Guy, and Kira Radinsky.
\newblock Node embedding over temporal graphs.
\newblock In {\em Proceedings of the Twenty-Eighth International Joint
  Conference on Artificial Intelligence, {IJCAI-19}}, pages 4605--4612.
  International Joint Conferences on Artificial Intelligence Organization, 7
  2019.

\bibitem{Mahdavi2018dynnode2vec}
Sedigheh Mahdavi, Shima Khoshraftar, and Aijun An.
\newblock dynnode2vec: Scalable dynamic network embedding.
\newblock In {\em 2018 IEEE International Conference on Big Data (Big Data)},
  pages 3762--3765. IEEE, 2018.

\bibitem{Bamler2017Dynamic}
Robert Bamler and Stephan Mandt.
\newblock Dynamic {W}ord {E}mbeddings.
\newblock {\em arXiv preprint arXiv:1702.08359}, 2017.

\bibitem{Heidari2018EvoNRL}
Farzaneh Heidari and Manos Papagelis.
\newblock Evo{N}{R}{L}: Evolving {N}etwork {R}epresentation {L}earning {B}ased
  on {R}andom {W}alks.
\newblock In {\em International Conference on Complex Networks and their
  Applications}, pages 457--469. Springer, 2018.

\bibitem{Sajjad2019Efficient}
Hooman~Peiro Sajjad, Andrew Docherty, and Yuriy Tyshetskiy.
\newblock Efficient {R}epresentation {L}earning using {R}andom {W}alks for
  {D}ynamic {G}raphs.
\newblock {\em arXiv preprint arXiv:1901.01346}, 2019.

\bibitem{Yu2018NetWalk}
Wenchao Yu, Wei Cheng, Charu~C Aggarwal, Kai Zhang, Haifeng Chen, and Wei Wang.
\newblock Net{W}alk: A {F}lexible {D}eep {E}mbedding {A}pproach for {A}nomaly
  {D}etection in {D}ynamic {N}etworks.
\newblock In {\em Proceedings of the 24th ACM SIGKDD Int. Conference on
  Knowledge Discovery \& Data Mining}, pages 2672--2681, 2018.

\bibitem{Pandhre2018stwalk}
Supriya Pandhre, Himangi Mittal, Manish Gupta, and Vineeth~N Balasubramanian.
\newblock S{T}{W}alk: {L}earning {T}rajectory {R}epresentations in {T}emporal
  {G}raphs.
\newblock In {\em Proceedings of the ACM India Joint International Conference
  on Data Science and Management of Data}, pages 210--219. ACM, 2018.

\bibitem{Khoshraftar2019Dynamic}
Shima Khoshraftar, Sedigheh Mahdavi, Aijun An, Yonggang Hu, and Junfeng Liu.
\newblock Dynamic {G}raph {E}mbedding via {L}{S}{T}{M} {H}istory {T}racking.
\newblock In {\em 2019 IEEE International Conference on Data Science and
  Advanced Analytics (DSAA)}, pages 119--127. IEEE, 2019.

\bibitem{wang2018attention}
Zhitao Wang, Chengyao Chen, and Wenjie Li.
\newblock Attention network for information diffusion prediction.
\newblock In {\em Companion Proceedings of the The Web Conference 2018}, pages
  65--66, 2018.

\bibitem{Wang2017Topological}
Jia Wang, Vincent~W Zheng, Zemin Liu, and Kevin Chen-Chuan Chang.
\newblock Topological {R}ecurrent {N}eural {N}etwork for {D}iffusion
  {P}rediction.
\newblock In {\em IEEE ICDM}, pages 475--484, 2017.

\bibitem{Yang2019Multi}
Cheng Yang, Jian Tang, Maosong Sun, Ganqu Cui, and Zhiyuan Liu.
\newblock Multi-{S}cale {I}nformation {D}iffusion {P}rediction with
  {R}einforced {R}ecurrent {N}etworks.
\newblock In {\em Proceedings of the 28th International Joint Conference on
  Artificial Intelligence}, pages 4033--4039. AAAI Press, 2019.

\bibitem{Sato2019DyANE}
Koya Sato, Mizuki Oka, Alain Barrat, and Ciro Cattuto.
\newblock Dy{A}{N}{E}: Dynamics-{A}ware {N}ode {E}mbedding for {T}emporal
  {N}etworks.
\newblock {\em arXiv preprint arXiv:1909.05976}, 2019.

\bibitem{Beres2019Node}
Ferenc B{\'e}res, Domokos~M Kelen, R{\'o}bert P{\'a}lovics, and Andr{\'a}s~A
  Bencz{\'u}r.
\newblock Node {E}mbeddings in {D}ynamic {G}raphs.
\newblock {\em Applied Network Science}, 4(1):64, 2019.

\bibitem{Liu2019Towards}
Zhining Liu, Dawei Zhou, and Jingrui He.
\newblock Towards {E}xplainable {R}epresentation of {T}ime-{E}volving {G}raphs
  via {S}patial-{T}emporal {G}raph {A}ttention {N}etworks.
\newblock In {\em Proceedings of the 28th ACM International Conference on
  Information and Knowledge Management}, pages 2137--2140. ACM, 2019.

\bibitem{Wu2019Efficiently}
Tianxing Wu, Arijit Khan, Huan Gao, and Cheng Li.
\newblock Efficiently {E}mbedding {D}ynamic {K}nowledge {G}raphs.
\newblock {\em arXiv preprint arXiv:1910.06708}, 2019.

\bibitem{Jiang2016Encoding}
Tingsong Jiang, Tianyu Liu, Tao Ge, Lei Sha, Sujian Li, Baobao Chang, and
  Zhifang Sui.
\newblock Encoding {T}emporal {I}nformation for {T}ime-{A}ware {L}ink
  {P}rediction.
\newblock In {\em Proceedings of the 2016 Conference on Empirical Methods in
  Natural Language Processing}, pages 2350--2354, 2016.

\bibitem{Bordes2013Translating}
Antoine Bordes, Nicolas Usunier, Alberto Garcia-Duran, Jason Weston, and Oksana
  Yakhnenko.
\newblock Translating {E}mbeddings for {M}odeling {M}ulti-{R}elational {D}ata.
\newblock In {\em Advances in Neural Information Processing Systems}, pages
  2787--2795, 2013.

\bibitem{Levy2014Neural}
Omer Levy and Yoav Goldberg.
\newblock Neural {W}ord {E}mbedding as {I}mplicit {M}atrix {F}actorization.
\newblock In {\em Advances in Neural Information Processing Systems}, pages
  2177--2185, 2014.

\bibitem{Tang2015Line}
Jian Tang, Meng Qu, Mingzhe Wang, Ming Zhang, Jun Yan, and Qiaozhu Mei.
\newblock L{I}{N}{E}: Large-{S}cale {I}nformation {N}etwork {E}mbedding.
\newblock In {\em Proceedings of the 24th International Conference on World
  Wide Web}, pages 1067--1077. International World Wide Web Conferences
  Steering Committee, 2015.

\bibitem{Liu2019General}
Xin Liu, Tsuyoshi Murata, Kyoung-Sook Kim, Chatchawan Kotarasu, and Chenyi
  Zhuang.
\newblock A {G}eneral {V}iew for {N}etwork {E}mbedding as {M}atrix
  {F}actorization.
\newblock In {\em Proceedings of the Twelfth ACM International Conference on
  Web Search and Data Mining}, pages 375--383. ACM, 2019.

\bibitem{Rahman2018dylink2vec}
Mahmudur Rahman, Tanay~Kumar Saha, Mohammad~Al Hasan, Kevin~S Xu, and Chandan~K
  Reddy.
\newblock Dy{L}ink2{V}ec: Effective {F}eature {R}epresentation for {L}ink
  {P}rediction in {D}ynamic {N}etworks.
\newblock {\em arXiv preprint arXiv:1804.05755}, 2018.

\bibitem{Lu2019Temporal}
Yuanfu Lu, Xiao Wang, Chuan Shi, Philip~S Yu, and Yanfang Ye.
\newblock Temporal {N}etwork {E}mbedding with {M}icro-and {M}acro-dynamics.
\newblock In {\em Proceedings of the 28th ACM International Conference on
  Information and Knowledge Management}, pages 469--478. ACM, 2019.

\bibitem{Yin2019mhdne}
Ying Yin, Jianpeng Zhang, Yulong Pei, Xiaotao Cheng, and Lixin Ji.
\newblock M{HDNE}: Network {E}mbedding {B}ased on {M}ultivariate {H}awkes
  {P}rocess.
\newblock In {\em Joint European Conference on Machine Learning and Knowledge
  Discovery in Databases}, pages 409--421. Springer, 2019.

\bibitem{Rossi2018Relational}
Ryan~A Rossi.
\newblock Relational {T}ime {S}eries {F}orecasting.
\newblock {\em The Knowledge Engineering Review}, 33, 2018.

\bibitem{Pei2016Node}
Yulong Pei, Jianpeng Zhang, GH~Fletcher, and Mykola Pechenizkiy.
\newblock Node {C}lassification in {D}ynamic {S}ocial {N}etworks.
\newblock {\em Proceedings of AALTD}, page~54, 2016.

\bibitem{junuthula2016evaluating}
Ruthwik~R Junuthula, Kevin~S Xu, and Vijay~K Devabhaktuni.
\newblock Evaluating link prediction accuracy in dynamic networks with added
  and removed edges.
\newblock In {\em 2016 IEEE international conferences on big data and cloud
  computing (BDCloud), social computing and networking (SocialCom), sustainable
  computing and communications (SustainCom)(BDCloud-SocialCom-SustainCom)},
  pages 377--384. IEEE, 2016.

\bibitem{Lamprier2018Variational}
Sylvain Lamprier.
\newblock A {V}ariational {T}opological {N}eural {M}odel for {C}ascade-based
  {D}iffusion in {N}etworks.
\newblock {\em arXiv preprint arXiv:1812.10962}, 2018.

\bibitem{Zhang2018Cosine}
Yuan Zhang, Tianshu Lyu, and Yan Zhang.
\newblock C{O}{S}{I}{N}{E}: Community-{P}reserving {S}ocial {N}etwork
  {E}mbedding from {I}nformation {D}iffusion {C}ascades.
\newblock In {\em Thirty-Second AAAI Conference on Artificial Intelligence},
  2018.

\bibitem{Maaten2008Visualizing}
Laurens van~der Maaten and Geoffrey Hinton.
\newblock Visualizing {D}ata using t-sne.
\newblock {\em Journal of Machine Learning Research}, 9(Nov):2579--2605, 2008.

\bibitem{Goyal2018DynamicGEM}
Palash Goyal, Sujit~Rokka Chhetri, Ninareh Mehrabi, Emilio Ferrara, and
  Arquimedes Canedo.
\newblock Dynamic{G}{E}{M}: A {L}ibrary for {D}ynamic {G}raph {E}mbedding
  {M}ethods.
\newblock {\em arXiv preprint arXiv:1811.10734}, 2018.

\bibitem{Feng2019Hypergraph}
Yifan Feng, Haoxuan You, Zizhao Zhang, Rongrong Ji, and Yue Gao.
\newblock Hypergraph {N}eural {N}etworks.
\newblock In {\em Proceedings of the AAAI Conference on Artificial
  Intelligence}, volume~33, pages 3558--3565, 2019.

\bibitem{Zhang2018DynamicHypergraph}
Zizhao Zhang, Haojie Lin, Yue Gao, and KLISS BNRist.
\newblock Dynamic {H}ypergraph {S}tructure {L}earning.
\newblock In {\em IJCAI}, pages 3162--3169, 2018.

\bibitem{Cheng2016Graph}
Xiaodong Cheng, Yu~Kawano, and Jacquelien~MA Scherpen.
\newblock Graph {S}tructure-{P}reserving {M}odel {R}eduction of {L}inear
  {N}etwork {S}ystems.
\newblock In {\em 2016 European Control Conference (ECC)}, pages 1970--1975.
  IEEE, 2016.

\end{thebibliography}
\end{document}